\def\longversion{}\documentclass[letterpaper]{article} % DO NOT CHANGE THIS
\ifdefined\longversion\usepackage{myaaai2026}\else\usepackage{aaai2026}\fi  % DO NOT CHANGE THIS
\ifdefined\longversion\usepackage[hidelinks]{hyperref}\fi
\IfPackageLoadedTF{hyperref}{\def\hyperhyper}{}

\usepackage[utf8]{inputenc}
\usepackage{times}  % DO NOT CHANGE THIS
\usepackage{helvet}  % DO NOT CHANGE THIS
\usepackage{courier}  % DO NOT CHANGE THIS
\ifdefined\longversion\else\usepackage[hyphens]{url}\fi  % DO NOT CHANGE THIS
\usepackage{graphicx} % DO NOT CHANGE THIS
\usepackage{natbib}  % DO NOT CHANGE THIS AND DO NOT ADD ANY OPTIONS TO IT
\usepackage{caption} % DO NOT CHANGE THIS AND DO NOT ADD ANY OPTIONS TO IT
\ifdefined\longversion\nocopyright\fi % -- Your paper will not be published if you use this command
\title{Non-Monotonic S4F Standpoint Logic\iflong\\\mbox{(Extended Version with Proofs)}\fi}
\author{%
    Piotr Gorczyca\textsuperscript{\rm 1},
    Hannes Strass\textsuperscript{\rm 1, 2}
}
\affiliations{
    \textsuperscript{\rm 1}Computational Logic Group, Faculty of Computer Science, TUD Dresden University of Technology, Germany

    \textsuperscript{\rm 2}ScaDS.AI Center for Scalable Data Analytics and Artificial Intelligence, Dresden/Leipzig, Germany

    firstname.lastname@tu-dresden.de
}

\usepackage{amssymb,amsmath}
\usepackage{xcolor}
\usepackage{xspace}
\usepackage[thmmarks,amsmath]{ntheorem}
\usepackage{cleveref}
\usepackage{enumerate}
\usepackage{enumitem}

\usepackage{mathtools}
\ifdefined\longversion\else\usepackage{scalerel}\fi

\newcommand{\wrt}{w.r.t.\xspace}

\newcommand{\define}[1]{\emph{#1}}

\newcommand{\ebnfeq}{\mathrel{::=}}
\newcommand{\ebnfalt}{\mathbin{\mid}}

\newcommand{\lang}{\mathcal{L}}

\newcommand{\eqdef}{\mathrel{:=}}

\newcommand{\proves}{\vdash}
\newcommand{\modelfor}{\Vdash}

\newcommand{\thry}[1]{\mathit{Th}(#1)}
\newcommand{\entails}{\models}

\newcommand{\subf}{\mathit{Sub}}

\newcommand{\limplies}{\rightarrow}

\newcommand{\sfstruct}{\mathfrak{X}}
\newcommand{\val}{E}

\newcommand{\restrict}[2]{#1\mkern-1mu|_{#2}} % restrict function #1's domain to #2

\newcommand{\set}[1]{\left\{#1\right\}}
\newcommand{\guard}{\ \middle\vert\ }
\newcommand{\card}[1]{\left\vert#1\right\vert}
\newcommand{\size}[1]{\left\Vert#1\right\Vert}
\newcommand{\tuple}[1]{\left(#1\right)}
\newcommand{\tand}{\text{ and }\,}
\newcommand{\tiff}{\text{ iff }}

\newcommand{\iffdef}{\mathbin{\,\mathord{:}\mkern-6mu\mathord{\iff}}}

\newcommand{\phiff}{\mathbin{\phantom{\iff}}}
\newcommand{\phiffdef}{\mathbin{\phantom{\iffdef}}}

\newcommand{\cclass}[1]{\ensuremath{\mathsf{#1}}\xspace}
\newcommand{\PTime}{\cclass{P}}
\newcommand{\NP}{\cclass{NP}}
\newcommand{\coNP}{\cclass{coNP}}
\newcommand{\SigmaP}[1][2]{\cclass{\Sigma^P_{#1}}}
\newcommand{\PiP}[1][2]{\cclass{\Pi^P_{#1}}}
\newcommand{\PSpace}{\ensuremath{\mathsf{PSpace}}\xspace}

\newcommand{\thr}{T}

\newcommand{\rex}{\textsc{o}} % Running EXample
\newcommand{\thrinit}{D^{\rex}}
\newcommand{\throne}{\ensuremath{\thrinit_1}}
\newcommand{\thrtwo}{\ensuremath{\thrinit_2}}
\newcommand{\sffsstructtwo}{\sffsstructn{\rex}}
\newcommand{\SFFSstructtwo}{\SFFSstructn{\rex}}
\newcommand{\sffsstructtwomin}{\sffsstructtwo'}
\newcommand{\SFFSstructtwomin}{\SFFSstructn[']{\rex}}
\newcommand{\Precstwo}{\Precs_{\rex}}
\newcommand{\stanotwo}{\stano_{\rex}}
\newcommand{\stanitwo}{\stani_{\rex}}
\newcommand{\evaltwo}{\eval_{\rex}}
\newcommand{\stanitwomin}{\stani_{\rex}'}

\newcommand{\Atoms}{\mathcal{A}}
\newcommand{\Atomsmod}{\Atoms^{\pm}}

\newcommand{\slogic}{\ensuremath{\mathbb{S}}\xspace}
\newcommand{\Stands}{\mathcal{S}}
\newcommand{\langs}{\lang_\slogic} % standpoint logic, formulas only
\newcommand{\langss}{\langs^{\scalebox{0.666}{\ensuremath{\mathord{\sharpens}}}}} % standpoint logic *with* sharpening statements

\newcommand{\Precs}{\Pi}
\newcommand{\stan}{\sigma}
\newcommand{\eval}{\gamma}
\newcommand{\spform}[1]{\mathsf{#1}}
\ifdefined\longversion\newcommand{\spformtext}[1]{\ensuremath{\spform{#1}}}\else\newcommand{\spformtext}[1]{\ensuremath{\scaleobj{0.9}{\ensuremath{\spform{#1}}}}}\fi
\newcommand{\all}{\ast}
\newcommand{\sts}{\spform{s}}
\newcommand{\stt}{\spform{t}}
\newcommand{\stu}{\spform{u}}
\newcommand{\str}{\spform{r}}

\newcommand{\pr}{\pi}

\newcommand{\prcopy}[1]{#1'}

\newcommand{\standb}[1]{\mathord{\mathop{\Box}\nolimits_{\spform{#1}}}}
\ifdefined\longversion\newcommand{\standbx}[1]{\mathord{\mathop{\Box}\nolimits_{\spform{#1}}}}\else\newcommand{\standbx}[1]{\mathord{\mathop{\Box}\nolimits_{\scaleobj{0.8}{\spform{#1}}}}}\fi
\newcommand{\standd}[1]{\mathord{\mathop{\Diamond}\nolimits_{\spform{#1}}}}
\ifdefined\longversion\newcommand{\standdx}[1]{\mathord{\mathop{\Diamond}\nolimits_{\spform{#1}}}}\else\newcommand{\standdx}[1]{\mathord{\mathop{\Diamond}\nolimits_{\scaleobj{0.8}{\spform{#1}}}}}\fi
\newcommand{\standbs}{\standb{s}}
\newcommand{\standbu}{\standb{u}}
\newcommand{\standbt}{\standb{t}}
\newcommand{\standbr}{\standb{r}}
\newcommand{\standball}{\standb{\all}}
\newcommand{\standds}{\standd{s}}

\newcommand{\sstruct}{\mathfrak{N}}

\newcommand{\Sstruct}{\tuple{\Precs,\stan,\eval}}

\newcommand{\prp}[1]{\pr'}

\newcommand{\stani}{\left(\stanis\right)_{\sts\in\Stands}}

\newcommand{\sff}{S4F\xspace}

\newcommand{\know}{\mathord{\mathbf{K}}}

\newcommand{\langk}{\lang_{\know}}

\newcommand{\partnk}{\Phi}
\newcommand{\partk}{\Psi}

\newcommand{\sffstruct}{\mathfrak{M}}

\newcommand{\SFFstruct}{\tuple{\wf,\ws,\wm}}

\newcommand{\wf}{V}
\newcommand{\ws}{W}
\newcommand{\wt}{X}
\newcommand{\wm}{\xi}
\newcommand{\wmv}{\chi}
\newcommand{\wrld}{w}

\newcommand{\SFstruct}{\tuple{\wt,\wmv}}
\newcommand{\entailssff}{\entails_{\text{S4F}}}

\newcommand{\mlax}[1]{\textbf{#1}\xspace}
\newcommand{\axK}{\mlax{K}}
\newcommand{\axT}{\mlax{T}}
\newcommand{\axf}{\mlax{4}}
\newcommand{\axfive}{\mlax{5}}
\newcommand{\axF}{\mlax{F}}
\newcommand{\lpand}{,\,}
\newcommand{\lpnot}{\mathord{\sim}}
\newcommand{\akset}{A^{\know}}
\newcommand{\sffpartall}{\Theta}
\newcommand{\sfpart}{\tuple{\partnk,\partk}}

\renewcommand{\stani}{\stan}
\newcommand{\stano}{\tau}

\newcommand{\sffsstruct}{\mathfrak{F}}
\newcommand{\sffsstructt}{\sffsstruct'}

\newcommand{\SFFSstruct}{\tuple{\Precs,\stani,\stano,\eval}}
\newcommand{\SFFSstructt}{\tuple{\Precs',\stani',\stano',\eval'}}

\newcommand{\sffsstructn}[1]{\mathfrak{F}_{#1}}
\newcommand{\SFFSstructn}[2][]{\tuple{\Precs_{#2}#1,\stani_{#2}#1,\stano_{#2}#1,\eval_{#2}#1}}

\newcommand{\provess}{\proves_{\slogic}}
\newcommand{\entailss}{\entails_{\slogic}}
\newcommand{\sharpens}{\preccurlyeq}

\newcommand{\sprefeq}{\trianglelefteq}
\newcommand{\sprefeqs}[1][\sts]{\sprefeq_{#1}}
\newcommand{\sprefiso}{\simeq}
\newcommand{\spref}{\vartriangleleft}

\newcommand{\uniq}[1][E]{\psi_{\bar{#1}}}
\newcommand{\bigland}{\bigwedge}

\newcommand{\modalsub}[1][\thr]{#1^{\Box}}
\newcommand{\partition}[1][\thr]{\Xi(#1)}
\newcommand{\Partition}[1][]{\tuple{\partnk#1,\partk#1}}

\newcommand{\partprop}[1][\sts]{\Theta_{#1}}
\newcommand{\sffsstructpartition}[1][\partition]{\sffsstruct_{#1}}

\ifdefined\longversion\ifdefined\hyperhyper\else\newcommand{\hypertarget}[2]{#2\label{#1}}\fi\else\newcommand{\hypertarget}[2]{#2\label{#1}}\fi
\ifdefined\longversion\ifdefined\hyperhyper\else\newcommand{\hyperlink}[2]{#2}\fi\else\newcommand{\hyperlink}[2]{#2}\fi

\newcommand{\caid}{{\normalfont(\textbf{C1})}}
\newcommand{\cbid}{{\normalfont(\textbf{C2})}}
\newcommand{\ccid}{{\normalfont(\textbf{C3})}}
\newcommand{\ca}{\hyperlink{itm:ca}{\caid}\xspace}
\newcommand{\cb}{\hyperlink{itm:cb}{\cbid}\xspace}
\newcommand{\cc}{\hyperlink{itm:cc}{\ccid}\xspace}

\newcommand{\unknown}[1][\sffsstruct]{U_{#1}}

\newcommand{\cred}{\mathsf{cred}}
\newcommand{\scep}{\mathsf{scep}}
\newcommand{\Tcred}[1][\xi]{\thr_{\cred}^{#1}}
\newcommand{\Tscep}[1][\xi]{\thr_{\scep}^{#1}}

\newcommand{\standbf}[2]{\standbx{#1}(#2)}
\newcommand{\default}[3]{{#1\mathbin{:}{#2}{/}{#3}}}

\newcommand{\dentails}{\mathbin{\mathord{\mid}\mkern-3.5mu\mathord{\approx}}}
\newcommand{\ndentails}{\mathbin{\mathord{\mid}\mkern-3.5mu\mathord{\not\approx}}}
\newcommand{\dentailsscep}{\dentails_{\text{scep}}}
\newcommand{\ndentailsscep}{\ndentails_{\text{scep}}}
\newcommand{\dentailscred}{\dentails_{\text{cred}}}

\newcommand{\atom}[1]{\ensuremath{\mathtt{#1}}}

\newcommand{\preg}{\atom{Preg}}

\newcommand{\ovudis}{\atom{OvuDis}}

\newcommand{\fha}{\atom{FHA}}

\newcommand{\pcos}{\atom{PCOS}}

\newcommand{\horm}{\atom{Horm}}

\newcommand{\spcos}{\atom{P}}
\newcommand{\shorm}{\atom{H}}
\newcommand{\spreg}{\atom{Pr}}
\newcommand{\sfha}{\atom{F}}
\newcommand{\sovu}{\atom{O}}

\newcommand{\smed}{\spform{M}}
\newcommand{\smedtext}{\spformtext{M}}
\newcommand{\sdone}{\spform{D1}}
\newcommand{\sdonetext}{\spformtext{D1}}
\newcommand{\sdtwo}{\spform{D2}}
\newcommand{\sdtwotext}{\spformtext{D2}}

\newcommand{\myparagraph}[1]{\vskip2pt\noindent\textbf{#1}\xspace}

\newlength{\mygatherskip}
\newenvironment{mygather}[1][2pt]{\setlength{\mygatherskip}{#1}\vskip\mygatherskip\mbox{}\hfill\(}{\)\hfill\mbox{}\vskip\mygatherskip}

\newcommand{\labeleddot}[3]{%
    \node at #1 {#2:};
    \node[below, xshift=12] at #1 {#3};
}

\definecolor{myPastelPink}{cmyk}{0,0.25,0.20,0}
\definecolor{myPastelYellow}{cmyk}{0,0,0.20,0}
\definecolor{myPastelBlue}{cmyk}{0.25,0.06,0,0.10}
\definecolor{myPastelGreen}{cmyk}{0.13,0,0.13,0.10}

\theoremseparator{.}
\newtheorem{theorem}{Theorem}
\newtheorem{definition}[theorem]{Definition}
\newtheorem{lemma}[theorem]{Lemma}
\newtheorem{proposition}[theorem]{Proposition}
\newtheorem{corollary}[theorem]{Corollary}
\newtheorem{example}[theorem]{Example}

\theoremstyle{plain}
\theoremheaderfont{\normalfont\itshape}
\theorembodyfont{\normalfont}
\newtheorem{claim}{Claim}

\theoremstyle{nonumberplain}
\theoremheaderfont{\normalfont\bfseries}
\theorembodyfont{\normalfont}
\theorempreskip{1pt}
\theorempostskip{-\topsep}
\theoremsymbol{\ensuremath{\square}}
\newtheorem{proof}{Proof}
\newtheorem{proofsketch}{Proof (Sketch)}
\theoremheaderfont{\normalfont\itshape}
\theoremsymbol{\ensuremath{\vartriangle}}
\newtheorem{claimproof}{Proof of Claim~\theclaim}

\usepackage{ifthen}
\usepackage{environ}
\newif\ifshowdiscussion
\newif\ifshowsketch
\newif\iflong
\NewEnviron{shortproof}{\iflong\else\expandafter\begin{proof}\BODY\end{proof}\fi}
\NewEnviron{shortproofsketch}{\iflong\else\begin{proofsketch}\BODY\end{proofsketch}\fi}
\NewEnviron{longproof}{\iflong\begin{proof}\BODY\end{proof}\fi}
\NewEnviron{discussion}{\ifshowdiscussion\expandafter\BODY\fi}

\NewEnviron{longonly}{\iflong\expandafter\BODY\fi}
\NewEnviron{shortonly}{\iflong\else\expandafter\BODY\fi}
\newcommand{\shortonlymac}[1]{\iflong\else\expandafter{#1}\fi}
\newcommand{\longonlymac}[1]{\iflong\expandafter{#1}\fi}

\usepackage{tikz,pgf}
\usetikzlibrary{shadings}
\ifdefined\longversion\longtrue\else\longfalse\fi

\begin{document}

\maketitle

% expand -- \input{main}
%%%%%%%%%%%%%%%%%%%%%%%%%%%%%%%%%%%%%%%%
\begin{abstract}
    % expand -- \input{sections/abstract}
    Standpoint logics offer unified modal logic-based formalisms for representing multiple heterogeneous viewpoints.
    At the same time, many non-monotonic reasoning frameworks can be naturally captured using modal logics -- in particular using the modal logic S4F.
    In this work, we propose a novel formalism called S4F Standpoint Logic, which generalises both S4F and propositional standpoint logic and is therefore capable of expressing multi-viewpoint, non-monotonic semantic commitments.
    We define its syntax and semantics and analyze its computational complexity, obtaining the result that S4F Standpoint Logic is not computationally harder than its constituent logics, whether in monotonic or non-monotonic form.
    We also outline mechanisms for credulous and sceptical acceptance and illustrate the framework with an example.
\end{abstract}

\iflong\else
    \begin{links}
        \link{Code}{https://github.com/cl-tud/nm-s4fsl-asp}
        \link{Extended version}{https://arxiv.org/abs/2511.10449}
    \end{links}
\fi

% expand -- \input{sections/introduction}
%%%%%%%%%%%%%%%%%%%%%%%%%%%%%%%%%%%%%%%%%%%
\section{Introduction}

% BEGIN import from NMR
Standpoint logic is a modal logic-based formalism for representing multiple diverse (and potentially conflicting) viewpoints within a single framework.
Its main appeal derives from its conceptual simplicity and its attractive properties:
In the presence of conflicting information, standpoint logic sacrifices neither consistency nor logical conclusions about the shared understanding of common vocabulary~\citep{GomezAlvarezR21}.
The underlying idea is to start from a base logic (originally propositional logic; \citealp{GomezAlvarezR21}) and to enhance it with two modalities pertaining to what holds according to certain \emph{standpoints}.
There, a standpoint is a specific point of view that an agent or other entity can take, and that has a bearing on how the entity understands and employs a given logical vocabulary (that may at the same time be used by other entities with a potentially different understanding).
The two modalities are, respectively:
$\standbs\phi$, expressing
“it is unequivocal [from the point of view $\sts$] that $\phi$”;
and its dual
$\standds\phi$, where
“it is conceivable [from the point of view $\sts$] that $\phi$”.

Standpoint logic escapes global inconsistency by keeping conflicting pieces of knowledge separate, yet avoids duplication of vocabulary and in this way conveniently keeps portions of common understanding readily available.
% TODO: cite van Fraassen on supervaluationism?
It has historic roots within the philosophical theory of supervaluationism~\citep{Bennett11}, which explains semantic variability “by the fact that natural language can be interpreted in many different yet equally acceptable ways, commonly referred to as \emph{precisifications}”~\citep{GomezAlvarezR21}.

In our work, such semantic commitments can be made on the basis of incomplete knowledge using a form of default reasoning.
Consequently, in our logic each standpoint embodies a consistent (but possibly partial) point of view, potentially using non-monotonic reasoning (NMR) to arrive there.
This entails that the overall formalism becomes non-monotonic with respect to its logical conclusions.

Several non-monotonic formalisms that could be employed for default reasoning within standpoints come to mind, and obvious criteria for selection among the candidates are not immediate.
We choose to employ the non-monotonic modal logic S4F~\citep{segerberg,SchwarzT94}, which is a very general formalism that subsumes several other NMR languages, decidedly allowing the possibility for later specialisation via restricting to proper fragments.
In this way, we obtain standpoint versions of
default logic~\citep{Reiter80} (see also \Cref{example} below),
answer set programming~\citep{GelfondL91}, and
abstract argumentation~\citep{Dung95},
all as corollaries of our general approach.
The usefulness of non-monotonic \sff for knowledge representation and especially non-monotonic reasoning has been aptly demonstrated by \citet{SchwarzT94} (among others), but seems to be underappreciated in the literature to this day.
% END import from NMR

We illustrate the logic we propose by showcasing a worked example in \emph{standpoint default logic}, a standpoint variant of \citeauthor{Reiter80}'s default logic~\citeyearpar{Reiter80}, where defaults and definite knowledge can be annotated with standpoint modalities.
In~\Cref{example}, defaults are of the standard form, namely $\default{\phi}{\psi_1,\ldots,\psi_n}{\xi}$ where (as usual) if the \emph{prerequisite} $\phi$ is established and there is no evidence to the contrary of the justifications $\psi_1,\ldots,\psi_n$, then the consequence $\xi$ is concluded.
An \emph{extension} is a deductively closed set of formulas representing one possible belief set derived by maximally applying the defaults.
A sentence follows \emph{credulously} if some extension entails it, and \emph{sceptically} if all extensions do.
\begin{example}\label{example}
    Ovulation disorders are among the leading causes of female infertility.
    Their origins and diagnostics vary, and medical communities do not agree on a unified diagnosis or treatment.
    For example, community $\sdonetext$ typically attributes ovulation disorders to \emph{polycystic ovary syndrome} ($\pcos$), while community $\sdtwotext$ often sees \emph{functional hypothalamic amenorrhea} ($\fha$) as their main source (unless the patient is pregnant, $\preg$).
    The initial treatment for PCOS, as generally accepted in the overall medical community $\smedtext$ -- including its subcommunities $\sdonetext$ and $\sdtwotext$ -- involves hormone therapy ($\horm$);
    however, this should be avoided in FHA, as it may be ineffective and could mask the underlying issue.%
    % \footnote{This simplified scenario illustrates a real issue in medical diagnosis, but has been shortened for brevity.
    %       In reality, the divisions are more fine-grained, with definitions put forward by groups like the Androgen Excess and PCOS Society or developed in various 
    %       % such as the 2003 ESHRE/ASRM %Rotterdam
    %       % workshop or the 1990 NIH consensus workshop
    %       workshops~\cite[Table~1]{teede2010pcos}.
    %       %\pg{TODO: If this can be addressed in the current logic, mention that the complete example will appear in the full version.}
    % }
    \footnote{We show a simplified example of a real-world case where different bodies define PCOS by the presence or absence of certain symptoms~\cite[Table~1]{teede2010pcos}, showcasing the
        standpoint
        non-monotonic reasoning captured by our logic.}
    % 
    % \footnote{This simplified scenario illustrates a real issue in medical diagnosis, but has been shortened for brevity.
    %       In reality, the divisions are more fine-grained, with definitions put forward by groups like the Androgen Excess and PCOS Society or developed in workshops such as the 2003 ESHRE/ASRM %Rotterdam
    %       workshop or the 1990 NIH consensus workshop~\cite[Table~1]{teede2010pcos}.
    %       %\pg{TODO: If this can be addressed in the current logic, mention that the complete example will appear in the full version.}
    % }
    This can be formalised as a standpoint default theory\/:
    \begin{align*}
        \thrinit\eqdef \big\{ & \mathrlap{\standbf{\sdone}{\default{\ovudis}{\pcos}{\pcos}},}           &  &  & \sdonetext \sharpens \smedtext,         &        \\
                              & \mathrlap{\standbf{\sdtwo}{\default{\ovudis}{\neg \preg, \fha}{\fha}},} &  &  & \sdtwotext \sharpens \smedtext,         &        \\
                              & \;\standbf{\smed}{\pcos\limplies\horm},                                 &  &  & \standbf{\smed}{\fha\limplies\neg\horm} & \big\}
    \end{align*}
    Now assume a patient is diagnosed with ovulation disorders, \mbox{$\throne\eqdef\thrinit\cup\set{\standball\ovudis}$}.
    The unique standpoint extension of $\throne$ yields $\standbf{\sdone}{\pcos \land \horm}$ and $\standbf{\sdtwo}{\fha \land \neg\horm}$, so these conclusions follow sceptically.
    The patient or physician may choose a treatment based on the reputation and trust attributed to the community from which the conclusion derives.
    If it is later learned that the patient is in fact pregnant
    -- \mbox{$\thrtwo\eqdef\throne\cup\set{\standball\preg}$} --
    then, $\standbf{\sdtwo}{\fha\land\neg\horm}$ is withdrawn, whereas $\standbf{\sdone}{\pcos\land\horm}$ remains.

    Compare this with related logics:
    A plain default theory is obtained by dropping the standpoint modalities.
    It yields two extensions corresponding to the previous conclusions, which follow credulously but not sceptically.
    Additionally, the information about which standpoint each conclusion derives from is lost.
    % \iflong 
    % % extended example for the long version
    % Alternatively, if we turn the default rules into strict implications -- i.e. $\standbf{\sdone}{\ovudis \limplies \pcos}$ (1) and $\standbf{\sdtwo}{\ovudis \land \neg\preg \limplies \fha}$ (2) -- we get a propositional standpoint logic theory.
    % In this case, the only consequence regarding diagnosis and treatment is $\standbf{\sdone}{\pcos \land \horm}$ obtained from the rule (1). Rule (2) requires that the pregnancy status be known to obtain the conclusion. Once $\standball\neg\preg$ is established, we also conclude $\standbf{\sdtwo}{\fha \land \neg\horm}$ from (2). However, in this case, adding $\standball\preg$ makes the theory obviously inconsistent.
    % \else % short version example
    Alternatively, propositional standpoint logic with strict implications -- e.g., \mbox{$\standbf{\sdone}{\ovudis \limplies \pcos}$} and \mbox{$\standbf{\sdtwo}{(\ovudis \land \neg\preg) \limplies \fha}$} -- does not support default reasoning, as it lacks means of expressing non-monotonicity.
    % \fi
\end{example}

% Specifically, 
In this work, we introduce syntax and semantics of \emph{\sff standpoint logic}, a combination of \sff and standpoint logic that generalises both, including a non-monotonic semantics for default reasoning.
We study the computational complexity of the logic and observe that the extension of unimodal \sff to multiple standpoint modalities does not incur any additional computational cost.
We conclude with an implementation of our logic in disjunctive answer set programming and a discussion of avenues for future work.

%\iflong\else An extended version of the paper containing all proofs and additional results is available as supplementary material.\fi

\paragraph{Related Work.}%
% expand -- \input{sections/related-work}
Several \emph{monotonic} logics have been “standpointified” so far:
Apart from propositional logic in the original work of \citet{GomezAlvarezR21}, also first-order logic and several description logics (\citealp{AlvarezRS22}, \citeyear{AlvarezRS23a}, \citeyear{AlvarezRS23b}), the temporal logic LTL~\citep{GiganteAL23,DemriW24,abs-2502-20193},
and most recently even monodic fragments of first-order logic with counting~\citep{GomezAlvarezR24,GomezAlvarezR25}.

In previous, preliminary work~(\citeyear{GS2024}), we introduced a different, more restrictive non-monotonic standpoint logic framework:
a two-dimensional modal logic, a \emph{product logic}~\cite{Kurucz03} of standpoint logic with \sff, however only considering a fragment of the language (where all formulas are of the form $\standbs\phi$ or $\standds\phi$ with $\phi$ not containing further standpoint modalities).
%
% \citet{LeisegangMR24} integrated standpoint modalities into another non-monotonic framework, that of KLM propositional logics~\citep{KrausLM90}, albeit in a restricted setting that disallows negation/disjunction of formulas with standpoint modalities.
%
% \citet{LeisegangMV2025} integrated both regular and novel defeasible standpoint modalities into propositional KLM, thereby extending rather than restricting the syntax of propositional standpoint logic, though at the cost of PSPACE reasoning complexity.
\citet{LeisegangMR24} integrated standpoint modalities into KLM propositional logics~\citep{KrausLM90} in a restricted setting (disallowing negation or disjunction of formulas with modalites).
\citet{LeisegangMV2025} proposed a slightly different KLM-based defeasible standpoint logic offering standpoint modalities and sharpenings that can both be defeasible, but where logical entailment is still monotonic and defeasible implication is among propositional formulas only.
Regarding non-monotonic multi-modal logics, \citet{Rosati06} generalised the logic of GK~\citep{LinS92} to the multi-agent case, however with an explicit focus on \emph{epistemic} interpretations of modalities and resulting introspection capabilities.

% \input{sections/background}
%%%%%%%%%%%%%%%%%%%%%%%%%%%%%%%%%%%%%%%
\section{Background}
\label{sec:background}

All languages we henceforth consider build upon propositional logic, denoted $\lang$, built from a set $\Atoms$ of atoms according to $\varphi\ebnfeq p\ebnfalt\neg \varphi\ebnfalt\varphi\land\varphi$ where $p\in\Atoms$, allowing the usual notational shorthands $\varphi\lor\psi$ and $\varphi\limplies\psi$.
Its model-theoretic semantics is given by valuations $\val\subseteq\Atoms$ containing exactly the true atoms.
We denote satisfaction of a formula $\varphi$ by a valuation $\val$ as \mbox{$\val\modelfor\varphi$} and entailment of a formula $\varphi$ by a set \mbox{$\thr\subseteq\lang$} of formulas as \mbox{$\thr\entails\varphi$}.
By $\subf(\varphi)$ we denote the set of all subformulas of $\varphi$ (also for logics introduced later and similarly for theories $\thr$).
For functions \mbox{$f\colon A\to B$}, we denote $f(C)\eqdef\set{f(a)\guard a\in C}$ for $C\subseteq A$.

\subsection{Standpoint Logic}
Standpoint logic (SL) was introduced as a modal logic-based formalism for representing multiple (potentially contradictory) perspectives in a single framework~\cite{GomezAlvarezR21}.
Building upon propositional logic, in addition to a set $\Atoms$ of propositional atoms, it uses a set $\Stands$ of \define{standpoint names}, where a standpoint represents a point of view an agent or other entity can take, with \mbox{$\all\in\Stands$} being the \define{universal} standpoint.
Formally, the syntax of propositional standpoint logic is given by
\(
\varphi \ebnfeq \bot \ebnfalt p \ebnfalt \neg\varphi \ebnfalt \varphi\land\varphi \ebnfalt \standbs\varphi \ebnfalt \sts\sharpens\stu
\)
where \mbox{$p\in\Atoms$}, and \mbox{$\sts,\stu\in\Stands$}.
An expression \mbox{$\sts\sharpens\stu$} is called a \emph{sharpening statement} and states that all semantic commitments of standpoint $\stu$ are inherited by standpoint $\sts$.
%Modality $\standbs$ has a dual, $\standds\varphi\eqdef\neg\standbs\neg\varphi$, read as “according to standpoint $\sts$, it is \emph{conceivably} the case that $\varphi$”.

The semantics of standpoint logic is given by \define{standpoint structures} $\sstruct=\Sstruct$, where $\Precs$ is a non-empty set of \define{precisifications} (i.e.~worlds), $\stan\colon\Stands\to 2^{\Precs}$ assigns a set of precisifications to each standpoint name (with $\stan(\all)=\Precs$ fixed), and $\eval\colon\Precs\to 2^\Atoms$ assigns a propositional valuation $E\subseteq\Atoms$ to each precisification.
The satisfiaction relation $\sstruct,\pr\modelfor\varphi$ for $\pr\in\Precs$ is defined by structural induction as follows\/:%
\begin{align*}
    \sstruct,\pr & \modelfor p                      &  & \iffdef p\in\eval(\pr)                                                      \\
    \sstruct,\pr & \modelfor\neg\varphi             &  & \iffdef \sstruct,\pr\not\modelfor\varphi                                    \\
    \sstruct,\pr & \modelfor\varphi_1\land\varphi_2 &  & \iffdef \sstruct,\pr\modelfor\varphi_1 \tand \sstruct,\pr\modelfor\varphi_2 \\
    \sstruct,\pr & \modelfor\standbs\varphi         &  & \iffdef \sstruct,\pr'\modelfor\varphi \text{ for all } \pr'\in\stan(\sts)   \\
    \sstruct,\pr & \modelfor\sts\sharpens\stu       &  & \iffdef \stan(\sts)\subseteq\stan(\stu)
    % HS: following line maybe not needed
    %\sstruct\phantom{,\pr} & \modelfor \varphi                &  & \iffdef \sstruct,\pr\modelfor\varphi\text{ for all } \pr\in\Precs
\end{align*}
%A standpoint structure $\sstruct=\Sstruct$ is a \define{model} for a formula $\varphi$ iff $\sstruct\modelfor\varphi$; a formula $\varphi\in\langs$ is \define{satisfiable} iff there exists an $\sstruct=\Sstruct$ and a $\pr\in\Precs$ such that $\sstruct,\pr\modelfor\varphi$.
%
Standpoint structures can be regarded as a restricted form of ordinary (multi-modal) Kripke structures $\tuple{\Precs,\set{R_\sts}_{\sts\in\Stands},\eval}$, where the worlds are the precisifications $\Precs$, the evaluation function of worlds is $\eval$, and the reachability relation for a standpoint name (i.e.\ modality) $\sts\in\Stands$ is $R_\sts=\Precs\times\stan(\sts)$.
\subsection{Modal Logic \sff}

The modal logic \sff is a unimodal logic extending the logic S4 (characterised by axioms \axK, \axT, and \axf) by axiom schema
% HS: preferably take the F axiom from Marek, Schwarz, and Trusz
\vskip4pt
\mbox{}\hfill
$(\varphi \land \neg\know\neg\know\psi) \limplies \know(\neg\know\neg\varphi \lor \psi)$
\hfill($\axF$)
\vskip4pt
\noindent and was studied in depth by \citet{segerberg}.
We are chiefly interested in its non-monotonic semantics and thus approach the logic from its (more accessible) model theory.
The syntax of \sff $\langk$ is given by
\(
\varphi \ebnfeq p \ebnfalt \neg\varphi \ebnfalt \varphi\land\varphi \ebnfalt \know\varphi
\)
with $p\in\Atoms$.
The semantics of \sff is given by \define{\sff structures}, tuples $\sffstruct=\SFFstruct$ where
$V$ and $W$ are disjoint sets of \define{worlds} with $W\neq\emptyset$, and
$\wm\colon\wf\cup\ws\to 2^\Atoms$ assigns to each world $w$ a valuation $\wm(\wrld)\subseteq\Atoms$.
The satisfaction relation $\sffstruct,\wrld\modelfor\varphi$ for $\wrld\in\wf\cup\ws$ is then defined by induction\/:%
\begin{align*}
    \sffstruct,w & \modelfor p                       &  & \iffdef p\in\wm(\wrld)                                                      \\
    \sffstruct,w & \modelfor \neg\varphi             &  & \iffdef \sffstruct,w\not\modelfor \varphi                                   \\
    \sffstruct,w & \modelfor \varphi_1\land\varphi_2 &  & \iffdef \sffstruct,w\modelfor\varphi_1 \tand \sffstruct,w\modelfor\varphi_2 \\
    \sffstruct,w & \modelfor \know\varphi            &  & \;\iffdef                                                                   \\ % HS: unclear why this :<=> is moved to the left 
                 & \mathrlap{
        \begin{cases}
            \sffstruct,v\modelfor\varphi \text{ for all } v\in\wf\cup\ws & \text{ if } w\in\wf, \\
            \sffstruct,v\modelfor\varphi\text{ for all } v\in\ws         & \text{ otherwise.}
        \end{cases}
    }
\end{align*}
%A pointed \sff structure $\sffstruct,\wrld$ is a \define{model} of a formula $\varphi$ iff $\sffstruct,\wrld\modelfor\varphi$;
%$\sffstruct,\wrld$ is a \define{model} of a theory $T\subseteq\langk$ iff $\sffstruct,\wrld\modelfor\varphi$ for all $\varphi\in T$.
%A formula $\varphi\in\langk$ is \define{satisfiable} iff there exists an \sff structure $\sffstruct=\SFFstruct$ and a world $w\in\wf\cup\ws$ such that $\sffstruct,\wrld\modelfor\varphi$ (likewise for theories $T$).
An \sff structure $\sffstruct=\SFFstruct$ is a \define{model} of a formula $\varphi\in\langk$ (theory $A\subseteq\langk$), written $\sffstruct\modelfor\varphi$ ($\sffstruct\modelfor A$) iff for all $\wrld\in\wf\cup\ws$, we have $\sffstruct,\wrld\modelfor\varphi$ (for each $\varphi\in A$).
A theory $A\subseteq\langk$ is \define{satisfiable} iff there is an \sff structure $\sffsstruct$ such that $\sffsstruct\modelfor A$.\iflong\footnote{There is sometimes a distinction between \emph{local} satisfiability of a formula $\varphi\in\langk$ -- there exists an \sff structure $\sffsstruct=\SFFstruct$ and a world $w\in\wf\cup\ws$ such that $\sffsstruct,w\modelfor\varphi$ -- and \emph{global} satisfiability (the notion we use). In our context, the two are interchangeable: $\varphi$ is globally satisfiable iff $\standball\varphi$ is locally satisfiable; $\varphi$ is locally satisfiable iff $\standball\varphi$ is globally satisfiable.}\else\ \fi
A formula $\varphi\in\langk$ is \define{entailed} by a theory $A$, written \mbox{$A\entailssff\varphi$}, iff every model of $A$ is a model of $\varphi$.

\sff structures can also be seen as a restricted form of Kripke structures \mbox{$(\wf\cup\ws, R, \wm)$} with reachability relation \mbox{\( R \eqdef(\wf\!\times\!\wf) \cup (\wf\!\times\!\ws) \cup (\ws\!\times\!\ws) \)},
comprising two clusters of worlds: \emph{outer} worlds $\wf$ and \emph{inner} worlds $\ws$.
%The outer worlds $\wf$ can reach all (inner and outer) worlds, while the inner worlds $\ws$ can only reach all inner worlds.
% Intuitively, while all worlds $\wf\cup\ws$ are globally possible, in any inner world, the outer worlds are \emph{not known to be possible}. I.e. for $\sffstruct=\SFFstruct$, for all \mbox{$\wrld\in\ws$}, \mbox{$v\in\wf\cup\ws$} we have \mbox{$\sffstruct,\wrld\modelfor\varphi \iff \sffstruct,v\modelfor\know\neg\know\neg\varphi$}.
%Intuitively, the inner worlds are \emph{known to be possible}, %i.e.,~for $\sffstruct=\SFFstruct$ and all $\wrld\in\ws$, if $\sffstruct,\wrld\modelfor\varphi$ then for all $v\in\wf\cup\ws$ it holds that $\sffstruct,v\modelfor\know\neg\know\neg\varphi$;
%whereas this does not necessarily hold for the outer worlds $\wf$.
% Intuitively, for $\sffstruct=\SFFstruct$ only the $\ws$ worlds are necessarily \emph{known to be possible}, i.e. for all \mbox{$\wrld\in\ws$}, \mbox{$v\in\wf\cup\ws$} we have \mbox{$\sffstruct,\wrld\modelfor\varphi \iff \sffstruct,v\modelfor\know\pos\varphi$}.
%
Intuitively, for an \sff structure $\SFFstruct$ all worlds \mbox{$\wf\cup\ws$} are globally possible, but there is an important distinction whether this possibility is \emph{known}:
All \emph{inner} worlds are \emph{known to be possible} in any world, while this does not necessarily hold for the \emph{outer} worlds.
Thus $\wf$ can possibly affect what is known overall, but not what is known in $\ws$.
% More technically, if for an inner world $w\in\ws$ we have $\sffstruct,w\modelfor\varphi$, then $\sffstruct\modelfor\know\neg\know\neg\varphi$; on the other hand, if for an outer world $v\in\wf$ we have $\sffstruct,v\modelfor\psi$, then possibly $\sffstruct\not\modelfor\know\neg\know\neg\psi$.
%
\sff has a small model property; its satisfiability problem is \NP-complete~\citep{Schwarz-Truszczynski-1993}, entailment “$A\entailssff\varphi$?” is \coNP-complete.
% An \sff structure $\sffstruct=\tuple{\emptyset,\ws,\wm}$ is called S5, denoted \mbox{$\sfstruct=\SFstruct$}.
% An \sff structure $\sffstruct=\tuple{\emptyset,\ws,\wm}$ is called S5, denoted \mbox{$\sfstruct=\tuple{\ws,\wm}$}.
An \sff structure $\SFFstruct$ with \mbox{$\wf=\emptyset$} is called an \define{S5 structure} and just denoted $\tuple{\ws,\wm}$.

\subsubsection{Non-Monotonic \sff}

A non-monotonic logic is obtained by restricting attention to models where what is known is minimal~\citep{SchwarzT94}:
Given an S5 structure \mbox{$\sfstruct=\SFstruct$}, an \sff structure \mbox{$\sffstruct=\SFFstruct$} is \define{strictly preferred} over $\sfstruct$ iff
\mbox{$\ws=\wt$}, \mbox{$\restrict{\wm}{\ws}=\wmv$}, and %there exists a \mbox{$\wrld\in\wf$} such that for all \mbox{$\wrldp\in\wt$}, \mbox{$\wm(\wrld)\neq\wmv(\wrldp)$}.
there is a \mbox{$\wrld\in\wf$} with \mbox{$\wm(\wrld)\notin\wmv(\wt)$} (in other words $\wm(\wf)\not\subseteq\wmv(\wt)$).
An S5 structure $\sfstruct=\SFstruct$ is a \define{minimal model} of a theory $A\subseteq\langk$ iff
$\sfstruct$ is a model for $A$ and
there is no \sff structure $\sffstruct$ that is strictly preferred to $\sfstruct$ with $\sffstruct$ a model for $A$.
%Intuitively, $\sfstruct$ having a strictly preferred alternative means that the knowledge of $\sfstruct$ is not minimal.
Intuitively, if $\sffstruct$ is strictly preferred to $\sfstruct$ then in $\sffstruct$ we know strictly less and the knowledge of $\sfstruct$ is thus not minimal.

\subsubsection{\sff in Knowledge Representation}
\label{subsect:sff-in-kr}

The logic \sff is immensely useful for knowledge representation purposes~\cite{Schwarz-Truszczynski-1993,SchwarzT94}, as it allows to naturally embed several important logics and formalisms for non-monotonic reasoning, including, but not limited to\/:

\myparagraph{Default Logic.}
For a general default
\(
\default{\phi}{\psi_1,\ldots,\psi_m}{\xi}
\)
\citep{Reiter80},
the corresponding \sff formula is given by
\mbox{$( \know\phi\land \know\neg\know\neg\psi_1 \land \ldots \land \know\neg\know\neg\psi_m ) \limplies \know\xi$}.

\myparagraph{Logic Programs.}
A given normal logic program rule
\mbox{\(
    p_0\gets p_1\lpand \ldots \lpand p_m \lpand \lpnot p_{m+1} \lpand \ldots \lpand \lpnot p_{m+n}
    \)}
is translated into
\(
(\know p_1 \land \ldots \land \know p_m \land \know\neg\know p_{m+1} \land \ldots \land \know\neg\know p_{m+n}) \limplies \know p_0
\), and similar translation results exist for extended and disjunctive logic programs~\cite{SchwarzT94}.

\myparagraph{Argumentation Frameworks.}
An AF \mbox{$F=(A,R)$} (under stable semantics) is translated into the \sff theory
$T_F\eqdef\set{ \know\neg\know\neg a \limplies \know a \guard a\in A } \cup \set{ \know a\limplies\know\neg b \guard (a,b)\in R }$, which follows from Dung's translation of AFs into normal logic programs~[\citeauthor{Dung95}, \citeyear{Dung95}, Section~5; \citeauthor{Strass13}, \citeyear{Strass13}].

The above formalisms can be modularly (piece by piece, without looking at the entire theory) and faithfully (preserving the semantics one-to-one) embedded into \sff.
Further \sff embeddings are possible:
e.g.\ the (bimodal) logic of GK by \citet{LinS92} as well as the (bimodal) logic of MKNF by \citet{Lifschitz94}, all with \sff being unimodal and thus arguably offering a simpler semantics.

Among the non-monotonic modal logics capable of faithfully embedding the above formalisms, \sff stands out
% It is worth noting that other non-monotonic modal logics are also capable of faithfully embedding the above formalisms.
% In fact, there exists a \define{range} of modal logics with that capability, including e.g.\ T$^{-}$ (the logic containing the T axiom only), K, T or S4.
%\sff however stands out 
as a prominent candidate for several reasons, including (but not limited to) its intuitive model theory and its relative computational easiness compared to other logics
(e.g.\ S4, for which satisfiability is \PSpace-complete;\ \citealt{HalpernM92}).
Furthermore, other logics need not guarantee modularity (e.g.\ KD45;\ \citealt{Gottlob95a,Moore85}) or have issues with {explicit definitions}~\cite{SchwarzT94}.

\subsubsection{Complexity of Non-Monotonic \sff}
\label{sec:sff-complexity}
The decision problems associated with non-monotonic \sff were found to reside on the second level of the polynomial hierarchy~\cite{Schwarz-Truszczynski-1993}.
As we later generalise it, below we sketch the procedure for deciding whether a given \sff theory \mbox{$A\subseteq\langk$} has a minimal model.
The actual minimal model $\sfstruct$ cannot always be explicitly constructed due to potentially containing exponentially many worlds (\wrt $A$), thus a “smaller” representation is required.
The idea for this -- going back to \citet{Shvarts90} -- is to represent $\sfstruct$ by giving all subformulas $\know\phi$ of $A$ with $\sfstruct\modelfor\know\phi$.

More technically, denote \mbox{$\akset\eqdef\set{\varphi\guard\know\varphi\in\subf(A)}$};
a \emph{partition} $\sfpart$ of $\akset$ then intuitively represents an S5 structure $\sfstruct$ in which all formulas in $\partk$ are known and all formulas in $\partnk$ are not known.
Surprisingly, it can be decided whether $\sfstruct$ is a minimal model of $A$ by consulting only $\sfpart$:
Formally, \citet{Schwarz-Truszczynski-1993} define
\begin{mygather}
    \sffpartall\eqdef A\cup\set{\neg\know\varphi\guard\varphi\in\partnk}\cup\set{\know\psi\guard\psi\in\partk}\cup\partk
\end{mygather}
\noindent which is read as a theory of propositional logic with subformulas $\know\phi$ regarded as atoms that are independent of $\phi$.
Now the pair $\sfpart$ corresponds to a minimal \sff model $\sfstruct$ of $A$ iff
(a) $\sffpartall$ is satisfiable in propositional logic,
(b) for each $\phi\in\partnk$, we have $\sffpartall\not\entails\phi$, and
(c) for each $\psi\in\partk$, we have $A\cup\set{\neg\know\phi\guard\phi\in\partnk}\entailssff\psi$.
%Intuitively, as $\sfstruct$ is an S5 structure, for every $\varphi\in\akset$ it holds that $\sfstruct\modelfor\know\varphi$ or $\sfstruct\modelfor\neg\know\varphi$.
%Towards knowledge minimisation, the subset of $\akset$ for which the latter holds should be $\subseteq$-maximal (guaranteed by condition c) as well as $\sfpart$ must actually correspond to a proper S5 model of $A$ (conditions a and b).
Since the number of \NP oracle calls to verify (a--c) is polynomial (actually linear) in $\card{\akset}$, we get containment in $\SigmaP[2]$~\cite{Schwarz-Truszczynski-1993}.
A matching lower bound follows from the faithful embedding of default logic~\cite{Reiter80} into \sff~\citep{SchwarzT94} and \citeauthor{Gottlob92}'s result on the complexity of extension existence in default logic~\citeyearpar{Gottlob92}.

% \input{sections/s4f-standpoint-logic}
%%%%%%%%%%%%%%%%%%%%%%%%%%%%%%%%%%%%%%%%%%%%%%%%%
\section{Non-Monotonic S4F Standpoint Logic}

Herein, we propose non-monotonic multi-modal S4F with modalities restricted as in standpoint logic.
% We proceed to show that all “nice” properties of the combined logics will be preserved through the combination.
We proceed to show that the combination of the two logics exhibits all ``nice'' properties of the constituents.
The syntax of the new logic is almost the same as ordinary standpoint logic;
we just disallow to nest sharpening statements into formulas.\footnote{This is not a severe restriction; several standpoint logics or precursors have similar conditions, with no known issues~\citep{GomezAlvarez19,AlvarezRS23a,AlvarezRS23b}.}

\begin{definition}
    \label{def:sffs:syntax}
    The language $\langss$ of \sff standpoint logic over atoms $\Atoms$ and standpoint names $\Stands$ contains
    \define{expressions} $\varphi$ that are either
    \define{sharpening statements} $\sts\sharpens\stu$
    or \define{formulas} $\psi$ of propositional multi-modal logic with modalities $\Stands$,~i.e.,
    \[
        \varphi \ebnfeq \sts\sharpens\stu \ebnfalt \psi
        \quad\text{with}\quad
        \psi \ebnfeq p \ebnfalt \neg\psi \ebnfalt \psi_1\land\psi_2 \ebnfalt \standbs\psi
    \]
    where \mbox{$p\in\Atoms$} and \mbox{$\sts,\stu\in\Stands$}.
    A \define{theory} is a subset \mbox{$\thr\subseteq\langss$}.
\end{definition}
As usual, we require the universal standpoint \mbox{$\all\in\Stands$}. %: intuitively, $\all$ is the \emph{universal} standpoint that encompasses all (other) points of view.
By $\langs$ we denote the proper fragment that contains formulas only.

Thus in the actual syntax of our logic, a stand\-point-an\-notated default \mbox{$\standbs(\default{\phi}{\psi_1,\ldots,\psi_n}{\xi})$} is syntactic sugar for a formula
\mbox{$(\standbs\phi\land\standbs\neg\standbs\neg\psi_1\land\ldots\land\standbs\neg\standbs\neg\psi_n)\limplies\standbs\xi$}.
For instance, \mbox{$\standbf{M}{\default{\ovudis}{\pcos}{\pcos}}$} of \Cref{example} stands for \mbox{$(\standbx{M}\ovudis\land\standbx{M}\neg\standbx{M}\neg\pcos)\limplies\standbx{M}\pcos$}.

The major novel aspects of our logic are the new monotonic and non-monotonic semantics.
We start with introducing the structures that are used in both model theories.

\begin{definition}
    \label{def:sffs:structures}
    Let $\Atoms$ be a set of propositional atoms and $\Stands$ be a set of standpoint names.
    An \define{S4F standpoint structure} (over $\Atoms$ and $\Stands$) is a tuple
    \mbox{$\sffsstruct=\SFFSstruct$} where
    \begin{itemize}
        \item $\Precs$ is a non-empty set of precisifications,
        \item $\stani,\stano\colon\Stands\to 2^{\Precs}$ are functions such that
              $\stani(\all)\cup\stano(\all)=\Precs$ and
              for all $\sts,\stu\in\Stands$:
              %   \begin{itemize}
              %       \item $\stani(\sts)\neq\emptyset$,
              %       \item $\stani(\sts)\cap\stano(\stu)=\emptyset$, and
              %   \end{itemize}
              $\stani(\sts)\neq\emptyset$ and
              $\stani(\sts)\cap\stano(\stu)=\emptyset$;
        \item $\eval\colon\Precs\to 2^{\Atoms}$ maps every precisification $\pr\in\Precs$ to a propositional valuation $\val\subseteq\Atoms$.
    \end{itemize}
\end{definition}

\sff standpoint structures thus mainly comprise precisifications \mbox{$\pr\in\Precs$} that each have a propositional valuation $\eval(\pr)$ attached as before.
Additionally, each standpoint $\sts$ distinguishes \emph{inner precisifications} $\stani(\sts)$ and \emph{outer precisifications} $\stano(\sts)$ that are disjoint across all standpoints, with $\all$ still being universal.
Intuitively, every standpoint $\sts\in\Stands$ has “its own” associated \sff structure $(\stano(\sts),\stani(\sts),\eval)$. %$(\stano(\sts),\stani(\sts),\restrict{\eval}{\stani(\sts)\cup\stano(\sts)})$,
This insight motivates the following definition of the satisfaction relation.

\begin{definition}
    \label{def:sffs:satisfaction}
    Let \mbox{$\sffsstruct=\SFFSstruct$} be an S4F standpoint structure and \mbox{$\varphi\in\langss$}.
    For \mbox{$\pr\in\Precs$}, the (pointed) satisfaction relation \mbox{$\sffsstruct,\pr\modelfor\varphi$} is defined by structural induction\/:\\[-4pt]
    \begin{align*}
        \sffsstruct,\pr           & \modelfor p                 &  & \iffdef p\in\eval(\pr)                                                            \\
        \sffsstruct,\pr           & \modelfor \neg\psi          &  & \iffdef \sffsstruct,\pr\not\modelfor\psi                                          \\
        \sffsstruct,\pr           & \modelfor \psi_1\land\psi_2 &  & \iffdef \sffsstruct,\pr\modelfor\psi_1 \tand \sffsstruct,\pr\modelfor\psi_2       \\
        \sffsstruct,\pr           & \modelfor \standbs\psi      &  & \;\iffdef                                                                         \\
                                  & \;\mathrlap{
            \begin{cases}
                \sffsstruct,\pr'\modelfor\psi \;\text{ for all } \pr'\in\stani(\sts)                 & \text{ if } \pr\in\stani(*), \\
                \sffsstruct,\pr'\modelfor\psi \;\text{ for all } \pr'\in\stani(\sts)\cup\stano(\sts) & \text{ otherwise.}
            \end{cases}
        }                                                                                                                                              \\
        \sffsstruct,\pr           & \modelfor \sts\sharpens\stu &  & \iffdef \stani(\sts)\subseteq\stani(\stu) \tand \stano(\sts)\subseteq\stano(\stu) \\
        \sffsstruct\phantom{,\pr} & \modelfor \varphi           &  & \iffdef \sffsstruct,\pr\modelfor\varphi \;\text{ for all } \pr\in\Precs
    \end{align*}
\end{definition}
As usual,
%a pointed structure $\sffsstruct,\pr$ is a \define{model} of \mbox{$\varphi\in\langss$} iff \mbox{$\sffsstruct,\pr\modelfor\varphi$};
a structure $\sffsstruct$ is a \define{model} of a theory \mbox{$\thr\subseteq\langss$} iff \mbox{$\sffsstruct\modelfor\varphi$} for all \mbox{$\varphi\in\thr$};
%a formula \mbox{$\varphi\in\langss$} (theory $\thr\subseteq\langss$) is \define{satisfiable} iff there exists a structure \mbox{$\sffsstruct=\SFFSstruct$} and a precisification \mbox{$\pr\in\Precs$} such that \mbox{$\sffsstruct,\pr\modelfor\varphi$} (\mbox{$\sffsstruct\modelfor\thr$});
a theory $\thr\subseteq\langss$ is \define{satisfiable} iff there exists a structure \mbox{$\sffsstruct$} such that \mbox{$\sffsstruct\modelfor\thr$};
a formula \mbox{$\varphi\in\langss$} is \define{entailed} by a theory \mbox{$\thr\subseteq\langss$}, denoted \mbox{$\thr\entailss\varphi$}, iff every model of $\thr$ is a model of $\varphi$.
Again, \sff standpoint structures can be recast as ordinary (multi-modal) Kripke structures \mbox{$(\Precs,\set{R_\sts}_{\sts\in\Stands},\eval)$} where each \mbox{$\sts\in\Stands$} has reachability relation
\mbox{$R_\sts\eqdef \stano(\all)\times\stano(\sts)\cup\stano(\all)\times\stani(\sts)\cup\stani(\all)\times\stani(\sts)$}.
\begin{figure}
    \centering
    \bgroup
    \footnotesize
    \begin{tikzpicture}
        \path[use as bounding box] (0,-0.2) rectangle (8.2,4.2);

        \begin{scope}[yshift=2.3cm]
            \draw[rounded corners=2pt, left color=black!5, right color=white, shading angle=0] (0, 1.9) rectangle (8.2, -0.2);
            \path[draw, fill=white]
            (.9, 1.48) -- (0, 1.48) -- (0, 1.83) arc[start angle=180, end angle=90, radius=2pt] -- (.9, 1.9) -- cycle;

            \node at (0.45, 1.7) {\small $\stanotwo(\all)$};

            \draw[rounded corners=2pt, left color=myPastelYellow, right color=white, shading angle=0] (2.3, 0) rectangle (8.08, 1.7);
            \path[draw, fill=white] (3.25, 1.28) -- (2.3, 1.28) -- (2.3, 1.63) arc[start angle=180, end angle=90, radius=2pt] -- (3.25, 1.7) -- cycle;
            \node at (2.775, 1.5) {\small $\stanotwo(\spform{M})$};

            \draw[rounded corners=2pt, left color=myPastelPink, right color=white, shading angle=0] (3.35, 0.15) rectangle (7.96, 1.55);
            \path[draw, fill=white] (4.44, 1.13) -- (3.35, 1.13) -- (3.35, 1.48) arc[start angle=180, end angle=90, radius=2pt] -- (4.44, 1.55) -- cycle;
            \node at (3.9,1.33) {\small $\stanotwo(\spform{D2})$};

            \draw[rounded corners=2pt, left color=myPastelBlue, right color=white, shading angle=0] (6, 0.3) rectangle (7.85, 1.4);
            \path[draw, fill=white] (7.07, 0.98) -- (6, 0.98) -- (6, 1.33) arc[start angle=180, end angle=90, radius=2pt] -- (7.07, 1.4) -- cycle;
            \node at (6.525, 1.17) {\small $\stanotwo(\spform{D1})$};

            \labeleddot{(3.7,0.75)}{$\pr_{3}$}{$\set{\sovu, \spreg, \sfha}$}

            \labeleddot{(5.05, .75)}{$\;\pr_4$}{$\!\!\!\!\set{\sovu, \spreg}$}

            \labeleddot{(0.4,1.1)}{$\pr_1$}{$\qquad\set{\sovu, \spreg, \shorm, \spcos, \sfha}$}
            \labeleddot{(0.4, 0.4)}{$\pr_2$}{$\quad\set{\sovu, \spreg, \shorm, \sfha}$}

        \end{scope}

        % INNER
        \draw[rounded corners=2pt, left color=black!5, right color=white, shading angle=0] (0, 1.9) rectangle (8.2, -0.2);
        \path[draw, fill=white] (.9, 1.48) -- (0, 1.48) -- (0, 1.83) arc[start angle=180, end angle=90, radius=2pt] -- (.9, 1.9) -- cycle;
        \node at (0.45, 1.7)  {\small $\stanitwo(\all)$};

        \draw[rounded corners=2pt, left color=myPastelYellow, right color=white, shading angle=0] (2.3, 0) rectangle (8.08, 1.7);
        \path[draw, fill=white] (3.25, 1.28) -- (2.3, 1.28) -- (2.3, 1.63) arc[start angle=180, end angle=90, radius=2pt] -- (3.25, 1.7) -- cycle;
        \node at (2.775, 1.5) {\small $\stanitwo(\spform{M})$};

        % \draw[rounded corners=2pt, left color=myPastelBlue, right color=white, shading angle=0] (4.05, 0.2) rectangle (6.05, 1.5);
        % \node at (4.65, 1.25) {$\underline{\stani(\spform{D1})}$};

        \draw[rounded corners=2pt, left color=myPastelPink, right color=white, shading angle=0] (3.35, 0.15) rectangle (7.96, 1.55);
        \path[draw, fill=white] (4.44, 1.13) -- (3.35, 1.13) -- (3.35, 1.48) arc[start angle=180, end angle=90, radius=2pt] -- (4.44, 1.55) -- cycle;
        \node at (3.9,1.33) {\small $\stanitwo(\spform{D2})$};

        \draw[rounded corners=2pt, left color=myPastelBlue, right color=white, shading angle=0] (6, 0.3) rectangle (7.85, 1.4);
        \path[draw, fill=white] (7.07, 0.98) -- (6, 0.98) -- (6, 1.33) arc[start angle=180, end angle=90, radius=2pt] -- (7.07, 1.4) -- cycle;
        \node at (6.55, 1.17) {\small $\stanitwo(\spform{D1})$};

        \labeleddot{(6.5, 0.8)}{$\!\!\!\!\!\!\pr_{8}$}{$\set{\sovu, \spreg, \shorm, \spcos}$}

        \labeleddot{(3.7,0.75)}{$\pr_{7}$}{$\set{\sovu, \spreg, \shorm}$}

        \labeleddot{(0.4,1.1)}{$\pr_5$}{$\set{\sovu, \spreg, \spcos}$}
        \labeleddot{(0.4, 0.4)}{$\pr_6$}{$\quad\set{\sovu, \spreg, \spcos, \sfha}$}
    \end{tikzpicture}
    \egroup
    \caption{%
        An \sff standpoint structure \mbox{$\sffsstructtwo=\SFFSstructtwo$} that is a model of theory $\thrtwo$ from~\Cref{example}.
        Precisifications \mbox{$\Precstwo=\set{\pr_1,\ldots,\pr_8}$} within a box belong to the outer (upper) or inner (lower) set of precisifications of the standpoint labelling the box,
        % e.g.~\mbox{$\stano(\smedtext)=\stano(\sdonetext)=\set{\pr_3}$}. % $\stani(\sdonetext)=\set{\pr_7}$.
        e.g.~\mbox{$\pr_8\in\stanitwo(\sdonetext)\subseteq\stanitwo(\sdtwotext)$}. % $\stani(\sdonetext)=\set{\pr_7}$.
        (Satisfaction \mbox{$\sffsstructtwo\modelfor\sdonetext\sharpens\sdtwotext$} is coincidental and not required by $\thrtwo$.)
        Precisification $\pr$'s valuation is shown as a set $\evaltwo(\pr)$ of atoms below $\pr$.
        Atoms are abbreviated thus: $\ovudis$ ($\sovu$), $\preg$ ($\spreg$), $\horm$ ($\shorm$), $\pcos$ ($\spcos$), and $\fha$ ($\sfha$).
        For example, \mbox{$\sffsstructtwo\modelfor\standbf{\all}{\spreg\land\sovu}$} (as \mbox{$\sffsstructtwo,\pr\modelfor\spreg\land\sovu$} for all \mbox{$\pr\in\Precstwo$}) and \mbox{$\sffsstructtwo,\pr_5\modelfor\standbx{\smed}\shorm$} (since \mbox{$\sffsstructtwo,\pr\modelfor\shorm$} for all \mbox{$\pr\in\stanitwo(\smedtext)$}), while \mbox{$\sffsstructtwo,\pr_1\not\modelfor\standbx{\smed}\shorm$} (as \mbox{$\sffsstructtwo,\pr_3\not\modelfor\shorm$} with \mbox{$\pr_3\in\stanotwo(\smedtext)$}), whence $\sffsstructtwo,\pr_1\modelfor\standdx{\smed}\neg\shorm$.
        Intuitively, from the medical standpoint $\smedtext$, $\neg\shorm$ is conceivable at $\pr_1$, whereas $\shorm$ is unequivocal at $\pr_5$.
    }
    \label{fig:structure}
\end{figure}
For illustration, \Cref{fig:structure} graphically depicts an \sff standpoint structure for \Cref{example} along with some satisfied formulas.

The sharpening statements of a theory $\thr$ form a hierarchy of standpoints (with $\all$ at the top) that we sometimes need.
\begin{definition}
    \label{def:standpoint-hierarchy}
    Let \mbox{$\thr\subseteq\langss$} be a theory over standpoint names $\Stands$.
    For $\sts,\stu\in\Stands$ we say that \define{$\sts$ sharpens $\stu$} and write $\thr\provess\sts\sharpens\stu$, which is defined by induction as follows\/:
    \begin{itemize}
        \item $\thr\provess\sts\sharpens\stu$ if $\sts\sharpens\stu\in\thr$
              or $\stu=\all$
              or $\sts=\stu$; \hfill (base cases)
        \item $\thr\provess\sts\sharpens\stu$ if there is some $\stt\in\Stands$ such that $\sts\sharpens\stt\in\thr$ and $\thr\provess\stt\sharpens\stu$. \hfill (inductive case)
    \end{itemize}
\end{definition}
It is clear that for finite \mbox{$\thr\subseteq\langss$} and \mbox{$\sts,\stu\in\Stands$},
the question whether \mbox{$\thr\provess\sts\sharpens\stu$} can be decided in deterministic polynomial time by checking reachability in the directed graph $(\Stands,\set{(\sts,\stu)\guard\sts\sharpens\stu\in\thr}\cup\Stands\times\set{\all})$.
%It is equally clear that we can show by induction that whenever $T\entailss\sts\sharpens\stt$ and $T\entailss\stt\sharpens\stu$, then also $T\entailss\sts\sharpens\stu$.
%
Furthermore, all sharpening statements that we can establish via reachability are correct with respect to the model theory, as this result shows.
\begin{lemma}
    \label{lem:sharpening-correctness}
    For any theory $\thr\subseteq\langss$ and $\sts,\stu\in\Stands$,
    \begin{enumerate}
        \item $\thr\provess\sts\sharpens\stu$ implies $\thr\entailss\sts\sharpens\stu$, and
        \item if $\thr$ is satisfiable, then $\thr\provess\sts\sharpens\stu$ iff $\thr\entailss\sts\sharpens\stu$.
    \end{enumerate}
    \iflong
        \newcommand{\prs}{\pr_{\sts}}
        \newcommand{\prsc}{\prcopy{\prs}}
        \newcommand{\prt}{\pr_{\stt}}
        \newcommand{\prr}{\pr_{\str}}
        \begin{proof}
            \begin{enumerate}
                \item
                      By induction on the definition of $\thr\provess\sts\sharpens\stu$.
                      Assume $\thr\provess\sts\sharpens\stu$ and consider $\sffsstruct=\SFFSstruct$ with $\sffsstruct\modelfor\thr$.
                      In the base cases, the case $\sts\sharpens\stu\in\thr$ is clear;
                      the case $\stu=\all$ follows by definition, and the case $\sts=\stu$ is trivial.
                      In the induction step, assume that $\sts\sharpens\stt\in\thr$ and $\thr\provess\stt\sharpens\stu$.
                      By $\sffsstruct\modelfor\thr$, we obtain $\sffsstruct\modelfor\sts\sharpens\stt$.
                      By the induction hypothesis, $\sffsstruct\modelfor\stt\sharpens\stu$.
                      Thus $\stani(\sts)\subseteq\stani(\stt)\subseteq\stani(\stu)$ and $\stano(\sts)\subseteq\stano(\stt)\subseteq\stano(\stu)$, whence $\sffsstruct\modelfor\sts\sharpens\stu$.
                      Since $\sffsstruct$ was arbitrary, $\thr\entailss\sts\sharpens\stu$.
                \item It remains to show the “if” direction.
                      We show the contrapositive, thus assume $\thr\not\provess\sts\sharpens\stu$.
                      Since $\thr$ is satisfiable, there is $\sffsstruct=\SFFSstruct$ such that $\sffsstruct\modelfor\thr$.
                      We will use $\sffsstruct$ to construct $\sffsstructt$ with $\sffsstructt\modelfor\thr$ and $\sffsstructt\not\modelfor\sts\sharpens\stu$.

                      To this end, consider an arbitrary $\prs\in\stani(\sts)\neq\emptyset$.
                      Create a copy $\prsc\notin\Precs$ and define $\sffsstructt=\SFFSstructt$ as follows:
                      \begin{itemize}
                          \item $\Precs'=\Precs\cup\set{\prsc}$;
                          \item for any $\stt\in\Stands$ we set
                                \[
                                    \stani'(\stt) \eqdef \begin{cases}
                                        \stani(\stt) \cup \set{ \prsc } & \text{if } \thr\provess\sts\sharpens\stt, \\
                                        \stani(\stt)                    & \text{otherwise;}
                                    \end{cases}
                                \]
                          \item $\stano'(\stt)=\stano(\stt)$ for all $\stt\in\Stands$; and
                          \item $\eval'(\pr)=\eval(\pr)$ for all $\pr\in\Precs$, and $\eval'(\prsc)=\eval(\pr_{\sts})$.
                      \end{itemize}
                      By construction, since $\thr\not\provess\sts\sharpens\stu$, we have $\prsc\notin\stani(\stu)$.
                      Thus by $\prsc\in\stani'(\sts)$, $\sffsstructt\not\modelfor\sts\sharpens\stu$.
                      It remains to show that $\sffsstructt\modelfor\thr$.
                      To this end, we first establish a helper claim.
                      \begin{claim}
                          \label{claim:sharpening-correctness:witness}
                          For every $\stt\in\Stands$ with $\prsc\in\stani'(\stt)$, there is a $\prt\in\stani(\stt)$ such that for all $\psi\in\langs$, we have
                          \[
                              \sffsstructt,\prsc\modelfor\psi \iff \sffsstructt,\prt\modelfor\psi
                          \]
                          \begin{claimproof}
                              Recall that $\prsc$ is a copy of $\prs\in\stani(\sts)$ with $\eval'(\prsc)=\eval'(\prs)$.
                              Since $\prsc\in\stani'(\stt)$, then by construction $\thr\provess\sts\sharpens\stt$.
                              Thus from $\sffsstruct\modelfor\thr$ and the first item of the main proof, we get $\sffsstruct\modelfor\sts\sharpens\stt$ and thus $\prs\in\stani(\stt)$.
                              Therefore, we can choose $\prt\eqdef\prs$.
                              It remains to show that for all formulas $\psi\in\langs$, we have
                              \[
                                  \sffsstructt,\prsc\modelfor\psi \iff \sffsstructt,\prt\modelfor\psi
                              \]
                              This is shown by induction on $\psi$ with the base case clear from $\eval'(\prsc)=\eval'(\prs)=\eval'(\prt)$.
                              Of the remaining cases, the only non-trivial one is $\psi=\standbr\xi$.
                              \begin{description}
                                  \item[\normalfont“$\!\implies\!$”:]
                                        Let $\sffsstructt,\prsc\modelfor\standbr\xi$.
                                        Since $\prsc\in\stani'(\sts)\subseteq\stani'(\all)$, we obtain $\sffsstructt,\pr_{\str}\modelfor\xi$ for all $\pr_{\str}\in\stani'(\str)$ and thus, since $\prt\in\stani'(\stt)\subseteq\stani'(\all)$, we get $\sffsstructt,\prt\modelfor\standbr\xi$.
                                  \item[\normalfont“$\!\impliedby\!$”:]
                                        Let $\sffsstructt,\prsc\not\modelfor\standbr\xi$.
                                        Then there is a $\prr\in\stani'(\str)$ with $\sffsstructt,\prr\not\modelfor\xi$.
                                        Thus we get $\sffsstructt,\prt\not\modelfor\standbr\xi$.
                              \end{description}
                          \end{claimproof}
                      \end{claim}
                      This enables us to show the next helper result.
                      \begin{claim}
                          \label{claim:sharpening-correctness:formulas}
                          For any formula $\psi\in\langs$ and $\pr\in\Precs$, we have
                          \[
                              \sffsstruct,\pr\modelfor\psi \iff \sffsstructt,\pr\modelfor\psi
                          \]
                          \begin{claimproof}
                              By induction on the structure of $\psi$.
                              The only non-trivial case is $\psi=\standbt\xi$.
                              \begin{description}
                                  \item[\normalfont“$\!\implies\!$”:]
                                        Let $\sffsstructt,\pr\not\modelfor\standbt\xi$.
                                        \begin{itemize}
                                            \item $\pr\in\stani(\all)$:
                                                  Then there is a $\pr'\in\stani'(\stt)$ such that $\sffsstructt,\pr'\not\modelfor\xi$.
                                                  If $\pr'\in\stani(\stt)$ then the induction hypothesis yields $\sffsstruct,\pr'\not\modelfor\xi$ and $\sffsstruct,\pr\not\modelfor\standbt\xi$.
                                                  Otherwise, $\pr'=\prsc$.
                                                  By \Cref{claim:sharpening-correctness:witness}, there exists a $\prt\in\stani(\stt)$ such that $\sffsstructt,\prt\not\modelfor\xi$.
                                                  The induction hypothesis again yields $\sffsstruct,\prt\not\modelfor\xi$ and thus we obtain $\sffsstruct,\pr\not\modelfor\standbt\xi$.
                                            \item $\pr\in\stano(\all)$:
                                                  Then there is a $\pr'\in\stani'(\stt)\cup\stano'(\stt)$ such that $\sffsstructt,\pr'\not\modelfor\xi$.
                                                  If $\pr'\in\stani'(\stt)$, then we can reason as in the previous case.
                                                  Otherwise, $\pr'\in\stano'(\stt)=\stano(\stt)$ and the induction hypothesis yields $\sffsstruct,\pr'\not\modelfor\xi$, whence $\sffsstruct,\pr\not\modelfor\standbt\xi$.
                                        \end{itemize}
                                  \item[\normalfont“$\!\impliedby\!$”:]
                                        Let $\sffsstructt,\pr\modelfor\standbt\xi$.
                                        \begin{itemize}
                                            \item $\pr\in\stani(\all)$:
                                                  Then for all $\prt\in\stani'(\stt)$, we have $\sffsstructt,\prt\modelfor\xi$.
                                                  In particular, for all $\prt\in\stani(\stt)\subseteq\stani'(\stt)$, we have $\sffsstructt,\prt\modelfor\xi$.
                                                  By the induction hypothesis, we thus have $\sffsstruct,\prt\modelfor\xi$ for all $\prt\in\stani(\stt)$, therefore $\sffsstruct,\pr\modelfor\standbt\xi$.
                                            \item $\pr\in\stano(\all)$:
                                                  Similar to the previous case, as we have $\stani(\stt)\cup\stano(\stt)\subseteq\stani'(\stt)\cup\stano'(\stt)$.
                                        \end{itemize}
                              \end{description}

                              %   We do a case distinction.
                              %   \begin{itemize}
                              %       \item $\thr\provess\sts\sharpens\stt$.
                              %       \item $\thr\not\provess\sts\sharpens\stt$.
                              %   \end{itemize}

                          \end{claimproof}
                      \end{claim}
                      We now use the helper claims to establish $\sffsstructt\modelfor\thr$.
                      Let $\varphi\in\thr$ be arbitrary.
                      \begin{itemize}
                          \item $\varphi\in\langs$ is a formula:
                                From $\sffsstruct\modelfor\thr$ we obtain $\sffsstruct\modelfor\varphi$, that is, $\sffsstruct,\pr\modelfor\varphi$ for all $\pr\in\Precs$.
                                Now consider an arbitrary $\pr'\in\Precs'$.
                                If $\pr'\in\Precs$, then \Cref{claim:sharpening-correctness:formulas} yields $\sffsstructt,\pr'\modelfor\varphi$.
                                Otherwise, $\pr'=\prsc$ and by \Cref{claim:sharpening-correctness:witness} there is a $\prt\in\stani(\all)$ with $\sffsstructt,\pr'\modelfor\varphi\iff\sffsstructt,\prt\modelfor\varphi$.
                                From $\sffsstruct,\prt\modelfor\varphi$ and \Cref{claim:sharpening-correctness:formulas} we thus obtain $\sffsstructt,\prt\modelfor\varphi$ and hence $\sffsstructt,\prsc\modelfor\varphi$.
                                In any case, $\sffsstructt,\pr'\modelfor\varphi$ and thus $\sffsstructt\modelfor\varphi$.
                          \item $\varphi=\str\sharpens\stt$ is a sharpening statement:
                                By $\sffsstruct\modelfor\thr$ we get $\stani(\str)\subseteq\stani(\stt)$ and $\stano'(\str)=\stano(\str)\subseteq\stano(\stt)=\stano'(\stt)$.
                                It remains to show $\stani'(\str)\subseteq\stani'(\stt)$.
                                Consider an arbitrary $\pr'\in\stani'(\str)$.
                                If $\pr'\in\stani(\str)$, then $\pr'\in\stani(\stt)\subseteq\stani'(\stt)$ is immediate.
                                Otherwise, $\pr'=\prsc$ and thus by construction $\thr\provess\sts\sharpens\str$.
                                Now since we have $\str\sharpens\stt\in\thr$, we thus obtain $\thr\provess\sts\sharpens\stt$, whence $\prsc\in\stani'(\stt)$ by construction.
                      \end{itemize}
                      In any case, $\sffsstructt\modelfor\varphi$.
                      Since $\varphi\in\thr$ was arbitrary, we obtain $\sffsstructt\modelfor\thr$.
            \end{enumerate}
        \end{proof}
    \fi
\end{lemma}

\subsection{Non-Monotonic Semantics}

In uni-modal \sff, the non-monotonic semantics seeks to minimise the amount of \emph{knowledge} contained in a structure $\sfstruct$.
Similarly, in the non-monotonic semantics of \sff standpoint logic we seek to minimise the amount of \emph{determination} contained in the structure of each standpoint.
Intuitively, this leads to standpoints making semantic commitments only insofar it is absolutely necessitated by a given theory.
To this end, we firstly introduce a preference ordering on structures.

\begin{definition}
    \label{def:sffs:preference-ordering}
    Consider the \sff standpoint structures \mbox{$\sffsstructn{1}=\SFFSstructn{1}$} and \mbox{$\sffsstructn{2}=\SFFSstructn{2}$} over standpoint names $\Stands$ and \mbox{$\sts\in\Stands$}.
    We define the following orderings\/:%
    \begin{align*}
        \sffsstructn{1}\sprefeqs\sffsstructn{2} & \iffdef \eval_2(\stani_2(\sts))=\eval_1(\stani_1(\sts)) \tand                                   \\
                                                & \phiffdef \eval_2(\stano_2(\sts))\subseteq\eval_1(\stani_1(\sts)\cup\stano_1(\sts))             \\
        \sffsstructn{1}\sprefeq\sffsstructn{2}  & \iffdef \sffsstructn{1}\sprefeqs\sffsstructn{2} \,\text{ for all }\, \sts\in\Stands             \\
        \sffsstructn{1}\spref\sffsstructn{2}    & \iffdef \sffsstructn{1}\sprefeq\sffsstructn{2} \tand \sffsstructn{2}\not\sprefeq\sffsstructn{1} \\
        \sffsstructn{1}\sprefiso\sffsstructn{2} & \iffdef \sffsstructn{1}\sprefeq\sffsstructn{2} \tand \sffsstructn{2}\sprefeq\sffsstructn{1}
    \end{align*}
\end{definition}
Intuitively, \mbox{$\sffsstructn{1}\sprefeq\sffsstructn{2}$} expresses that for all standpoints, $\sffsstructn{2}$ is at least as determined as $\sffsstructn{1}$, with respect to the structures' semantic commitments.
In our non-monotonic semantics, we are interested in structures where such commitments are minimal (subject to still satisfying a given theory).
We call an \sff standpoint structure an \define{S5 standpoint structure} whenever for all \mbox{$\sts\in\Stands$} we have \mbox{$\stano(\sts)=\emptyset$}.

\begin{definition}
    \label{def:minimal-model}
    Let \mbox{$\thr\subseteq\langss$}.
    An S5 standpoint structure $\sffsstruct$ is a \define{minimal model of $\thr$} iff
    (1) \mbox{$\sffsstruct\modelfor\thr$} and
    (2) for all \sff standpoint structures \mbox{$\sffsstructt\sprefeq\sffsstruct$} with \mbox{$\sffsstructt\modelfor\thr$}, we find \mbox{$\sffsstructt\sprefiso\sffsstruct$}.
\end{definition}
As usual, an expression $\varphi\in\langss$ is \define{credulously} (\define{sceptically}) entailed by a theory $\thr\subseteq\langss$, denoted $\thr\dentailscred\varphi$ ($\thr\dentailsscep\varphi$), iff
$\sffsstruct\modelfor\varphi$ for some (all) minimal model(s) $\sffsstruct$ of $\thr$.
E.g.\ a minimal model $\sffsstructtwomin=\SFFSstructtwomin$ of theory $\thrtwo$ from \Cref{example} is obtained from $\sffsstructtwo$ (see~\Cref{fig:structure}) by moving all outer precisifications into their respective inner sets (yielding e.g.\
\mbox{$\stanitwomin(\sdtwotext) = \stanotwo(\sdtwotext)\cup\stanitwo(\sdtwotext)$})
thus emptying the outer sets.
%, i.e., setting $\stano'(\sts) = \emptyset$ for all $\sts \in \Stands$.
% $\stanitwomin(\sdtwo)=\set{\pr_3, \pr_4, \pr_7}$) 

We observe that S4F standpoint logic generalises both propositional standpoint logic~\citep{GomezAlvarezR21} (via S5 standpoint structures; only atomic sharpening statements) as well as unimodal \sff~\citep{SchwarzT94} (via $\Stands=\set{\all}$ and replacing every occurrence of $\know$ in a theory $\thr$ by $\standball$, denoted $\thr[\know/\standball]$).

\begin{proposition}
    \label{thm:generalisation}
    \begin{enumerate}
        \item\label{itm:generalisation:sl} For any theory $\thr$ of propositional standpoint logic such that sharpening statements occur in $\thr$ as atoms only, there is a bijection between the set of standpoint structures $\sstruct$ that satisfy $\thr$ and the set of S5 standpoint structures $\sffsstruct$ that satisfy $\thr$.
        \item\label{itm:generalisation:sff} For any theory $\thr\subseteq\langk$ of \sff, there is a bijection between the set of \sff structures $\sffstruct$ that satisfy $\thr$ and the set of \sff standpoint structures $\sffsstruct$ that satisfy $\thr[\know/\standball]$.
    \end{enumerate}
    \iflong
        \begin{proof}
            \begin{enumerate}
                \item Given a standpoint structure \mbox{$\sstruct=\Sstruct$}, we can define an \sff standpoint structure \mbox{$\sffsstruct=\SFFSstruct$} by just setting \mbox{$\stano(\sts)\eqdef\emptyset$} for all \mbox{$\sts\in\Stands$}.
                      Conversely, an S5 standpoint structure \mbox{$\sffsstruct=\SFFSstruct$} has \mbox{$\stano(\sts)=\emptyset$} and thus \mbox{$\sstruct=\Sstruct$} is an equivalent structure.
                \item Given an \sff structure \mbox{$\sffstruct=\SFFstruct$}, we can define an \sff standpoint structure \mbox{$\sffsstruct=\SFFSstruct$} by setting \mbox{$\stani(\all)\eqdef\ws$}, \mbox{$\stano(\all)\eqdef\wf$}, and \mbox{$\eval\eqdef\wm$}.
                      Conversely, an \sff standpoint structure \mbox{$\sffsstruct=\SFFSstruct$} for \mbox{$\Stands=\set{\all}$} determines an \sff structure \mbox{$\sffstruct=\SFFstruct$} via \mbox{$\wf\eqdef\stano(\all)$}, \mbox{$\ws\eqdef\stani(\all)$}, and \mbox{$\wm\eqdef\eval$}.
            \end{enumerate}
        \end{proof}
    \fi
\end{proposition}

\iflong
    % expand -- \input{sections/technical-lemmas}
    %%%%%%%%%%%%%%%%%%%%%%%%%%%%%%%%%%%%%%%%%%%%%%%
    In the remainder, we establish several key technical properties of the language.
    They will be instrumental in the results of later sections.

    The first result concerns \sff standpoint structures that are comparable in the determination ordering $\sprefeq$, especially the interplay of what is unequivocal in the less determined structure and how that can transfer to the more determined structure.
    \begin{lemma}
        \label{lemma-knowledge-passing}
        Consider the two \sff standpoint structures \mbox{$\sffsstructn{1}=\SFFSstructn{1}$} and \mbox{$\sffsstructn{2}=\SFFSstructn{2}$} over $\Atoms$ and $\Stands$ such that \mbox{$\sffsstructn{1}\sprefeq\sffsstructn{2}$}.
        For all formulas $\phi\in\langs$ and $\sts\in\Stands$\/:
        \begin{enumerate}[label=(\roman*)]
            \item\label{lemma2:item-1} for all $\pr_1\in\stani_1(\sts)$ and $\pr_2\in\stani_2(\sts)$ with $\eval_1(\pr_1)=\eval_2(\pr_2)$:
                  \[
                      \sffsstructn{1},\pr_1\modelfor\phi\iff\sffsstructn{2},\pr_2\modelfor\phi
                  \]
            \item\label{lemma2:item-2} for all $\pr_1\in\stano_1(\sts)$ and $\pr_2\in\stani_2(\sts)$:
                  \[
                      \sffsstructn{1},\pr_1\modelfor\standbu\phi \implies \sffsstructn{2},\pr_2\modelfor\standbu\phi
                  \]
            \item\label{lemma2:item-3} if $\sffsstructn{2}$ is an S5 standpoint structure, then
                  \[
                      \sffsstructn{1}\modelfor\standbu\phi \implies \sffsstructn{2}\modelfor\standbu\phi
                  \]
            \item\label{lemma2:item-4} if $\sffsstructn{2}$ is an S5 standpoint structure, then
                  \[
                      \sffsstructn{2}\modelfor\neg\standbu\phi \implies \sffsstructn{1}\modelfor\neg\standbu\phi
                  \]
        \end{enumerate}
        \begin{proof}
            \begin{enumerate}[label=(\roman*)]
                \item
                      By induction on the structure of $\phi$.
                      The only nontrivial case is $\phi=\standbu\psi$.
                      \begin{itemize}
                          \item “$\!\implies\!$”:
                                Let $\sffsstructn{1},\pr_1\modelfor\standbu\psi$ and consider an arbitrary $\pr_2'\in\stani_2(\stu)$.
                                By $\sffsstructn{1}\sprefeqs[\stu]\sffsstructn{2}$ we obtain $\eval_1(\stani_1(\stu))=\eval_2(\stani_2(\stu))$, whence there exists a $\pr_1'\in\stani_1(\stu)$ such that $\eval_1(\pr_1')=\eval_2(\pr_2')$.
                                By $\sffsstructn{1},\pr_1\modelfor\standbu\psi$ and $\pr_1'\in\stani_1(\stu)$, we obtain $\sffsstructn{1},\pr_1'\modelfor\psi$.
                                By the induction hypothesis, we get $\sffsstructn{2},\pr_2'\modelfor\psi$.
                                Since $\pr_2'$ was arbitrarily chosen, we get $\sffsstructn{2},\pr_2\modelfor\standbu\psi$.
                          \item “$\!\impliedby\!$”:
                                Assume $\sffsstructn{1},\pr_1\not\modelfor\standbu\psi$, which by the semantics yields $\sffsstructn{1},\pr'_1\not\modelfor\psi$ for some $\pr'_1\in\stani_1(\stu)$.
                                Since $\sffsstructn{1}\sprefeq\sffsstructn{2}$, we get $\eval_1(\stani_1(\stu))=\eval_2(\stani_2(\stu))$ and there must exist a $\pr'_2\in\stani_2(\stu)$ such that $\eval_1(\pr'_1)=\eval_2(\pr'_2)$.
                                By the induction hypothesis, then $\sffsstructn{2},\pr'_2\not\modelfor\phi$ and consequently $\sffsstructn{2},\pr_2\not\modelfor\standbu\phi$.
                      \end{itemize}
                \item
                      Let $\pr_1\in\stano_1(\sts)$,  $\pr_2\in\stani_2(\sts)$ and assume $\sffsstructn{1},\pr_1\modelfor\standbu \phi$.
                      Then for all $\pr'_1\in\stani_1(\stu)\cup\stano_1(\stu)$, $\sffsstructn{1},\pr_1'\modelfor\phi$.
                      In particular, for all $\pr_1'\in\stani_1(\stu)$, we have $\sffsstructn{1},\pr_1'\modelfor\phi$.
                      Consider an arbitrary $\pr^\dagger_2\in\stani_2(\stu)$.
                      Since $\eval_1(\stani_1(\stu))=\eval_2(\stani_2(\stu))$, there must be a $\pr^\dagger_1\in\stani_1(\stu)$ with $\eval_1(\pr^\dagger_1)=\eval_2(\pr^\dagger_2)$.
                      By $\sffsstructn{1},\pr^\dagger_1\modelfor\phi$ and \Cref{lemma2:item-1}, we get $\sffsstructn{2},\pr^\dagger_2\modelfor\phi$.
                      Since the choice of $\pr^\dagger_2\in\stani_2(\stu)$ was arbitrary, we obtain $\sffsstructn{2},\pr_2\modelfor\standbu\phi$.
                \item
                      Let $\sffsstructn{1}\modelfor\standbu\phi$.
                      If $\stano_1(\all)=\emptyset$, then the claim follows directly from \Cref{lemma2:item-1};
                      otherwise, for $\pr_1\in\stano_1(\all)$, we have $\sffsstructn{1},\pr_1\modelfor\standbu\phi$ and by \Cref{lemma2:item-2} we obtain $\sffsstructn{2},\pr_2\modelfor\standbu\phi$ for all $\pr_2\in\stani_2(\all)=\Precs_2$, thus $\sffsstructn{2}\modelfor\standbu\phi$.
                \item
                      Let $\sffsstructn{2}\modelfor\neg\standbu\phi$.
                      Then there is a $\pr_2^{\stu}\in\stani_2(\stu)$ with $\sffsstructn{2},\pr_2^{\stu}\not\modelfor\phi$.
                      By $\eval_1(\stani_1(\stu))=\eval_2(\stani_2(\stu))$, there is a $\pr_1^{\stu}\in\stani_1(\stu)$ with $\eval_1(\pr_1^{\stu})=\eval_2(\pr_2^{\stu})$.
                      By \Cref{lemma2:item-1}, we obtain $\sffsstructn{1},\pr_1^{\stu}\not\modelfor\phi$.
                      Thus for any $\pr_1\in\Precs_1$, we have $\sffsstructn{1},\pr_1\not\modelfor\standbu\phi$, that is, $\sffsstructn{1},\pr_1\modelfor\neg\standbu\phi$.
                      Thus $\sffsstructn{1}\modelfor\neg\standbu\phi$.
            \end{enumerate}
        \end{proof}
    \end{lemma}

    % \hs{
    %     Superseded by \Cref{lem:uniqd} below?

    %     \begin{definition}[Unique formula for a prop. evaluation $E$]\label{def:uniqd}
    %         Let $\alphabet$ be a set of propositional atoms and $E\subseteq \alphabet$ a propositional evaluation over $\alphabet$. We denote the unique formula for $E$ as $\uniqd$ and define as:
    %         \[
    %             \uniqd\eqdef \neg(\bigwedge_{p\in E} p \land \bigwedge_{p\in \alphabet\setminus E} \neg p)
    %         \]
    %     \end{definition}
    %     An interesting property of $\uniqd$ for $E\subseteq \alphabet$ is the following: Let $\sffsstruct=\SFFSstruct$ be a structure over $\alphabet$ and $\Stands$ and $\sts\in\Stands$ be a standpoint name such that $E\notin \eval(\stani(\sts))$. Then for $\pr\in\stani(\sts)$, $\sffsstruct,\pr\modelfor\standbs\uniqd$.
    %     Conversely, let $E\notin \eval(\stano(\sts)\cup\stani(\sts))$. Then for $\pr\in\stano(\sts)\cup\stani(\sts)$, we have $\sffsstruct,\pr\modelfor\standbs\uniqd$ and consequently $\sffsstruct\modelfor\standbs\uniqd$.

    % }

    We next introduce a way to identify the presence of certain valuations the inner/outer precisifications of a standpoint using propositional formulas.
    \begin{lemma}
        \label{lem:uniqd}
        Let $\sffsstruct=\SFFSstruct$ be an \sff standpoint structure over a finite set $\Atoms$ of atoms and standpoint names $\Stands$, let $E\subseteq\Atoms$ be a propositional valuation, and define
        \[
            \uniq[E] \eqdef \neg\left(\bigland_{p\in E}p\land\bigland_{p\in\Atoms\setminus E}\neg p\right)
        \]
        Then for any $\sts\in\Stands$ and $\pr\in\Precs$ we have:
        \begin{enumerate}
            \item if $\pr\in\stani(\all)$ then
                  \(
                  \sffsstruct,\pr\modelfor\standbs\uniq \tiff E\notin\eval(\stani(\sts))
                  \);
            \item if $\pr\in\stano(\all)$ then
                  \(
                  \sffsstruct,\pr\modelfor\standbs\uniq \tiff E\notin\eval(\stani(\sts)\cup\stano(\sts))
                  \);
            \item \(
                  \sffsstruct\modelfor\standbs\uniq \tiff E\notin\eval(\stani(\sts)\cup\stano(\sts))
                  \).
        \end{enumerate}
        \begin{proof}
            It is clear that for any propositional valuation $F\subseteq\Atoms$, we have $F\modelfor\uniq$ iff $F\neq E$.
            \begin{enumerate}
                \item Let $\pr\in\stani(\all)$.
                      \begin{align*}
                          \sffsstruct,\pr\modelfor\standbs\uniq
                           & \iff \forall\pr\in\stani(\sts): \sffsstruct,\pr\modelfor\uniq \\
                           & \iff \forall\pr\in\stani(\sts): \eval(\pr)\neq E              \\
                           & \iff E \notin \eval(\stani(\sts))
                      \end{align*}

                \item Let $\pr\in\stano(\all)$.
                      \begin{align*}
                          \sffsstruct,\pr\modelfor\standbs\uniq
                           & \iff \forall\pr\in\stani(\sts)\cup\stano(\sts): \sffsstruct,\pr\modelfor\uniq \\
                           & \iff \forall\pr\in\stani(\sts)\cup\stano(\sts): \eval(\pr)\neq E              \\
                           & \iff E \notin \eval(\stani(\sts)\cup\stano(\sts))
                      \end{align*}
                \item
                      \begin{align*}
                          \sffsstruct\modelfor\standbs\uniq
                           & \iff \forall\pr\in\stani(\all)\cup\stano(\all): \sffsstruct,\pr\modelfor\standbs\uniq \\
                           & \iff \forall\pr\in\stani(\sts)\cup\stano(\sts): \sffsstruct,\pr\modelfor\uniq         \\
                           & \iff \forall\pr\in\stani(\sts)\cup\stano(\sts): \eval(\pr)\neq E                      \\
                           & \iff E \notin \eval(\stani(\sts)\cup\stano(\sts))
                      \end{align*}
            \end{enumerate}
        \end{proof}
    \end{lemma}

    For S5 standpoint structures, what is undetermined in them serves to exactly syntactically characterise the upper bound with respect to the determination preference ordering.
    \begin{lemma}
        \label{lem:order-unknown}
        Let $\sffsstruct$ be an S5 standpoint structure and define
        \[
            \unknown \eqdef \set{ \neg\standbs\phi \guard \standbs\phi\in\langs \text{ and } \sffsstruct\not\modelfor\standbs\phi }
        \]
        Then it holds for all \sff standpoint structures $\sffsstructt$ that
        \[
            \sffsstructt\sprefeq\sffsstruct \iff \sffsstructt\modelfor\unknown
        \]
        \begin{proof}
            \begin{description}
                \item[\normalfont“$\!\implies\!$”:]
                      Let $\sffsstructt\sprefeq\sffsstruct$ and consider $\neg\standbs\phi\in\unknown$.
                      We have $\sffsstruct\not\modelfor\standbs\phi$ by definition, that is, there is some $\pr\in\Precs=\stani(\all)$ with $\sffsstruct,\pr\not\modelfor\standbs\phi$.
                      Since $\sffsstructt\sprefeq\sffsstruct$, there exists a $\pr'\in\stani'(\all)$ with $\eval(\pr)=\eval'(\pr')$.
                      \Cref{lemma2:item-1} of \Cref{lemma-knowledge-passing} then yields $\sffsstructt,\pr'\not\modelfor\standbs\phi$, that is, there is a $\pr_{\sts}'\in\stani'(\sts)$ such that $\sffsstructt,\pr_{\sts}'\not\modelfor\phi$.
                      Thus for any $\pr''\in\stani'(\all)\cup\stano'(\all)$, we have $\sffsstructt,\pr''\not\modelfor\standbs\phi$, that is, $\sffsstructt,\pr''\modelfor\neg\standbs\phi$.
                      Thus $\sffsstructt\modelfor\neg\standbs\phi$.
                      Since $\neg\standbs\phi$ was arbitrary, we get $\sffsstructt\modelfor\unknown$.
                \item[\normalfont“$\!\impliedby\!$”:]
                      Let \mbox{$\sffsstructt\modelfor\unknown$}.
                      We have to show
                      \mbox{$\eval(\stani(\sts))=\eval'(\stani'(\sts))$}
                      and
                      \mbox{$\eval(\stano(\sts))\subseteq\eval'(\stani'(\sts)\cup\stano'(\sts))$}
                      where by \mbox{$\stano(\sts)=\emptyset$} it suffices to show \mbox{$\eval(\stani(\sts))=\eval'(\stani'(\sts))$}.
                      \begin{itemize}
                          \item “$\subseteq$”:
                                Let $E\in\eval(\stani(\sts))$.
                                Then by \Cref{lem:uniqd} we obtain $\sffsstruct\not\modelfor\standbs\uniq$.
                                Thus $\neg\standbs\uniq\in\unknown$ and $\sffsstructt\modelfor\neg\standbs\uniq$.
                                We obtain that for any $\pr\in\stani'(\sts)\cup\stano'(\sts)$ we have $\sffsstruct,\pr\modelfor\neg\standbs\uniq$, that is, $\sffsstruct,\pr\not\modelfor\standbs\uniq$.
                                In particular, for any $\pr'\in\stani'(\sts)$ we have $\sffsstruct,\pr'\not\modelfor\standbs\uniq$ and by \Cref{lem:uniqd} we get $E\in\eval'(\stani'(\sts))$.
                          \item “$\supseteq$”:
                                Let $E\in\eval'(\stani'(\sts))\subseteq\eval'(\stani'(\sts)\cup\stano'(\sts))$.
                                Then by \Cref{lem:uniqd} we get that for all $\pr'\in\stani'(\all)\cup\stano'(\all)=\Precs'$, we have $\sffsstructt,\pr'\not\modelfor\standbs\uniq$, that is, $\sffsstructt,\pr'\modelfor\neg\standbs\uniq$.
                                Therefore $\sffsstructt\modelfor\neg\standbs\uniq$, and via axiom (\axf) of \sff we obtain $\sffsstructt\modelfor\standbs\neg\standbs\uniq$.
                                Thus $\sffsstructt\not\modelfor\neg\standbs\neg\standbs\uniq$, and $\neg\standbs\neg\standbs\uniq\notin\unknown$.
                                By definition of $\unknown$, we hence get $\sffsstruct\modelfor\standbs\neg\standbs\uniq$, that is, $\sffsstruct\not\modelfor\standbs\uniq$ and thus by \Cref{lem:uniqd} obtain $E\in\eval(\stani(\sts)\cup\stano(\sts))=\eval(\stani(\sts))$.
                      \end{itemize}
            \end{description}
        \end{proof}
    \end{lemma}

    The next result is important for the characterisation of expansions;
    it states that an S5 standpoint structure and an \sff standpoint structure that are mutually preferred (or equal) in the determination ordering are actually isomorphic in the sense that they satisfy exactly the same formulas.

    \begin{lemma}
        \label{lem:isomorphic-s5-s4f-equivalent}
        Consider an S5 standpoint structure $\sffsstruct$ and an \sff standpoint structure $\sffsstructt$ such that $\sffsstruct\sprefeq\sffsstructt$.
        Then for all formulas $\varphi\in\langs$, we have
        \iflong
            \[
                \sffsstruct\modelfor\varphi\iff\sffsstructt\modelfor\varphi
            \]
        \else
            $\sffsstruct\modelfor\varphi\iff\sffsstructt\modelfor\varphi$.
        \fi
        \begin{proof}
            Let $\sffsstruct=\SFFSstruct$ and $\sffsstructt=\SFFSstructt$.
            We will first show the following helper claim:
            \begin{claim}
                \label{claim:isomorphic-s5-s4f-equivalent}
                For all $\pr\in\Precs$ and $\pr'\in\Precs'$ with $\eval(\pr)=\eval'(\pr')$, we have that for all formulas $\varphi\in\langs$,
                \[
                    \sffsstruct,\pr\modelfor\varphi \iff \sffsstructt,\pr'\modelfor\varphi
                \]
                \begin{claimproof}
                    By structural induction on $\varphi$.
                    The only interesting case is $\varphi=\standbs\psi$.
                    \begin{description}
                        \item[\normalfont“$\!\implies\!$”:]
                              Let $\sffsstruct,\pr\modelfor\standbs\psi$.
                              Then for all $\pr^\sts\in\stani(\sts)$, we have $\sffsstruct,\pr^\sts\modelfor\psi$.
                              Now let $\pr_\sts'\in\stani'(\sts)\cup\stano'(\sts)$ be arbitrary.
                              By $\sffsstruct\sprefeq\sffsstructt$ we get $\eval'(\stani'(\sts)\cup\stano'(\sts))\subseteq\eval(\stani(\sts))$, and there exists a $\pr_\sts\in\stani(\sts)$ such that $\eval'(\pr_\sts')=\eval(\pr_\sts)$.
                              By assumption, we obtain $\sffsstruct,\pr_\sts\modelfor\psi$.
                              By the induction hypothesis, we thus obtain $\sffsstructt,\pr_\sts'\modelfor\psi$.
                              Since $\pr_\sts'$ was arbitrarily chosen, we get $\sffsstructt,\pr'\modelfor\standbs\psi$.
                        \item[\normalfont“$\!\impliedby\!$”:]
                              Let $\sffsstructt,\pr'\modelfor\standbs\psi$.
                              Then in any case of $\pr'\in\stani'(\sts)\cup\stano'(\sts)$, we get that for all $\pr_\sts''\in\stani'(\sts)$, we have $\sffsstructt,\pr_\sts''\modelfor\psi$.
                              Now let $\pr_\sts\in\stani(\sts)$ be arbitrary.
                              By $\sffsstruct\sprefeq\sffsstructt$, we obtain $\eval'(\stani'(\sts))=\eval(\stani(\sts))$ and there is a $\pr_\sts'\in\stani'(\sts)$ such that $\eval(\pr_\sts)=\eval'(\pr_\sts')$.
                              Thus $\sffsstructt,\pr_\sts'\modelfor\psi$ and by the induction hypothesis, $\sffsstruct,\pr_\sts\modelfor\psi$.
                              Since $\pr_\sts$ was arbitrarily chosen, we obtain $\sffsstruct,\pr\modelfor\standbs\psi$.
                    \end{description}
                \end{claimproof}
            \end{claim}
            Now for the main proof.
            Let $\varphi\in\langs$ be arbitrary.
            \begin{description}
                \item[\normalfont“$\!\implies\!$”:]
                      Let $\sffsstruct\modelfor\varphi$ and consider $\pr'\in\Precs'$.
                      By $\sffsstruct\sprefeq\sffsstructt$, there exists a $\pr\in\stani(\all)=\Precs$ with $\eval'(\pr')=\eval(\pr)$.
                      Thus $\sffsstruct,\pr\modelfor\varphi$ and by \Cref{claim:isomorphic-s5-s4f-equivalent}, we obtain $\sffsstructt,\pr'\modelfor\varphi$.
                      Since $\pr'$ was arbitrarily chosen, we get $\sffsstructt\modelfor\varphi$.
                \item[\normalfont“$\!\impliedby\!$”:]
                      Let $\sffsstructt\modelfor\varphi$ and consider $\pr\in\Precs=\stani(\all)$.
                      By $\sffsstruct\sprefeq\sffsstructt$, there exists a $\pr'\in\stani'(\all)$ with $\eval(\pr)=\eval'(\pr')$.
                      Thus $\sffsstructt,\pr'\modelfor\varphi$ and by \Cref{claim:isomorphic-s5-s4f-equivalent}, we get $\sffsstruct,\pr\modelfor\varphi$.
                      Since $\pr$ was arbitrarily chosen, we get $\sffsstructt\modelfor\varphi$.
            \end{description}
        \end{proof}
    \end{lemma}

    The statement of \Cref{lem:isomorphic-s5-s4f-equivalent} does not hold any more if we drop the requirement that (at least) one of the structures is S5.
    \begin{example}
        \label{exm:counter:isomorphic-s4f-equivalent}
        % Consider structure $\sffsstructn{1}=\SFFSstructn{1}$ with
        % \begin{itemize}
        %     \item $\Precs_1=\set{\pr_{\emptyset},\pr_{\set{p}},\pr_{\set{p}}'}$,
        %     \item $\stani_1(\sts)=\set{\pr_{\set{p}}}$ and $\stano_1(\sts)=\set{\pr_{\set{p}}'}$,
        %     \item $\stani_1(\stu)=\set{\pr_{\set{p}}}$ and $\stano_1(\stu)=\set{\pr_{\emptyset}}$;
        % \end{itemize}
        % and similarly structure $\sffsstructn{2}=\SFFSstructn{2}$ with
        % \begin{itemize}
        %     \item $\Precs_2=\set{\pr_{\emptyset},\pr_{\set{p}}}$,
        %     \item $\stani_2(\sts)=\set{\pr_{\set{p}}}$ and $\stano_2(\sts)=\emptyset$,
        %     \item $\stani_2(\stu)=\set{\pr_{\set{p}}}$ and $\stano_2(\stu)=\set{\pr_{\emptyset}}$;
        % \end{itemize}
        Consider $\sffsstruct=\tuple{\set{\pr_{\emptyset},\pr_{\set{p}}},\stani,\stano,\eval}$ with
        \begin{itemize}
            \item $\stani(\sts)=\set{\pr_{\set{p}}}$ and $\stano(\sts)=\emptyset$,
            \item $\stani(\stu)=\set{\pr_{\set{p}}}$ and $\stano(\stu)=\set{\pr_{\emptyset}}$;
        \end{itemize}
        where the valuations are denoted via $\eval(\pr_E)=E$.
        It is easy to verify $\sffsstruct,\pr_{\set{p}}\modelfor\standbu p$, thus $\sffsstruct\modelfor\standbs\standbu p$.

        Now consider a copy $\sffsstructt$ of $\sffsstruct$ with the minor modification that $\stano'(\sts)=\{\pr_{\set{p}}'\}$ gets a copy of $\pr_{\set{p}}$.
        We see that $\sffsstruct\sprefeq\sffsstructt$ and $\sffsstructt\sprefeq\sffsstruct$;
        but also, $\sffsstructt,\pr_{\set{p}}'\not\modelfor\standbu p$, thus $\sffsstructt\not\modelfor\standbs\standbu p$.
    \end{example}

\fi

% expand -- \input{sections/expansion}
%%%%%%%%%%%%%%%%%%%%%%%%%%%%%%%%%%%%%%%%%%%%%%
\subsection{Expansions}

Historically, the semantics of non-monotonic reasoning formalisms was not formulated in terms of minimal models from the start.
A more common formulation employed deductively closed sets where non-monotonic inferences have been maximally applied.
For example, \citet{Reiter80} defined the semantics of default logic via so-called \emph{extensions};
\citet{McDermottD80} devised a scheme for obtaining non-monotonic modal logics (of which unimodal \sff is an instance) and defined the semantics based on \emph{expansions};
\citet{Moore85} gave an equally named concept for autoepistemic logic (a logic that can actually be cast into the scheme of \citeauthor{McDermottD80} as non-monotonic modal logic KD45; \citealp{Shvarts90}).
Our definition below is a generalisation of these notions to the multi-modal (standpoint) case.

\begin{definition}
    Let \mbox{$\thr\subseteq\langs$}.
    A set \mbox{$U\subseteq\langs$} is an \define{expansion of $\thr$} iff
    \mbox{$U\!=\!\set{\psi\in\langs \guard \thr \cup \set{\neg \standbs \phi \guard  \standbs \phi \in \langs \!\setminus\! U}\entailss \psi}$.\hspace*{-1ex}}
\end{definition}
Intuitively, a theory $U$ is an expansion of $\thr$ if, by only using the original theory $\thr$ and negative introspection (\wrt $U$), the expansion $U$ can be reproduced exactly.

We can show that expansions are an alternative, but equivalent way to define the non-monotonic semantics of \sff standpoint logic.
The fundamental insight that expansions in the scheme of \citet{McDermottD80} can be equivalently formulated in terms of S5 structures goes back to \citet{Schwarz92};
\citet{MarekST93} later extensively studied such correspondences for various (non-monotonic) modal logics.
Our main result on this topic shows that the correspondence can be lifted from unimodal to \emph{multi-modal} non-monotonic (standpoint) logics.
%There, the \define{theory of a structure} $\sffsstruct$ is the set \mbox{$\thry{\sffsstruct}\eqdef\set{\psi\in\langs\guard\sffsstruct\modelfor\psi}$} (recall that $\langs$ is $\langss$ without sharpening statements).

\begin{theorem}
    Let \mbox{$\thr\subseteq\langs$} be a theory.
    An S5 standpoint structure $\sffsstruct$ is a minimal model of $\thr$
    iff
    the theory of $\sffsstruct$, the set \mbox{$\thry{\sffsstruct}\eqdef\set{\psi\in\langs\guard\sffsstruct\modelfor\psi}$}, is an expansion of $\thr$.
    \begin{longproof}
        \begin{description}
            \item[\normalfont“if”:]
                  Denote \mbox{$\sffsstruct=\SFFSstruct$} and \mbox{$U\eqdef\thry{\sffsstruct}$} and assume that $U$ is an expansion of $\thr$.
                  By $\thr\subseteq U$ and $\sffsstruct\modelfor U$, it is clear that $\sffsstruct\modelfor\thr$.
                  We have to show that $\sffsstruct$ is a minimal model of $\thr$.

                  Consider $\sffsstructt=\SFFSstructt$, such that $\sffsstructt\sprefeq\sffsstruct$.
                  We have to show that $\sffsstruct\sprefeq\sffsstructt$;
                  for this, it remains to show that for all $\sts\in\Stands$, we have
                  \[
                      \eval'(\stano'(\sts))\subseteq\eval(\stani(\sts))
                  \]
                  To this end, let $\sts\in\Stands$ and consider $E\notin\eval(\stani(\sts))$.
                  By \Cref{lem:uniqd}, we obtain $\sffsstruct\modelfor\standbs\uniq$ and $\uniq\in U$.
                  In turn, since $U$ is an expansion of $\thr$ we obtain:
                  \begin{gather}
                      \thr \cup \set{\neg \standbs \phi \guard  \standbs \phi \in \langs \setminus U}\entailss \standbs\uniq \tag{\ddag}
                  \end{gather}

                  Note that $\set{\neg\standbs\phi\guard\standbs\phi\in\langs \setminus U} = \set{\neg\standbs\phi\guard\standbs\phi\in\langs\text{ and }\sffsstruct\not\modelfor\standbs\phi} = \unknown$.
                  By \Cref{lem:order-unknown}, since $\sffsstructt\sprefeq\sffsstruct$, we get
                  \begin{align*}
                      \sffsstructt\modelfor\set{\neg\standbs\phi\guard\standbs\phi\in\langs \setminus U,\sts\in\Stands} \tag{\dag}
                  \end{align*}
                  Then from $\sffsstructt\modelfor\thr$, ($\dag$), and ($\ddag$), we obtain $\sffsstructt\modelfor\standbs\uniq$.
                  Thus by \Cref{lem:uniqd}, $E\notin\eval'(\stani'(\sts)\cup\stano'(\sts))$ and in particular $E\notin\eval'(\stano'(\sts))$.
                  Consequently, $\sffsstruct\sprefeq\sffsstructt$ and $\sffsstruct$ is a minimal model of $\thr$.
            \item[\normalfont“only if”:]
                  Let $\sffsstruct$ be a minimal model of $\thr$.
                  We will show that $U\eqdef\thry{\sffsstruct}$ is an expansion of $\thr$, i.e.\ that
                  \[
                      U=\set{\varphi\in\langs \guard \thr \cup \set{\neg \standbs \phi \guard  \standbs \phi \in \langs \setminus U}\entailss \varphi}
                  \]
                  \begin{description}
                      \item[\normalfont“$\subseteq$”:]
                            We show that for every formula $\psi \in U=\thry{\sffsstruct}$:
                            \[
                                \thr \cup \set{\neg\standbs \phi\guard\standbs \phi\in\langs\setminus \thry{\sffsstruct}}\entailss \psi
                            \]
                            To this end, consider any $\sffsstructt=\SFFSstructt$ such that:
                            \[
                                \sffsstructt\modelfor \thr \text{ and }
                                \sffsstructt\modelfor \set{ \neg\standbs\phi \guard \standbs\phi\in\langs\setminus \thry{\sffsstruct} }
                            \]
                            By \Cref{lem:order-unknown} we know that $\sffsstructt\sprefeq\sffsstruct$.
                            Since $\sffsstruct$ is a minimal model, then $\sffsstructt\sprefiso\sffsstruct$ and
                            since $\sffsstruct$ is an S5 standpoint structure, by \Cref{lem:isomorphic-s5-s4f-equivalent} we know that $\sffsstructn{2}\modelfor\psi\iff\sffsstruct\modelfor\psi$.
                            By $\sffsstruct\modelfor\psi$ we thus get $\sffsstructn{2}\modelfor\psi$.

                      \item[\normalfont“$\supseteq$”:]
                            Let $\varphi\in\langs$ with
                            \[
                                \thr \cup \set{\neg \standbs \phi \guard  \standbs \phi \in \langs \setminus \thry{\sffsstruct}}\entailss \varphi
                            \]
                            We have $\sffsstruct\modelfor\thr$ and for any $\standbs\phi\in\langs\setminus \thry{\sffsstruct}$ we have
                            $\sffsstruct\not\modelfor\standbs\phi$, and as $\sffsstruct$ is an S5 standpoint structure, by modal logic axiom (\axfive) we get $\sffsstruct\modelfor\neg\standbs\phi$.
                            Then we get $\sffsstruct\modelfor\varphi$ and thus $\varphi\in\thry{\sffsstruct}=U$.
                  \end{description}
        \end{description}
    \end{longproof}
\end{theorem}

\section{S4F Standpoint Logic: Complexity Analysis}

% expand -- \input{sections/complexity-monotonic}
%%%%%%%%%%%%%%%%%%%%%%%%%%%%%%%%%%%%%%%%%%%%%%%%%%%%%%%%%%%

\subsection{Complexity of Monotonic S4F Standpoint Logic}
\label{sec:sffs-complexity-mon}

Reasoning within monotonic \sff standpoint logic is interesting in its own right, but also relevant for reasoning with the non-monotonic minimal-model semantics.
We start out with \define{model checking}, that is, given a formula $\psi$ (finite theory $\thr$) and a structure \mbox{$\sffsstruct=\SFFSstruct$} (and possibly a precisification \mbox{$\pr\in\Precs$}), does \mbox{$\sffsstruct,\pr\modelfor\psi$} (respectively \mbox{$\sffsstruct\modelfor\thr$}) hold?

\begin{proposition}
    \label{p:sffs-p-model-checking}
    The model checking problem for S4F standpoint logic (formulas and finite theories) is in \PTime.
    \iflong
        \begin{proof}
            Let $\thr\subseteq\langss$ and $\sffsstruct=\SFFSstruct$ be a structure for the vocabulary of $\thr$.
            We perform the model check as follows:
            For every $\psi\in\subf(\thr)$ with increasing size and every $\pr\in\Precs$, we figure out (bottom-up) whether $\sffsstruct,\pr\modelfor\psi$.
            (In other words, we construct a table with subformulas in the rows and precisifications in the columns and then fill the table cells with yes/no entries.
            Every table entry only depends on entries for strictly simpler subformulas, but potentially on all precisifications.
            The overall size of the table is quadratic, it remains to show that every cell entry can be computed in polynomial time.)
            We show how to do this by induction on the structure of $\psi$:
            \begin{itemize}
                \item $\psi=\sts\sharpens\stu$: This requires at most $\card{\stani(\sts)}\cdot\card{\stani(\stu)}+\card{\stano(\sts)}\cdot\card{\stano(\stu)}$ comparisons of (encoded) precisifications.
                \item $\psi=p\in\Atoms$: Then $\sffsstruct,\pr\modelfor p$ iff $p\in\eval(\pr)$ which requires linear time as we only consult the input.
                \item $\psi=\neg\xi$: We set $\sffsstruct,\pr\modelfor\neg\xi$ iff $\sffsstruct,\pr\not\modelfor\xi$, so we only consult the cell for $(\pr,\xi)$.
                \item $\psi=\psi_1\land\psi_2$: Similar, we only need the two cells for $(\pr,\psi_1)$ and $(\pr,\psi_2)$.
                \item $\psi=\standbs\xi$:
                      By induction hypothesis, we have $\sffsstruct,\pr'\modelfor\xi$ figured out for all $\pr'\in\Precs$.
                      \begin{enumerate}
                          \item $\pr\in\stani(\all)$:
                                If there is a $\pr'\in\stani(\sts)$ such that $\sffsstruct,\pr'\not\modelfor\xi$, we mark $\sffsstruct,\pr\not\modelfor\standbs\xi$;
                                otherwise, we mark $\sffsstruct,\pr\modelfor\standbs\xi$.
                          \item $\pr\in\stano(\all)$:
                                If there is a $\pr'\in\stani(\sts)\cup\stano(\sts)$ such that $\sffsstruct,\pr'\not\modelfor\xi$, we mark $\sffsstruct,\pr\not\modelfor\standbs\xi$;
                                otherwise, we mark $\sffsstruct,\pr\modelfor\standbs\xi$.
                      \end{enumerate}
                      In any case, it suffices to consider the previously constructed modelhood relation for $\xi$ and all precisifications once, that is, consider only the full row for $\xi$ once.
            \end{itemize}
            Finally, the answer to the model checking problem can be read off the row for $\varphi$ (all $\varphi\in\thr$) (and column for $\pr$).
        \end{proof}
    \else
        \begin{proofsketch}
            We can check $\sffsstruct,\pr\modelfor\psi$ bottom-up by considering all subformulas in order of increasing size.
            There are linearly many such checks, and each check in the worst case (of $\standball\xi$) involves all (linearly many) precisifications.
        \end{proofsketch}
    \fi
\end{proposition}

% expand -- \input{sections/small-model-property}
%%%%%%%%%%%%%%%%%%%%%%%%%%%%%%%%%%%%%%%%%%%%%%%%%%%%%
%\subsection{Small Model Property}

As is the case for other standpoint logics~\citep{AlvarezRS22}, one useful aspect of (monotonic) \sff standpoint logic is that satisfiable theories always have small models, where “small” here means linear in the size of the theory.
The \define{size} of a (finite) theory $\thr$ is $\size{\thr}\eqdef\sum_{\varphi\in\thr}\size{\varphi}$ with the size of a formula $\varphi$ defined as the number of its subformulas, $\size{\varphi}\eqdef\card{\subf(\varphi)}$, and the size of a sharpening statement being $\size{\sts\sharpens\stu}\eqdef 2$.

% TODO: clarify the role of \thr having to be finite here
\begin{theorem}
    \label{thm:small-model-property}
    Let $\thr\subseteq\langss$ be finite.
    If $\thr$ is satisfiable, then there exists a model of $\thr$ with at most $\size{\thr}$ precisifications.
    \iflong
        \begin{proof}
            Assume $\sffsstruct\modelfor\thr$ with $\sffsstruct=\SFFSstruct$.
            We will construct a small model $\sffsstructt$ for $\thr$ using the following (standard) idea~\citep{HalpernM92,Schwarz-Truszczynski-1993,AlvarezRS22}:
            For every $\standbs\phi\in\subf(\thr)$ with $\sffsstruct\not\modelfor\standbs\phi$, there exists a precisification $\pr_{\standbs\phi}\in\stani(\sts)\cup\stano(\sts)$ such that $\sffsstruct,\pr_{\standbs\phi}\not\modelfor\phi$;
            for every such formula $\standbs\phi$ we keep one such witness $\pr_{\standbs\phi}$ with a preference for keeping $\pr_{\standbs\phi}\in\stani(\sts)$ if available and only otherwise keeping $\pr_{\standbs\phi}\in\stano(\sts)$.
            (In either case, every subformula $\standbu\phi\in\subf(\thr)$ leads to at most one remaining precisification.)
            We also make sure that \mbox{$\stani'(\stu)\neq\emptyset$} by keeping an arbitrary precisification if necessary.

            We observe that the construction guarantees $\stani'(\all)\subseteq\stani(\all)$ and $\stano'(\all)\subseteq\stano(\all)$.
            It also follows that the resulting structure has at most $\card{\subf(\thr)}$ many precisifications;
            furthermore, $\sffsstructt$ satisfies any \mbox{$\sts\sharpens\stu\in\thr$} by construction.
            It remains to establish the correctness of the construction.
            To this end, we first show a helper result
            \begin{claim}
                \label{claim:small-model-property}
                For each $\pr'\in\Precs'$ and every formula $\psi\in\subf(\thr)$,
                \[
                    \sffsstruct,\pr'\modelfor\psi \iff \sffsstructt,\pr'\modelfor\psi
                \]
                \begin{claimproof}
                    We use induction on the structure of $\psi$.
                    The only interesting proof case is the one for
                    % \begin{itemize}
                    %     \item $\psi=p\in\Atoms$: Clear.
                    %     \item $\psi=\neg\xi$:
                    %     \item $\psi=\psi_1\land\psi_2$:
                    %     \item
                    $\psi=\standbs\xi$:
                    \begin{description}
                        \item[\normalfont “$\!\implies\!$”:]
                              Let $\sffsstruct,\pr'\modelfor\standbs\xi$.
                              We know $\pr'\in\Precs'=\stani'(\all)\cup\stano'(\all)$ and do a case distinction.
                              \begin{enumerate}
                                  \item $\pr'\in\stani'(\all)$.
                                        Then for all $\pr_\sts\in\stani(\sts)$, we have $\sffsstruct,\pr_\sts\modelfor\xi$.
                                        Now let $\pr_\sts'\in\stani'(\sts)$ be arbitrary.
                                        By $\stani'(\sts)\subseteq\stani(\sts)$ we get $\pr_\sts'\in\stani(\sts)$ and thus obtain $\sffsstruct,\pr_\sts'\modelfor\xi$.
                                        By the induction hypothesis, $\sffsstructt,\pr_\sts'\modelfor\xi$.
                                        Since $\pr_\sts'$ was arbitrarily chosen, we thus obtain $\sffsstructt,\pr'\modelfor\standbs\xi$.
                                  \item $\pr'\in\stano'(\all)$.
                                        Then for all $\pr_\sts\in\stani(\sts)\cup\stano(\sts)$, we have $\sffsstruct,\pr_\sts\modelfor\xi$.
                                        Now let $\pr_\sts'\in\stani'(\sts)\cup\stano'(\sts)$ be arbitrary.
                                        From $\stani'(\sts)\subseteq\stani(\sts)$ and $\stano'(\sts)\subseteq\stano(\sts)$ we obtain that $\sffsstruct,\pr_\sts'\modelfor\xi$.
                                        By the induction hypothesis, we obtain $\sffsstructt,\pr_\sts'\modelfor\xi$.
                                        Now $\pr_\sts'$ was arbitrarily chosen, thus by the definition of the semantics, we get $\sffsstructt,\pr'\modelfor\standbs\xi$.
                              \end{enumerate}
                        \item[\normalfont “$\!\impliedby\!$”:]
                              We show the contrapositive.
                              Let $\sffsstruct,\pr'\not\modelfor\standbs\xi$.
                              We do a case distinction.
                              \begin{enumerate}
                                  \item $\pr'\in\stani'(\all)$.
                                        By the semantics and our construction, there is a $\pr_{\standbs\xi}\in\stani'(\sts)\subseteq\stani(\sts)$ with $\sffsstruct,\pr_{\standbs\xi}\not\modelfor\xi$ due to the preference for inner precisifications in our construction.
                                        Therefore the induction hypothesis yields $\sffsstructt,\pr_{\standbs\xi}\not\modelfor\xi$.
                                        Thus $\sffsstructt,\pr'\not\modelfor\standbs\xi$.
                                  \item $\pr'\in\stano'(\all)$.
                                        By semantics and construction, there again exists a precisification $\pr_{\standbs\xi}\in\stani'(\sts)\cup\stano'(\sts)$ such that $\sffsstruct,\pr_{\standbs\xi}\not\modelfor\xi$.
                                        Since $\pr_{\standbs\xi}\in\Precs'$, the induction hypothesis yields $\sffsstructt,\pr_{\standbs\xi}\not\modelfor\xi$.
                                        Thus we get $\sffsstructt,\pr'\not\modelfor\standbs\xi$.
                              \end{enumerate}
                              Thus in any case, $\sffsstructt,\pr'\not\modelfor\standbs\xi$.
                    \end{description}
                    %\end{itemize}
                \end{claimproof}
            \end{claim}
            In particular, for any formula $\varphi\in\thr\subseteq\subf(\thr)$ we obtain\/:
            \begin{align*}
                 & \phiff \sffsstruct\modelfor\varphi                                                                                \\
                 & \iff \forall\pr\in\Precs: \sffsstruct,\pr\modelfor\varphi                                                         \\
                 & \implies \forall\pr'\in\Precs': \sffsstruct,\pr'\modelfor\varphi & \text{ (as $\Precs'\subseteq\Precs$)}          \\
                 & \iff \forall\pr'\in\Precs': \sffsstructt,\pr'\modelfor\varphi    & \text{ (by \Cref{claim:small-model-property})} \\
                 & \iff \sffsstructt\modelfor\varphi
            \end{align*}
            This shows $\sffsstructt\modelfor\thr$ and concludes the overall proof.
        \end{proof}
    \else
        \begin{proofsketch}
            We employ a standard idea~\citep{HalpernM92,Schwarz-Truszczynski-1993,AlvarezRS22}:
            Each \mbox{$\standbu\xi\in\subf(\thr)$} with \mbox{$\sffsstruct\not\modelfor\standbu\xi$} has a witness precisification \mbox{$\pr\in\stani(\stu)\cup\stano(\stu)$} with \mbox{$\sffsstruct,\pr\not\modelfor\xi$}.
            We keep one such witness for each dissatisfied \mbox{$\standbu\xi\in\subf(\thr)$} to obtain a model of size at most $\size{\thr}$.
        \end{proofsketch}
    \fi
\end{theorem}

We next address the \emph{satisfiability} problem of \sff standpoint logic, that is, given a finite theory \mbox{$\thr\subseteq\langss$}, does there exist a structure $\sffsstruct$ that is a model of $\thr$?
To decide it, we use the small model property as expected.
The proposition below generalises the known results on unimodal S4F \citep{Schwarz-Truszczynski-1993} and multi-modal propositional standpoint logic \citep{GomezAlvarezR21}.
\begin{proposition}
    \label{t:sffs-p-satisfiability}
    The satisfiability problem for S4F standpoint logic is \NP-complete.
    \begin{proof}
        \NP-hardness carries over from the proper fragment of propositional logic, so it remains to show membership.
        Given a theory $\thr\subseteq\langs$, we guess an S4F standpoint structure $\sffsstruct$ with at most $\size{\thr}$ many precisifications and then check whether $\sffsstruct\modelfor\varphi$ for all $\varphi\in\thr$.
        The latter check can be done in deterministic polynomial time by \Cref{p:sffs-p-model-checking}.
    \end{proof}
\end{proposition}

Note that theory satisfiability cannot be reduced to formula satisfiability, as formulas lack sharpening statements.

% expand -- \input{sections/characterising-minimal-models}
%%%%%%%%%%%%%%%%%%%%%%%%%%%%%%%%%%%%%%%%%%%%%%%%%%%%%%%%%%%%
\subsection{Characterising Minimal Models}

In this section, we show how the minimal models of a theory $\thr\subseteq\langss$ can be parsimoniously represented, which paves the way for subsequent complexity analyses.
For the purposes of our constructions, we consider (w.l.o.g.) the vocabulary $\Atoms$ of $\thr$ to consist only of those atoms that actually occur in $\thr$.

The main idea of our syntactic characterisation of minimal models follows \citet{Shvarts90} in that we reduce to propositional logic over the extended vocabulary $\Atomsmod\supseteq\Atoms$ where
subformulas of the form $\standbs\xi$ %for some $\sts\in\Stands$
(and sharpening statements) are regarded as propositional atoms.
The major novelty of our construction is the incorporation of sharpening statements via the hierarchy of standpoints.
For brevity, we sometimes %abuse notation for $U\subseteq\langss$ in writing 
denote $\neg\Phi\eqdef\set{\neg\varphi\guard\varphi\in\Phi}$
for a set $\Phi\subseteq\langss$.

\begin{definition}
    \label{def:c1c2c3}
    Let \mbox{$\thr\!\subseteq\!\langss$} be an \sff standpoint theory and denote
    \mbox{$\modalsub[\thr]\eqdef\set{\standbs\phi\guard \standbs\phi\in\subf(\thr) \text{ for some } \sts\in\Stands}$}.
    For a partition $\Partition$ of $\modalsub[\thr]$ we define the following conditions\/: %\pg{Minor remark: Following \cite{Shvarts1990} we could just restrict ourselves to one element of the partition, e.g. only $\Psi$, then $\Phi$ would be easily recoverable as $\Phi\eqdef \thr^{\Box}\setminus\Psi$. It would be easier to have only one symbol in the subscript etc.}
    \begin{description}
        \item[\hypertarget{itm:ca}{\caid}] For every $\sts\in\Stands$,
              the theory
              \mbox{\(
                  \partprop\eqdef \thr \cup \neg\partnk \cup \partk \cup \partk_{\sts}
                  \)}
              is satisfiable in propositional logic, where we define
              \mbox{\(
                  \partk_{\sts}\eqdef \set{\psi\guard\standbu\psi\in\partk\text{ for some } \stu\in\Stands \text{ with } \thr\provess\sts\sharpens\stu}
                  \)}.
        \item[\hypertarget{itm:cb}{\cbid}] For every \mbox{$\sts\in\Stands$} and \mbox{$\standbu\phi\in\partnk$} for some \mbox{$\stu\in\Stands$} with \mbox{$\thr\provess\stu\sharpens\sts$},
              the theory \mbox{\(
                  \partprop \cup \set{ \neg\phi }
                  \)}
              is satisfiable in propositional logic.
        \item[\hypertarget{itm:cc}{\ccid}] For every $\sts\in\Stands$ and $\standbs\psi\in\partk$,
              \mbox{\(
                  \thr \cup \neg\partnk \entailss \standbs\psi
                  \)}.
    \end{description}
\end{definition}

Whenever a partition \mbox{$\partition=\Partition$} satisfies \ca, we can construct an S5 standpoint structure from it.
%(Without satisfaction of \ca, the structure is not well-defined due to empty sets of inner precisifications.)

\begin{definition}
    \label{def:sffsstructpartition}
    Given a partition \mbox{$\partition=\Partition$} satisfying \ca, define S5 standpoint structure \mbox{$\sffsstructpartition=\SFFSstruct$}
    where for \mbox{$\sts\in\Stands$},
    \mbox{$\stani(\sts) \eqdef \set{ F\cap\Atoms \guard F\subseteq\Atomsmod \text{ with } F\modelfor\partprop[\sts] }$}
    and
    \mbox{$\stano(\sts) \eqdef \emptyset$},
    and
    \mbox{$\Precs\eqdef\stani(\all)$} with \mbox{$\eval(\pr)\eqdef\pr$} for all \mbox{$\pr\in\Precs$}.
\end{definition}

Clearly, if $\partition$ satisfies \ca, then $\sffsstructpartition$ is well-defined because \mbox{$\stani(\sts)\neq\emptyset$} for any \mbox{$\sts\in\Stands$}.
Our subsequent main results show that the construction is correct and partitions therefore provide a valid way of characterising (minimal) models for theories.
We start with soundness.

\begin{theorem}
    \label{thm:characterisation-minimal-model:sound}
    Let \mbox{$\thr\subseteq\langss$} and \mbox{$\partition=\Partition$} be a partition of $\modalsub[\thr]$ such that $\partition$ satisfies \ca.
    \begin{enumerate}
        \item\label{itm:s:c12}
              If $\partition$ also satisfies \cb, then $\sffsstructpartition\modelfor\thr$;
        \item\label{itm:s:c123} if $\partition$ also satisfies \cb and \cc, then $\sffsstructpartition$ is a minimal model of $\thr$.
    \end{enumerate}
    \iflong
        \begin{proof}
            \begin{enumerate}
                \item
                      Let $\Partition$ satisfy \ca and \cb.
                      The key to showing the result is the following claim, which establishes that the propositional readings of the elements of $\modalsub$ and their modal readings coincide for the models of $\partprop[\sts]$ for all $\sts\in\Stands$.
                      \begin{claim}
                          \label{claim:a}
                          For all $\varphi\in\subf(\thr)$, for all $\sts\in\Stands$, and for all $F\subseteq\Atomsmod$ with $F\modelfor\partprop[\sts]$, for $E=F\cap\Atoms$ we have
                          \[
                              F\modelfor\varphi \iff \sffsstructpartition,E\modelfor\varphi
                          \]
                          \begin{claimproof}
                              By induction on the structure of $\varphi$.
                              \begin{description}
                                  \item[\normalfont$\varphi=p\in\Atoms$:]
                                        $F\modelfor p$ iff $E\modelfor p$ iff $\sffsstructpartition,E\modelfor p$.
                                  \item[\normalfont$\varphi=\neg\psi$:]
                                        $F\modelfor\neg\psi$
                                        iff $F\not\modelfor\psi$
                                        iff (IH) $\sffsstructpartition,E\not\modelfor\psi$
                                        iff $\sffsstructpartition,E\modelfor\neg\psi$.
                                  \item[\normalfont$\varphi=\psi_1\land\psi_2$:]
                                        $F\modelfor\psi_1\land\psi_2$
                                        iff $F\modelfor\psi_1 \tand F\modelfor\psi_2$
                                        iff (IH) $\sffsstructpartition,E\modelfor\psi_1 \tand \sffsstructpartition,E\modelfor\psi_2$
                                        iff $\sffsstructpartition,E\modelfor\psi_1\land\psi_2$.
                                  \item[\normalfont$\varphi=\standbu\xi$:]
                                        Since $\varphi=\standbu\xi\in\modalsub[\thr]$, we have $\standbu\xi\in\partk$ or $\standbu\xi\in\partnk$.
                                        \begin{description}
                                            \item[\normalfont“$\!\implies\!$”:]
                                                  Let $F\modelfor\standbu\xi$.
                                                  Then $\standbu\xi\in\partk$ and $\xi\in\partprop[\stu]$, whence $\partprop[\stu]\entails\xi$.
                                                  Now let $G\in\stani(\stu)$ be arbitrary.
                                                  By definition, there exists a $H\subseteq\Atomsmod$ such that $H\modelfor\partprop[\stu]$ and $G=H\cap\Atoms$.
                                                  Hence from $\partprop[\stu]\entails\xi$ we get $H\modelfor\xi$.
                                                  By the induction hypothesis, thus $\sffsstructpartition,G\modelfor\xi$.
                                                  Since $G$ was chosen arbitrarily, we obtain $\sffsstructpartition,E\modelfor\standbu\xi$.
                                            \item[\normalfont“$\!\impliedby\!$”:]
                                                  We show the contrapositive.
                                                  Let $F\not\modelfor\standbu\xi$.
                                                  Then $\standbu\xi\in\partnk$ and by \cb, $\partprop[\stu]\cup\set{\neg\xi}$ is satisfiable in propositional logic.
                                                  Let $H\modelfor\partprop[\stu]\cup\set{\neg\xi}$.
                                                  By $H\modelfor\partprop[\stu]$, denoting $G=H\cap\Atoms$ by construction we get that $G\in\stani(\stu)$.
                                                  Furthermore $H\modelfor\neg\xi$ means $H\not\modelfor\xi$, whence by the induction hypothesis we obtain $\sffsstructpartition,G\not\modelfor\xi$.
                                                  By $G\in\stani(\stu)$ we thus get $\sffsstructpartition,E\not\modelfor\standbu\xi$.
                                        \end{description}
                              \end{description}
                          \end{claimproof}
                      \end{claim}
                      Now for showing $\sffsstructpartition\modelfor\thr$.
                      \begin{itemize}
                          \item Let $\varphi\in\thr$ be a formula.
                                We have by construction that $\varphi\in\partprop[\all]$ whence $\partprop[\all]\entails\varphi$.
                                Thus for all $F\subseteq\Atomsmod$ with $F\modelfor\partprop[\all]$ we have $F\modelfor\varphi$;
                                by \Cref{claim:a} we obtain that $\sffsstructpartition,E\modelfor\varphi$ for all $E\in\stani(\all)$.
                                Since $\stani(\all)=\Precs$, we get $\sffsstructpartition\modelfor\varphi$.
                          \item Let $\sts\sharpens\stu\in\thr$.
                                We need to show $\stani(\sts)\subseteq\stani(\stu)$.
                                To this end, we show $\partprop[\stu]\subseteq\partprop[\sts]$, for which it suffices to show that
                                $\partk_{\stu}\subseteq\partk_{\sts}$:
                                If $\psi\in\partk_{\stu}$, then $\standb{\stu'}\psi\in\partk$ for some $\stu'\in\Stands$ with $\thr\provess\stu\sharpens\stu'$.
                                Therefore with $\sts\sharpens\stu\in\thr$ it follows by \Cref{def:standpoint-hierarchy} that $\thr\provess\sts\sharpens\stu'$.
                                Thus $\psi\in\partk_{\sts}$, and since $\psi$ was arbitrary, $\partk_{\stu}\subseteq\partk_{\sts}$ and hence $\partprop[\stu]\subseteq\partprop[\sts]$.
                                From $\partprop[\stu]\subseteq\partprop[\sts]$ it follows for every $F\subseteq\Atomsmod$ that $F\modelfor\partprop[\sts]$ implies $F\modelfor\partprop[\stu]$;
                                this shows that $\stani(\sts)\subseteq\stani(\stu)$ by definition.
                      \end{itemize}
                      Thus we obtain $\sffsstructpartition\modelfor\thr$.
                \item Let $\Partition$ satisfy \ca, \cb, and \cc;
                      furthermore denote $\sffsstructpartition=\SFFSstructt$.
                      By \Cref{itm:s:c12}, $\sffsstructpartition\modelfor\thr$.

                      For showing minimality, assume there is an \sff standpoint structure $\sffsstruct$ such that $\sffsstruct\modelfor\thr$ and $\sffsstruct\sprefeq\sffsstructpartition$.
                      We have to show $\sffsstructpartition\sprefeq\sffsstruct$.

                      From $\sffsstruct\sprefeq\sffsstructpartition$ by \Cref{lem:order-unknown} we obtain:
                      \begin{gather}
                          \sffsstruct\modelfor\set{\neg\standbu\phi\guard \sffsstructpartition\not\modelfor\standbu\phi }
                          \tag{$\dagger$}
                      \end{gather}
                      Now let $\standbu\phi\in\partnk$.
                      Then $\neg\standbu\phi\in\partprop[\all]$ and by definition for every $F\subseteq\Atomsmod$ with $F\modelfor\partprop[\all]$ we have $F\modelfor\neg\standbu\phi$.
                      By \Cref{claim:a}, we obtain that $\sffsstructpartition,F\cap\Atoms\modelfor\neg\standbu\phi$, that is, $\sffsstructpartition\not\modelfor\standbu\phi$.
                      Thus by ($\dagger$) we get $\sffsstruct\modelfor\neg\standbu\phi$.
                      Since $\standbu\phi\in\partnk$ was arbitrarily chosen, we get $\sffsstruct\modelfor\neg\partnk$.

                      Now let $\standbs\psi\in\partk$ be arbitrary.
                      We have $\sffsstruct\modelfor\thr$ and $\sffsstruct\modelfor\neg\partnk$, whence by \cc we obtain $\sffsstruct\modelfor\standbs\psi$.
                      Since $\standbs\psi$ was arbitrary, $\sffsstruct\modelfor\Psi$.

                      In order to establish $\sffsstructpartition\sprefeq\sffsstruct$, it remains to show that for all $\sts\in\Stands$, we have
                      \[
                          \eval(\stano(\sts))\subseteq\eval'(\stani'(\sts))
                      \]
                      Thus consider any $\pr\in\stano(\sts)$ and define
                      \(
                      F\eqdef \eval(\pr)\cup\Psi
                      \).
                      \begin{claim}
                          \label{claim:b}
                          For all $\varphi\in\subf(\thr)$:\;
                          \(
                          \sffsstruct,\pr\modelfor\varphi
                          \iff
                          F\modelfor\varphi
                          \).
                          \begin{claimproof}
                              By structural induction on $\varphi$.
                              The only interesting case is $\varphi=\standbu\xi$:

                              $\sffsstruct,\pr\modelfor\standbu\xi$
                              iff
                              $\sffsstruct\modelfor\standbu\xi$
                              iff
                              $\standbu\xi\in\partk$
                              iff
                              $F\modelfor\standbu\xi$.
                          \end{claimproof}
                      \end{claim}
                      Using \Cref{claim:b}, we thus obtain:
                      \begin{itemize}
                          \item $F\modelfor T$ because $\sffsstruct,\pr\modelfor\thr$ because $\sffsstruct\modelfor\thr$;
                          \item $F\modelfor\neg\partnk$ because $\sffsstruct,\pr\modelfor\neg\partnk$ because $\sffsstruct\modelfor\neg\partnk$;
                          \item $F\modelfor\partk$ because $\sffsstruct,\pr\modelfor\partk$ because $\sffsstruct\modelfor\partk$;
                          \item $F\modelfor\partk_{\sts}$ as for all $\standbu\psi\in\partk$ with $\thr\provess\sts\sharpens\stu$, we have that $\sffsstruct\modelfor\sts\sharpens\stu$ (\Cref{lem:sharpening-correctness})
                                whence $\pr\in\stano(\stu)$ and hence $\sffsstruct,\pr\modelfor\psi$;
                      \end{itemize}
                      Hence we obtain $F\modelfor\partprop[\sts]$, thus by \Cref{def:sffsstructpartition} %definition of $\sffsstructpartition$
                      we get $F\cap\Atoms=\eval(\pr)\in\stani'(\sts)=\eval'(\stani'(\sts))$.
            \end{enumerate}
        \end{proof}
    \else
        \begin{proofsketch}
            The key to the proof is showing that
            for all \mbox{$\sts\in\Stands$}, \mbox{$F\subseteq\Atomsmod$} with \mbox{$F\modelfor\partprop[\sts]$}, and \mbox{$\varphi\in\subf(\thr)$}, we have
            \mbox{$F\modelfor\varphi$} iff \mbox{$\sffsstructpartition,F\cap\Atoms\modelfor\varphi$}.
            With this, we can then show \mbox{$\sffsstructpartition\modelfor\thr$}, where satisfaction of sharpening statements holds by construction.
            For minimality, we assume \mbox{$\sffsstruct\sprefeq\sffsstructpartition$} with \mbox{$\sffsstruct\modelfor\thr$} and employ a helper result establishing that
            \mbox{$\sffsstruct\sprefeq\sffsstructpartition$} implies \mbox{$\sffsstruct\modelfor\set{\neg\standbu\phi\guard \sffsstructpartition\not\modelfor\standbu\phi }$}.
            This serves to show \mbox{$\eval(\stano(\sts))\subseteq\eval'(\stani'(\sts))$}, thus \mbox{$\sffsstructpartition\sprefeq\sffsstruct$}.
        \end{proofsketch}
    \fi
\end{theorem}

This shows soundness of the characterisation;
we can also show completeness, which is the more involved direction.

\begin{theorem}
    \label{thm:characterisation-minimal-model:complete}
    Let \mbox{$\thr\subseteq\langss$} be an S4F standpoint theory and $\sffsstruct$ be an S5 standpoint structure for the vocabulary of $\thr$.
    \begin{enumerate}
        \item\label{itm:c:c12} If \mbox{$\sffsstruct\modelfor\thr$}, then $\modalsub[\thr]$ has a partition $\Partition$ that satisfies \ca and \cb;
        \item\label{itm:c:c123} if $\sffsstruct$ is a minimal model of $\thr$, then $\modalsub[\thr]$ has a partition $\Partition$ that satisfies \ca, \cb, and \cc.
    \end{enumerate}
    \iflong
        \begin{proof}
            \begin{enumerate}
                \item
                      Let \mbox{$\sffsstruct\modelfor\thr$}.
                      We will construct a partition \mbox{$\partition[\thr,\sffsstruct]=\Partition$} and show that it satisfies \ca and \cb.
                      Define
                      \[
                          \partk\eqdef\set{\standbu\phi\in\modalsub[\thr]\guard\sffsstruct\modelfor\standbu\phi}
                          \tand
                          \partnk\eqdef\modalsub[\thr]\setminus\partk
                      \]
                      We first show a helper claim.
                      \begin{claim}
                          \label{claim:c}
                          For any \mbox{$\sts\in\Stands$} and \mbox{$\pr\in\stani(\sts)$} we define the propositional valuation
                          \[
                              F_\pr \eqdef \eval(\pr)\cup\set{ \standbu\phi \guard \sffsstruct\modelfor\standbu\phi }
                          \]
                          for which it holds for all $\varphi\in\subf(\thr)$ that
                          \[
                              F_\pr\modelfor\varphi \iff \sffsstruct,\pr\modelfor\varphi
                          \]
                          \begin{claimproof}
                              We use structural induction on $\varphi$.
                              \begin{description}
                                  \item[\normalfont$\varphi=p\in\Atoms$:]
                                        $F_\pr\modelfor\varphi$ iff $\eval(\pr)\modelfor\varphi$ iff $\sffsstruct,\pr\modelfor\varphi$.
                                  \item[\normalfont$\varphi=\neg\psi$:]
                                        $F_\pr\modelfor\neg\psi$ iff $F_\pr\not\modelfor\psi$ iff (IH) $\sffsstruct,\pr\not\modelfor\psi$ iff $\sffsstruct,\pr\modelfor\neg\psi$.
                                  \item[\normalfont$\varphi=\psi_1\land\psi_2$:]
                                        $F_\pr\modelfor\psi_1\land\psi_2$ iff $F_\pr\modelfor\psi_1$ and $F_\pr\modelfor\psi_2$ iff (IH)
                                        $\sffsstruct,\pr\modelfor\psi_1$ and $\sffsstruct,\pr\modelfor\psi_2$ iff $\sffsstruct,\pr\modelfor\psi_1\land\psi_2$.
                                  \item[\normalfont$\varphi=\standbu\xi$:]
                                        $F_\pr\modelfor\standbu\xi$ iff (by definition) $\sffsstruct\modelfor\standbu\xi$ iff $\sffsstruct,\pr\modelfor\standbu\xi$.
                              \end{description}
                          \end{claimproof}
                      \end{claim}
                      \begin{claim}
                          \label{claim:d}
                          For any $\sts\in\Stands$ and $\pr\in\stani(\sts)$, we have $F_\pr\modelfor\partprop[\sts]$.
                          \begin{claimproof}
                              Construct $F_\pr$ as in \Cref{claim:c}.
                              From assumption $\sffsstruct\modelfor\thr$ by \Cref{claim:c} it follows that $F_\pr\modelfor\thr$;
                              likewise, $F_\pr\modelfor\partk$ as well as $F_\pr\modelfor\neg\standbu\phi$ for every $\standbu\phi\in\partnk$, both by construction.
                              Finally, consider any $\psi\in\subf(\thr)$ such that $\standbu\psi\in\partk$ for some $\stu\in\Stands$ with $\thr\provess\sts\sharpens\stu$.
                              From $\standbu\psi\in\partk$ by construction we get $\sffsstruct\modelfor\standbu\psi$.
                              Additionally $\thr\provess\sts\sharpens\stu$ and $\sffsstruct\modelfor\thr$ entails (via \Cref{lem:sharpening-correctness})
                              that $\sffsstruct\modelfor\sts\sharpens\stu$ and therefore $\stani(\sts)\subseteq\stani(\stu)$.
                              Thus $\pr\in\stani(\sts)$ yields $\pr\in\stani(\stu)$, and from $\sffsstruct\modelfor\standbu\psi$ we thus get $\sffsstruct,\pr\modelfor\psi$.
                              By \Cref{claim:c} above, thus $F_\pr\modelfor\psi$.
                              Since $\psi$ was arbitrary, we get $F_\pr\modelfor\partk$;
                              hence $F_\pr\modelfor\partprop[\sts]$.
                          \end{claimproof}
                      \end{claim}
                      \begin{description}
                          \item[\normalfont${\partition[\thr,\sffsstruct]}$ satisfies \ca:]
                                    Let $\sts\in\Stands$ be arbitrary.
                                    We have to show that $\partprop[\sts]$ is satisfiable in propositional logic.

                                    Since $\stani(\sts)\neq\emptyset$, there is a $\pr\in\stani(\sts)$.
                                    As in \Cref{claim:c} above, we can construct a propositional evaluation $F_\pr$.
                                    By \Cref{claim:d}, we obtain $F_\pr\modelfor\partprop[\sts]$ and $\partprop[\sts]$ is satisfiable.
                          \item[\normalfont${\partition[\thr,\sffsstruct]}$ satisfies \cb:]
                                    Let $\sts\in\Stands$ and $\standbu\phi\in\partnk$ for some $\stu\in\Stands$ with $\thr\provess\stu\sharpens\sts$.
                                    We have to show that the set $\partprop[\sts]\cup\set{\neg\phi}$ is satisfiable in propositional logic.

                                    From $\standbu\phi\in\partnk$ it follows by definition that $\sffsstruct\not\modelfor\standbu\phi$, whence there exists a $\pr\in\stani(\stu)$ such that $\sffsstruct,\pr\not\modelfor\phi$, that is, $\sffsstruct,\pr\modelfor\neg\phi$.
                                    Since by assumption $\thr\provess\stu\sharpens\sts$, it follows with \Cref{lem:sharpening-correctness} that $\thr\entailss\stu\sharpens\sts$; hence
                                $\sffsstruct\modelfor\thr$ yields $\sffsstruct\modelfor\stu\sharpens\sts$, that is, $\stani(\stu)\subseteq\stani(\sts)$ and $\pr\in\stani(\sts)$.
                                    From $\eval(\pr)$, we construct a propositional valuation $F_\pr\subseteq\Atomsmod$ as in \Cref{claim:c} above;
                                    by \Cref{claim:c}, it directly follows that $F_\pr\modelfor\neg\phi$.
                                    From \Cref{claim:d}, as $\pr\in\stani(\sts)$, we additionally get $F_\pr\modelfor\partprop[\sts]$;
                                    overall, this yields $F_\pr\modelfor\partprop[\sts]\cup\set{\neg\phi}$. % witnessing satisfiability.
                      \end{description}
                \item Assume that $\sffsstruct$ is a minimal model of $\thr$.
                      We construct the partition $\partition[\thr,\sffsstruct]$ as above, whence by \Cref{itm:c:c12} it satisfies \ca and \cb.
                      It remains to show that $\partition[\thr,\sffsstruct]$ satisfies \cc.

                      %Let $\sts\in\Stands$ and $\standbs\psi\in\partk$ and let $\sffsstructt\modelfor \thr\cup\neg\partnk$.
                      %We have to show $\sffsstructt\modelfor\standbs\psi$.

                      Assume to the contrary that there is some $\standbs\psi\in\partk$ such that
                      $\thr\cup\neg\partnk\not\entailss\standbs\psi$.
                      Then there exists an \sff standpoint structure $\sffsstructt$ with $\sffsstructt\modelfor\thr\cup\neg\partnk$ and $\sffsstructt\not\modelfor\standbs\psi$.

                      We construct the \sff standpoint structure $\sffsstruct''$ as follows.
                      For every $\sts\in\Stands$, we assume (w.l.o.g.) that $\stani(\sts)\cap(\stani'(\sts)\cup\stano'(\sts))=\emptyset$ and set
                      \begin{align*}
                          \stani''(\sts) & \eqdef \stani(\sts)                   \\
                          \stano''(\sts) & \eqdef \stani'(\sts)\cup\stano'(\sts) \\
                          \eval''(\pr'') & \eqdef
                          \begin{cases}
                              \eval(\pr'')  & \text{ if } \pr''\in\stani(\sts), \\
                              \eval'(\pr'') & \text{ otherwise.}
                          \end{cases}
                      \end{align*}
                      Observe that $\sffsstruct''\sprefeq\sffsstruct$ by construction.
                      We next show the first main helper result.
                      \begin{claim}
                          \label{claim:e}
                          For all $\sts\in\Stands$, $\pr\in\stani(\sts)=\stani''(\sts)$ and all $\varphi\in\subf(\thr)$, we have
                          \[
                              \sffsstruct,\pr\modelfor\varphi \iff \sffsstruct'',\pr\modelfor\varphi
                          \]
                          \begin{claimproof}
                              We use induction on the structure of $\varphi$;
                              there is only one interesting case, namely $\varphi=\standbu\xi$.
                              \begin{align*}
                                  \sffsstruct,\pr\modelfor\standbu\xi & \iff \forall\pr_{\stu}\in\stani(\stu):\sffsstruct,\pr_{\stu}\modelfor\xi                   \\
                                                                      & \iff \forall\pr_{\stu}\in\stani(\stu):\sffsstruct'',\pr_{\stu}\modelfor\xi   & \text{(IH)} \\
                                                                      & \iff \forall\pr_{\stu}\in\stani''(\stu):\sffsstruct'',\pr_{\stu}\modelfor\xi               \\
                                                                      & \iff \sffsstruct'',\pr\modelfor\standbu\xi
                              \end{align*}
                          \end{claimproof}
                      \end{claim}
                      Now for the next helper result.
                      \begin{claim}
                          \label{claim:f}
                          For all $\sts\in\Stands$, $\pr'\in\stani'(\sts)\cup\stano'(\sts)=\stano''(\sts)$, and $\varphi\in\subf(\thr)$, we have
                          \[
                              \sffsstructt,\pr'\modelfor\varphi \iff \sffsstruct'',\pr'\modelfor\varphi
                          \]
                          \begin{claimproof}
                              We use induction on the structure of $\varphi$;
                              again, there is only one non-trivial case, $\varphi=\standbu\xi$.
                              \begin{description}
                                  \item[\normalfont“$\!\implies\!$”:]
                                        Let $\sffsstructt,\pr'\modelfor\standbu\xi$.

                                        Firstly, as $\sffsstructt\modelfor\standbu\xi$, by $\sffsstructt\modelfor\neg\partnk$ we have $\standbu\xi\notin\partnk$.
                                        Thus $\standbu\xi\in\partk$ whence $\sffsstruct\modelfor\standbu\xi$.
                                        Thus for all $\pr\in\stani(\stu)$ we have $\sffsstruct,\pr\modelfor\xi$.
                                        Thus by \Cref{claim:e} we get $\sffsstruct'',\pr_{\stu}\modelfor\xi$ for all $\pr_{\stu}\in\stani''(\stu)$, which we denote by $(\ddagger)$.

                                        \begin{itemize}
                                            \item $\pr'\in\stani'(\all)$:
                                                  Then $(\ddagger)$ directly yields $\sffsstruct'',\pr'\modelfor\standbu\xi$.
                                            \item $\pr'\in\stano'(\all)$:
                                                  Then for all $\pr_{\stu}'\in\stani'(\stu)\cup\stano'(\stu)$, we have $\sffsstructt,\pr_{\stu}'\modelfor\xi$.
                                                  By the induction hypothesis, thus, for all $\pr_{\stu}''\in\stano''(\stu)$ we have $\sffsstruct'',\pr_{\stu}''\modelfor\xi$.

                                                  By $(\ddagger)$, also for all $\pr_{\stu}''\in\stani''(\stu)$ we have $\sffsstruct'',\pr_{\stu}''\modelfor\xi$.

                                                  In combination, $\sffsstruct'',\pr''\modelfor\xi$ for all $\pr''\in\stani''(\stu)\cup\stano''(\stu)$.
                                                  Thus $\sffsstruct'',\pr'\modelfor\standbu\xi$.
                                        \end{itemize}
                                  \item[\normalfont“$\!\impliedby\!$”:]
                                        We show the contrapositive.
                                        Let $\sffsstructt,\pr'\not\modelfor\standbu\xi$.
                                        Then there is a $\pr_{\stu}'\in\stani'(\stu)\cup\stano'(\stu)$ with $\sffsstructt,\pr_{\stu}'\not\modelfor\xi$.
                                        By the induction hypothesis, we obtain $\sffsstruct'',\pr_{\stu}'\not\modelfor\xi$.
                                        Thus we finally get $\sffsstruct'',\pr'\not\modelfor\standbu\xi$.
                              \end{description}
                          \end{claimproof}
                      \end{claim}
                      Now towards showing $\sffsstruct''\modelfor\thr$.
                      \begin{itemize}
                          \item Let $\varphi\in\thr$ be a formula.
                                We have $\sffsstructt\modelfor\varphi$ by assumption, that is, for all $\pr'\in\Precs=\stani'(\all)\cup\stano'(\all)$ we have $\sffsstructt,\pr'\modelfor\varphi$.
                                By \Cref{claim:f}, we get that for all $\pr''\in\stano''(\all)$ we have $\sffsstruct'',\pr''\modelfor\varphi$.
                                Since $\sffsstruct\modelfor\thr$, we also have $\sffsstruct,\pr\modelfor\varphi$ for all $\pr\in\stani(\all)=\Precs$.
                                By \Cref{claim:e}, we get that for all $\pr''\in\stani''(\all)$ we have $\sffsstruct'',\pr''\modelfor\varphi$.
                                In combination, $\sffsstruct''\modelfor\varphi$.
                          \item Let $\sts\sharpens\stu\in\thr$.
                                By $\sffsstruct\modelfor\thr$ we get $\sffsstruct\modelfor\sts\sharpens\stu$ and thus $\stani(\sts)\subseteq\stani(\stu)$.
                                Likewise, $\sffsstructt\modelfor\thr$ yields $\sffsstructt\modelfor\sts\sharpens\stu$, that is,
                                $\stani'(\sts)\subseteq\stani'(\stu)$ as well as
                                $\stano'(\sts)\subseteq\stano'(\stu)$.
                                We thus obtain $\stani''(\sts)=\stani(\sts)\subseteq\stani(\stu)=\stani''(\stu)$;
                                likewise, $\stano''(\sts)=\stani'(\sts)\cup\stano'(\sts)\subseteq\stani'(\stu)\cup\stano'(\stu)=\stano''(\stu)$.
                                %(Sharpening statements $\sts\sharpens\stu$ are satisfied in $\sffsstruct''$ because they are satisfied in $\sffsstruct$ and $\sffsstructt$.)
                      \end{itemize}
                      Overall, we thus get $\sffsstruct''\modelfor\thr$.

                      Now we have $\sffsstructt\not\modelfor\standbs\psi$ by assumption, thus there is a $\pr_{\psi}'\in\stani'(\sts)\cup\stano'(\sts)=\stano''(\sts)$ with $\sffsstructt,\pr_{\psi}'\not\modelfor\psi$.
                      By \Cref{claim:f} we thus obtain $\sffsstruct'',\pr_{\psi}'\not\modelfor\psi$, that is, $\sffsstruct''\not\modelfor\standbs\psi$.
                      On the other hand, $\standbs\psi\in\partk$ means by construction that $\sffsstruct\modelfor\standbs\psi$, whence $\eval''(\stano''(\sts))\not\subseteq\eval(\stani(\sts))$.

                      In addition to $\sffsstruct''\sprefeq\sffsstruct$ by construction, this yields $\sffsstruct''\spref\sffsstruct$ with $\sffsstruct''\modelfor\thr$ although $\sffsstruct$ is a minimal model of $\thr$.
                      Contradiction.
            \end{enumerate}
        \end{proof}
    \else
        \begin{proofsketch}
            With \mbox{$\sffsstruct\modelfor\thr$}, it is clear to define the partition \mbox{$\partition[\thr,\sffsstruct]=\Partition$} by
            \mbox{$\partk\eqdef\set{\standbu\phi\in\modalsub[\thr]\guard\sffsstruct\modelfor\standbu\phi}$}
            and
            \mbox{$\partnk\eqdef\modalsub[\thr]\setminus\partk$}.
            A first helper claim then again connects the propositional and modal readings of formulas:
            for any \mbox{$\sts\in\Stands$} and \mbox{$\pr\in\stani(\sts)$} we define the propositional valuation
            %$F_\pr \eqdef \eval(\pr)\cup\set{ \standbu\phi\in\subf(\thr) \guard \sffsstruct\modelfor\standbu\phi }$
            \mbox{$F_\pr \eqdef \eval(\pr)\cup\partk\subseteq\Atomsmod$}
            for which it holds for all \mbox{$\varphi\in\subf(\thr)$} that
            \mbox{$F_\pr\modelfor\varphi$} iff \mbox{$\sffsstruct,\pr\modelfor\varphi$}.
            This then serves to establish another helper result by which
            for any \mbox{$\sts\in\Stands$} and \mbox{$\pr\in\stani(\sts)$}, we have \mbox{$F_\pr\modelfor\partprop[\sts]$};
            in turn, this can be used to prove \ca and \cb.
            For \cc, we do a proof by contradiction and assume there is some \mbox{$\standbs\psi\in\partk$} such that \mbox{$\thr\cup\neg\partnk\not\entailss\standbs\psi$}.
            Then there exists an \sff standpoint structure $\sffsstructt$ with \mbox{$\sffsstructt\modelfor\thr\cup\neg\partnk$} and \mbox{$\sffsstructt\not\modelfor\standbs\psi$}.
            The two structures $\sffsstruct$ and $\sffsstructt$ can be combined to a third structure $\sffsstruct''$ with \mbox{$\sffsstruct''\sprefeq\sffsstruct$} that can be shown to also be a model for $\thr$ with \mbox{$\sffsstruct''\not\modelfor\standbs\psi$} (this is actually the most laborious part of the proof).
            But this then yields \mbox{$\sffsstruct''\spref\sffsstruct$} with \mbox{$\sffsstruct''\modelfor\thr$} although $\sffsstruct$ is a minimal model of $\thr$, the desired contradiction.
        \end{proofsketch}
    \fi
\end{theorem}

In our running~\Cref{example}, the minimal model $\sffsstructtwomin$ of theory $\thrtwo$ given earlier can be characterised as pair $\Partition['_2]$, with
\mbox{$\set{\standball\sovu,\standball\spreg,\standbx{\sdone}\spcos,\standbx{\sdone}\shorm} \subseteq \partk_2'$} and \mbox{$\set{\standbx{\sdtwo}\sfha}\subseteq\partnk_2'$}, where the latter containment intuitively expresses that $\neg\fha$ is conceivable from standpoint $\sdtwotext$.

% expand -- \input{sections/complexity-non-monotonic}
%%%%%%%%%%%%%%%%%%%%%%%%%%%%%%%%%%%%%%%%%%%%%%%%%%%%%%%%

\subsection{Complexity of Non-Monotonic S4F SL}
\label{sec:sffs-complexity-nonmon}

Given that each minimal model of an \sff standpoint logic theory \mbox{$\thr\subseteq\langss$} can be parsimoniously represented via a partition $\partition$ of $\modalsub$, for dealing with various decision problems surrounding minimal models we can resort to computing with such representations instead of computing with actual models (that might be of worst-case exponential size).

\begin{theorem}
    \label{thm:complexity:non-monotonic:existence}
    Deciding existence of a minimal model for an \sff standpoint logic theory $\thr\subseteq\langss$ is $\SigmaP[2]$-complete.
    \begin{proof}
        The lower bound follows from unimodal \sff \citep{Schwarz-Truszczynski-1993}, so let us focus on containment.
        The general approach is clear:
        Given \mbox{$\thr\subseteq\langss$}, we obtain $\modalsub$, guess a partition \mbox{$\partition=\Partition$}, and verify \ca, \cb, and \cc.
        Verifying \ca and \cb can be done using the NP oracle for the polynomially many satisfiability checks of propositional logic, where all involved theories are polynomial in the size of $\thr$.
        For \cc, we make use of \Cref{t:sffs-p-satisfiability} and employ the \NP oracle to do \mbox{$\card{\partk}\leq\size{\thr}$} many satisfiability checks of monotonic \sff standpoint logic.
    \end{proof}
\end{theorem}

The idea for credulous and sceptical entailment is then to many-one-reduce it to minimal model existence as follows\/:

\begin{theorem}
    \label{thm:complexity:non-monotonic:entailment-reduction}
    Let \mbox{$\thr\subseteq\langss$} be a theory, \mbox{$\xi\in\langss$} be a formula, and assume atom \mbox{$z\in\Atoms$} does not occur in \mbox{$\thr\cup\set{\xi}$}.
    \begin{enumerate}
        \item $\thr\dentailscred\xi$ \;iff\; $\Tcred$ has a minimal model, where
              \begin{mygather}
                  \Tcred \eqdef \thr\cup\set{(\standball\neg\standball z \land \standball\neg\standball\xi ) \limplies \standball z}
              \end{mygather}
        \item $\thr\ndentailsscep\xi$ \;iff\; $\Tscep$ has a minimal model, where
              \begin{mygather}
                  \Tscep \eqdef \thr\cup\set{(\standball\neg\standball z \land \neg\standball\neg\standball\xi ) \limplies \standball z}
              \end{mygather}
    \end{enumerate}
    \iflong
        \begin{proof}
            We first observe the following:
            For any S5 standpoint structure $\sffsstruct$,
            % \begin{align*}
            %      & \phiff \sffsstruct\modelfor\xi                                                          \\
            %      & \iff \forall\pr\in\stani(\all): \sffsstruct,\pr\modelfor\xi                             \\
            %      & \iff \exists\pr\in\stani(\all): \sffsstruct,\pr\modelfor\standball\xi                   \\
            %      & \iff \exists\pr\in\stani(\all): \sffsstruct,\pr\not\modelfor\neg\standball\xi           \\
            %      & \iff \forall\pr\in\stani(\all): \sffsstruct,\pr\not\modelfor\standball\neg\standball\xi \\
            %      & \iff \forall\pr\in\stani(\all): \sffsstruct,\pr\modelfor\neg\standball\neg\standball\xi \\
            %      & \iff \sffsstruct\modelfor\neg\standball\neg\standball\xi
            % \end{align*}
            % which we will denote by $(\dagger)$.
            \begin{gather}
                \label{eq:xi-iff-nbnbxi}
                \sffsstruct\modelfor\xi \iff \sffsstruct\modelfor\standball\xi %\iff \sffsstruct\modelfor\standd{\all}\standball\xi
                \iff \sffsstruct\modelfor\neg\standball\neg\standball\xi
                \tag{$\dagger$}
            \end{gather}
            \begin{enumerate}
                \item
                      \begin{description}
                          \item[\normalfont“if”:]
                                Let $\sffsstruct$ be a minimal model of $\Tcred$.
                                We will show that $\sffsstruct$ is a minimal model of $\thr$ with $\sffsstruct\modelfor\xi$.

                                By $\sffsstruct\modelfor\Tcred$, in particular we get
                                \(
                                \sffsstruct\modelfor(\standball\neg\standball z \land \standball\neg\standball\xi ) \limplies \standball z
                                \).
                                We will next show that $\sffsstruct\not\modelfor\standball z$:
                                Assume to the contrary that $\sffsstruct\modelfor\standball z$.
                                We construct an \sff standpoint structure $\sffsstructt$ as follows\/:
                                \begin{align*}
                                    \Precs'       & \eqdef \Precs \cup \set{ \pr' \guard \pr\in\Precs }                               \\
                                    \stani'(\sts) & \eqdef \stani(\sts)                                 & \text{ for } \sts\in\Stands \\
                                    \stano'(\sts) & \eqdef \set{ \pr' \guard \pr\in\stani(\sts) }       & \text{ for } \sts\in\Stands \\
                                    \eval'(\pr)   & \eqdef \eval(\pr)                                   & \text{ for } \pr\in\Precs   \\
                                    \eval'(\pr')  & \eqdef \eval(\pr)\setminus\set{z}                   & \text{ for } \pr\in\Precs
                                \end{align*}
                                By construction, $\sffsstructt\spref\sffsstruct$ (as $\sffsstructt\not\modelfor\standball z$).
                                Since $\sffsstruct$ is a minimal model of $\Tcred$, we obtain $\sffsstructt\not\modelfor\Tcred$.
                                On the other hand, $\sffsstructt\modelfor\thr$ as $z$ does not occur in $\thr$.
                                Therefore, $\sffsstructt\not\modelfor(\standball\neg\standball z \land \standball\neg\standball\xi ) \limplies \standball z$, that is, we get
                                $\sffsstructt\modelfor\standball\neg\standball z \land \standball\neg\standball\xi$ and in particular $\sffsstructt\modelfor\standball\neg\standball z$, that is, $\sffsstructt\not\modelfor\standball z$.
                                Contradiction.

                                This establishes $\sffsstruct\not\modelfor\standball z$.
                                Therefore, we obtain that $\sffsstruct\not\modelfor \standball\neg\standball z \land \standball\neg\standball\xi$, that is, (i) $\sffsstruct\not\modelfor\standball\neg\standball z$ or (ii) $\sffsstruct\not\modelfor\standball\neg\standball\xi$.

                                In case (i), there exists a precisification $\pr\in\stani(\all)$ with $\sffsstruct,\pr\not\modelfor\neg\standball z$, that is, $\sffsstruct,\pr\modelfor\standball z$.
                                As we have established above, this is impossible.
                                Thus (ii) must be the case, that is,
                                $\sffsstruct\not\modelfor\standball\neg\standball\xi$.
                                Then there is a $\pr\in\Precs$ with $\sffsstruct,\pr\not\modelfor\neg\standball\xi$, that is, $\sffsstruct,\pr\modelfor\standball\xi$.
                                Thus we obtain $\sffsstruct\modelfor\xi$.

                                It remains to show that $\sffsstruct$ is a minimal model of $\thr$.
                                By $\thr\subseteq\Tcred$ it is clear that $\sffsstruct\modelfor\thr$.
                                Assume that $\sffsstruct''\sprefeq\sffsstruct$ with $\sffsstruct''\modelfor\thr$.
                                Then by $\sffsstruct\modelfor\xi$ and $(\dagger)$ we get $\sffsstruct\modelfor\neg\standball\neg\standball\xi$, which in turn by \Cref{lemma2:item-4} of \Cref{lemma-knowledge-passing} yields $\sffsstruct''\modelfor\neg\standball\neg\standball\xi$.
                                Therefore,
                                \[
                                    \sffsstruct''\modelfor (\standball\neg\standball z \land \standball\neg\standball\xi ) \limplies \standball z
                                \]
                                whereby $\sffsstruct''\modelfor\Tcred$.
                                Since $\sffsstruct$ is a minimal model of $\Tcred$, we obtain $\sffsstruct''\sprefiso\sffsstruct$ as required.
                          \item[\normalfont“only if”:]
                                Let $\sffsstruct$ be a minimal model of $\thr$ with $\sffsstruct\modelfor\xi$.
                                By $(\dagger)$, we immediately obtain
                                %Then by definition also $\sffsstruct\modelfor\standball\xi$, thus since $\stani(\all)=\Precs\neq\emptyset$, there is a $\pr\in\stani(\all)$ with $\sffsstruct,\pr\modelfor\standball\xi$, that is, $\sffsstruct,\pr\not\modelfor\neg\standball\xi$.
                                %By definition, we obtain that for all $\pr'\in\stani(\all)$ we have $\sffsstruct,\pr'\not\modelfor\standball\neg\standball\xi$, that is, $\sffsstruct,\pr'\modelfor\neg\standball\neg\standball\xi$, and therefore
                                $\sffsstruct\modelfor\neg\standball\neg\standball\xi$, whence
                                \[
                                    \sffsstruct\modelfor (\standball\neg\standball z \land \standball\neg\standball\xi ) \limplies \standball z
                                \]
                                and consequently $\sffsstruct\modelfor\Tcred$.

                                Now consider an arbitrary $\sffsstructt\sprefeq\sffsstruct$ with $\sffsstructt\modelfor\Tcred$.
                                By $\thr\subseteq\Tcred$, we obtain $\sffsstructt\modelfor\thr$;
                                since $\sffsstruct$ is a minimal model of $\thr$, we obtain $\sffsstructt\sprefiso\sffsstruct$.
                                Since $\sffsstructt$ was arbitrary, we get that $\sffsstruct$ is a minimal model of $\Tcred$.
                                %It thus follows from \Cref{lem:minimal-model-superset} and $\thr\subseteq\Tcred$ that $\sffsstruct$ is a minimal model of $\Tcred$.
                      \end{description}
                \item Similar:
                      \begin{description}
                          \item[\normalfont“if”:]
                                Let $\sffsstruct$ be a minimal model of $\Tscep$.
                                As above, we can establish that $\sffsstruct\not\modelfor\standball z$ and thus $\sffsstruct\not\modelfor\neg\standball\neg\standball\xi$.
                                By \eqref{eq:xi-iff-nbnbxi}, we thus get $\sffsstruct\not\modelfor\xi$.
                                It remains to show that $\sffsstruct$ is a minimal model of $\thr$.
                                Consider $\sffsstructt\sprefeq\sffsstruct$ with $\sffsstructt\modelfor\thr$.
                                From \Cref{lemma2:item-3} of \Cref{lemma-knowledge-passing} and $\sffsstruct\not\modelfor\standball\neg\standball\neg\standball\xi$ we get $\sffsstructt\not\modelfor\standball\neg\standball\neg\standball\xi$ and thus $\sffsstructt\modelfor\standball\neg\standball\xi$.
                                But then $\sffsstructt\modelfor\Tscep$ and $\sffsstructt\sprefiso\sffsstruct$.
                          \item[\normalfont“only if”:]
                                If $\sffsstruct$ is a minimal model of $\thr$ with $\sffsstruct\not\modelfor\xi$, then by \eqref{eq:xi-iff-nbnbxi} we obtain $\sffsstruct\not\modelfor\neg\standball\neg\standball\xi$.
                                Then $\sffsstruct\modelfor\standball\neg\standball\xi$ and $\sffsstruct\modelfor\Tscep$.
                                For any $\sffsstructt\modelfor\Tscep$ with $\sffsstructt\sprefeq\sffsstruct$ we clearly have $\sffsstructt\modelfor\thr$, whence $\sffsstructt\sprefiso\sffsstruct$.
                      \end{description}
            \end{enumerate}
        \end{proof}
    \else
        \begin{proofsketch}
            Intuitively, the additional atom $z$ and implications serve as integrity constraints that eliminate all minimal models that do (not) contain the formula $\xi$ to be queried:
            A minimal model $\sffsstruct$ is an S5 standpoint structure by definition, so either
            (a) \mbox{$\sffsstruct\modelfor\standball z$} or
            (b) \mbox{$\sffsstruct\modelfor\neg\standball z$}, and in case (b) then \mbox{$\sffsstruct\modelfor\standball\neg\standball z$} due to negative introspection.
            It can be shown that (a) is impossible because the only reason for it requires (b); thus the other conjunct $(\neg)\standball\neg\standball\xi$ in the implication's prerequisite must be dissatisfied.
            For example, in credulous entailment, \mbox{$\sffsstruct\not\modelfor\standball\neg\standball\xi$} yields \mbox{$\sffsstruct\modelfor\standball\xi$};
            the conclusion is dual for sceptical reasoning.
        \end{proofsketch}
    \fi
\end{theorem}

Given $\thr$ and $\xi$, the theories $\Tcred$ and $\Tscep$ can clearly be constructed in deterministic polynomial time.
Thus the respective complexities follow, with lower bounds obtained from the proper fragment of default logic~\citep{Gottlob92}.

\begin{corollary}
    \label{cor:complexity:non-monotonic:reasoning}
    The problem “Given $T$ and $\xi$, does $T\dentailscred\xi$ hold?” is \SigmaP[2]-complete.
    The problem “Given $T$ and $\xi$, does $T\dentailsscep\xi$ hold?” is \PiP[2]-complete.
\end{corollary}

% expand -- \input{sections/asp-encoding}
\section{Disjunctive ASP Encoding}
\label{sec:asp-encoding}

We provide a proof-of-concept implementation of \sff standpoint logic by developing an encoding in disjunctive answer set programming using the saturation technique~\citep{EiterG95}.
The main insight underlying the encoding is to equivalently reformulate the guess-and-check approach described in the previous section %\Cref{sec:sffs-complexity-nonmon}
% -- guess a partition of the set $T^{\know}$ of given theory $T$'s subformulas of the form $\know\varphi$ and check consistency and minimality --
as follows:
\begin{enumerate}
    \item[(1)] We guess, independently, the following:
          \begin{enumerate}
              \item a partition $\partition=\Partition$ of $\modalsub$;
              \item for every standpoint name \mbox{$\sts\in\Stands$}, a propositional valuation $\val_{\sts}$ of the extended vocabulary $\Atomsmod=\Atoms\cup\modalsub$;
              \item for each $\standbu\xi\in\modalsub$, a valuation $\val_{\standbu\xi}\subseteq\Atomsmod$.
          \end{enumerate}
    \item[(2)] We next verify (in deterministic polynomial time):
          \begin{enumerate}
              \item $\val_{\sts}\modelfor\partprop[\sts]$ for all $\sts\in\Stands$;
              \item for every $\standbu\phi\in\partnk$, we have that $\val_{\standbu\phi}\modelfor\partprop[\sts]\cup\set{\neg\phi}$ for all $\sts\in\Stands$ with $T\provess\stu\sharpens\sts$. %, where the latter relation is directly expressed in the encoding via positive rules.
          \end{enumerate}
    \item[(3)] We finally verify (using the \NP~oracle) that
          for every \mbox{$\standbs\psi\in\partk$}, we have that $T\cup\neg\partnk\entailss\standbs\psi$.
\end{enumerate}

\noindent Our ASP encoding now guesses just like in (1) above, verifies (2) via straightforward evaluation of propositional formulas (with $\thr\provess\stu\sharpens\sts$ being obtained directly via rules), and implements (3) via a saturation encoding that checks, for every \mbox{$\standbs\psi\in\partk$} independently, that \mbox{$T\cup\neg\partnk\cup\set{\neg\standbs\psi}$} is unsatisfiable.
Each such check makes use of the small model property of \sff standpoint logic (\Cref{thm:small-model-property}) and works by verifying that all \sff standpoint structures up to the maximal possible size are not models;
technically, model candidates are disjunctively guessed and then checked off if they violate some required property.
\iflong In order to evaluate the $\standbs$ modalities in a strictly positive way (for saturation), we introduce a total ordering on subformulas of the logic, which is induced by a total ordering on the atoms $\Atoms$ and standpoint names $\Stands$ that has to be given as additional input by the user. \fi
The full encoding is available at \url{https://github.com/cl-tud/nm-s4fsl-asp}.

Our implementation continues and generalises a long line of research implementing default and autoepistemic reasoning formalisms via answer set programming:
\citet{JunkerK90} implemented default and autoepistemic logics via truth maintenance systems, which are known to be equivalent to logic programs under the stable model semantics~\citep{ReinfrankDB89}.
The system {\textsf{dl2asp}}~\citep{ChenWZZ10} works similarly to the work of \citeauthor{JunkerK90}.
\citet{JiS14} provided a disjunctive ASP encoding of default logic via the logic of GK~\citep{LinS92}.% , where the latter can be modularly translated into \sff~\citep{SchwarzT94}.
%Our approach instead uses \sff, which is at least as general as GK (\citet{SchwarzT94} show how to modularly translate the logic of GK into \sff) and arguably easier than GK because there is only one modality.

\iflong
    As a by-product, our encoding also provides an implementation of ordinary propositional standpoint logic~\citep{GomezAlvarezR21}:
    According to \Cref{thm:characterisation-minimal-model:sound,thm:characterisation-minimal-model:complete} the S5 standpoint structure represented by a partition $\partition$ is a model of a theory $\thr$ iff $\partition$ satisfies \ca and \cb.
    In combination with \Cref{thm:generalisation} (\Cref{itm:generalisation:sl}), we obtain the desired correctness.
    To leave out the check for \cc, one simply removes the corresponding part of the encoding (which is in a separate file).
    While some standpoint description logics have been implemented before \citep{EmmrichAS23,AlvarezRS23b}, this is, to the best of our knowledge, the first implementation of propositional (S5) standpoint logic.
\fi

% expand -- \input{sections/discussion}
%%%%%%%%%%%%%%%%%%%%%%%%%%%%%%%%%%%%%%%%%%%

\section{Discussion}
\label{sec:discussion}

%% CONCLUSION
In this paper, we introduced \sff standpoint logic, which combines and generalises propositional standpoint logic \citep{GomezAlvarezR21} and (non-monotonic) \sff \citep{SchwarzT94}.
It constitutes the first full-fledged, unrestricted non-monotonic standpoint logic covering, by corollary, standpoint default logic, standpoint answer set programming, and standpoint argumentation frameworks.
We demonstrated that the addition of multiple standpoints to non-monotonic \sff comes at no additional computational cost, and based on this insight presented a disjunctive ASP encoding that implements our logic.

%% RELATED TOPICS

%% FUTURE WORK

For future work, we are interested in simplifying our decision procedure for the proper fragments of standpoint logic programs and standpoint argumentation frameworks, as the satisfiability problems of the base languages are easier (\NP-complete; \citealp{BidoitF91}, \citealp{MarekT91}, LPs; \citealp{DunneW09}, AFs) unless the polynomial hierarchy collapses.
We also want to study strong equivalence for \sff standpoint logic;
the case of unimodal S4F was studied by \citet{Truszczynski07}.
% HS: create some space for further intuitions and explanations.
\iflong
    Finally, it is worthwhile to develop a proof system for our new logic.
    \sff has a proof system via the axioms (\axK), (\axT), (\axf), and (\axF) \citep{segerberg};
    propositional SL has proof systems as well \citep{GomezAlvarezR21,LyonA22}.
    It is open whether these proof systems can be combined into one for \sff standpoint logic.
\fi

\newcommand{\projectname}[1]{#1}
\section*{Acknowledgements}

This work was supported by funding from BMFTR (Federal Ministry of Research, Technology and Space) within projects
\projectname{KIMEDS} (grant no.~GW0552B),
\projectname{MEDGE} (grant no.~16ME0529),
\projectname{SEMECO} (grant no.~03ZU1210B), and
\projectname{SECAI} (via DAAD project 57616814, School of Embedded Composite AI, \url{https://secai.org/}, as part of the program Konrad Zuse Schools of Excellence in Artificial Intelligence).

\bibliography{references}

@inproceedings{Truszczynski07,
  author    = {Mirosław Truszczyński},
  title     = {The Modal Logic {S4F}, the Default Logic, and the Logic Here-and-There},
  booktitle = {{Proceedings of the Twenty-Second {AAAI} Conference on Artificial Intelligence, July 22--26, 2007, Vancouver, British Columbia, Canada}},
  pages     = {508--514},
  publisher = {{AAAI} Press},
  year      = {2007},
  url       = {http://www.aaai.org/Library/AAAI/2007/aaai07-080.php}
}

@article{SchwarzT94,
  author  = {Grigori Schwarz and Mirosław Truszczyński},
  title   = {Minimal Knowledge Problem: {A} New Approach},
  journal = {Artif. Intell.},
  volume  = {67},
  number  = {1},
  pages   = {113--141},
  year    = {1994},
  url     = {https://doi.org/10.1016/0004-3702(94)90013-2}
}

@inproceedings{Schwarz92,
  author    = {Grigori Schwarz},
  title     = {Minimal Model Semantics for Nonmonotonic Modal Logics},
  booktitle = {Proceedings of the Seventh Annual Symposium on Logic in Computer Science ({LICS} '92), Santa Cruz, California, USA, June 22--25, 1992},
  pages     = {34--43},
  publisher = {{IEEE} Computer Society},
  year      = {1992},
  url       = {https://doi.org/10.1109/LICS.1992.185517}
}

@article{McDermottD80,
  author  = {Drew V. McDermott and Jon Doyle},
  title   = {Non-Monotonic Logic {I}},
  journal = {Artif. Intell.},
  volume  = {13},
  number  = {1--2},
  pages   = {41--72},
  year    = {1980},
  url     = {https://doi.org/10.1016/0004-3702(80)90012-0}
}

@inproceedings{GomezAlvarezR21,
  author    = {Lucía {Gómez Álvarez} and Sebastian Rudolph},
  editor    = {Fabian Neuhaus and Boyan Brodaric},
  title     = {Standpoint Logic: Multi-Perspective Knowledge Representation},
  booktitle = {Formal Ontology in Information Systems -- Proceedings of the Twelfth International Conference, {FOIS} 2021, Bozen-Bolzano, Italy, September 11--18, 2021},
  series    = {Frontiers in Artificial Intelligence and Applications},
  volume    = {344},
  pages     = {3--17},
  publisher = {{IOS} Press},
  year      = {2021},
  url       = {https://doi.org/10.3233/FAIA210367}
}

@inproceedings{Schwarz-Truszczynski-1993,
  author    = {Grigori Schwarz and
               Mirosław Truszczyński},
  editor    = {Georg Gottlob and
               Alexander Leitsch and
               Daniele Mundici},
  title     = {Nonmonotonic Reasoning is Sometimes Simpler},
  booktitle = {Computational Logic and Proof Theory, Third Kurt G{\"{o}}del Colloquium, KGC'93, Brno, Czech Republic, August 24--27, 1993, Proceedings},
  series    = {Lecture Notes in Computer Science},
  volume    = {713},
  pages     = {313--324},
  publisher = {Springer},
  year      = {1993},
  url       = {https://doi.org/10.1007/BFb0022579},
  optdoi    = {10.1007/BFB0022579}
}

@inproceedings{AlvarezRS22,
  author    = {Lucía {Gómez Álvarez} and
               Sebastian Rudolph and
               Hannes Strass},
  editor    = {Ulrike Sattler and
               Aidan Hogan and
               C. Maria Keet and
               Valentina Presutti and
               Jo{\~{a}}o Paulo A. Almeida and
               Hideaki Takeda and
               Pierre Monnin and
               Giuseppe Pirr{\`{o}} and
               Claudia d'Amato},
  title     = {{How to Agree to Disagree -- Managing Ontological Perspectives using Standpoint Logic}},
  booktitle = {The Semantic Web -- {ISWC} 2022 -- 21st International Semantic Web Conference, Virtual Event, October 23--27, 2022, Proceedings},
  series    = {Lecture Notes in Computer Science},
  volume    = {13489},
  pages     = {125--141},
  publisher = {Springer},
  year      = {2022},
  url       = {https://doi.org/10.1007/978-3-031-19433-7\_8}
}

@phdthesis{segerberg,
  author = {Segerberg, Karl Krister},
  school = {Stanford University, Department of Philosophy},
  title  = {An Essay in Classical Modal Logic},
  year   = {1971}
}

@inproceedings{GiganteAL23,
  author    = {Nicola Gigante and Lucía {Gómez Álvarez} and Tim S. Lyon},
  editor    = {Pierre Marquis and Tran Cao Son and Gabriele Kern{-}Isberner},
  title     = {Standpoint Linear Temporal Logic},
  booktitle = {Proceedings of the 20th International Conference on Principles of Knowledge Representation and Reasoning, {KR} 2023, Rhodes, Greece, September 2--8, 2023},
  pages     = {311--321},
  year      = {2023},
  url       = {https://doi.org/10.24963/kr.2023/31}
}

@inproceedings{AlvarezRS23a,
  author    = {Lucía {Gómez Álvarez} and Sebastian Rudolph and Hannes Strass},
  title     = {Tractable Diversity: Scalable Multiperspective Ontology Management via Standpoint {EL}},
  booktitle = {Proceedings of the Thirty-Second International Joint Conference on Artificial Intelligence, {IJCAI} 2023, 19th--25th August 2023, Macao, SAR, China},
  pages     = {3258--3267},
  publisher = {ijcai.org},
  year      = {2023},
  url       = {https://doi.org/10.24963/ijcai.2023/363}
}

@inproceedings{AlvarezRS23b,
  author    = {Lucía {Gómez Álvarez} and Sebastian Rudolph and Hannes Strass},
  editor    = {Pierre Marquis and Tran Cao Son and Gabriele Kern{-}Isberner},
  title     = {Pushing the Boundaries of Tractable Multiperspective Reasoning: {A} Deduction Calculus for Standpoint {EL+}},
  booktitle = {Proceedings of the 20th International Conference on Principles of Knowledge Representation and Reasoning, {KR} 2023, Rhodes, Greece, September 2--8, 2023},
  pages     = {333--343},
  year      = {2023},
  url       = {https://doi.org/10.24963/kr.2023/33}
}

@inproceedings{LyonA22,
  author    = {Tim S. Lyon and Lucía {Gómez Álvarez}},
  editor    = {Gabriele Kern{-}Isberner and Gerhard Lakemeyer and Thomas Meyer},
  title     = {Automating Reasoning with Standpoint Logic via Nested Sequents},
  booktitle = {Proceedings of the 19th International Conference on Principles of Knowledge Representation and Reasoning, {KR} 2022, Haifa, Israel, July 31 -- August 5, 2022},
  year      = {2022},
  url       = {https://proceedings.kr.org/2022/26/}
}

@article{Reiter80,
  author  = {Raymond Reiter},
  title   = {A Logic for Default Reasoning},
  journal = {Artif. Intell.},
  volume  = {13},
  number  = {1--2},
  pages   = {81--132},
  year    = {1980},
  url     = {https://doi.org/10.1016/0004-3702(80)90014-4}
}

@article{LinS92,
  author  = {Fangzhen Lin and Yoav Shoham},
  title   = {A Logic of Knowledge and Justified Assumptions},
  journal = {Artif. Intell.},
  volume  = {57},
  number  = {2--3},
  pages   = {271--289},
  year    = {1992},
  url     = {https://doi.org/10.1016/0004-3702(92)90019-T}
}

@article{Lifschitz94,
  author  = {Vladimir Lifschitz},
  title   = {Minimal Belief and Negation as Failure},
  journal = {Artif. Intell.},
  volume  = {70},
  number  = {1--2},
  pages   = {53--72},
  year    = {1994},
  url     = {https://doi.org/10.1016/0004-3702(94)90103-1}
}

@article{Dung95,
  author  = {Phan Minh Dung},
  title   = {On the Acceptability of Arguments and its Fundamental Role in Nonmonotonic
             Reasoning, Logic Programming and $n$-Person Games},
  journal = {Artif. Intell.},
  volume  = {77},
  number  = {2},
  pages   = {321--358},
  year    = {1995},
  url     = {https://doi.org/10.1016/0004-3702(94)00041-X}
}

@article{Strass13,
  author  = {Hannes Strass},
  title   = {Approximating operators and semantics for abstract dialectical frameworks},
  journal = {Artif. Intell.},
  volume  = {205},
  pages   = {39--70},
  year    = {2013},
  url     = {https://doi.org/10.1016/j.artint.2013.09.004}
}

@book{Kurucz03,
  title     = {Many-dimensional modal logics: {T}heory and applications},
  author    = {Kurucz, Agi and Wolter, Frank and Zakharyaschev, Michael and Gabbay, Dov M},
  year      = {2003},
  publisher = {Elsevier}
}

@article{Gottlob95a,
  author  = {Georg Gottlob},
  title   = {Translating Default Logic into Standard Autoepistemic Logic},
  journal = {J. {ACM}},
  volume  = {42},
  number  = {4},
  pages   = {711--740},
  year    = {1995},
  url     = {https://doi.org/10.1145/210332.210334}
}

@article{Gottlob92,
  author  = {Gottlob, Georg},
  title   = {{Complexity Results for Nonmonotonic Logics}},
  journal = {Journal of Logic and Computation},
  volume  = {2},
  number  = {3},
  pages   = {397--425},
  year    = {1992},
  month   = {06},
  issn    = {0955--792X},
  url     = {https://doi.org/10.1093/logcom/2.3.397}
}

@article{Moore85,
  author  = {Robert C. Moore},
  title   = {Semantical Considerations on Nonmonotonic Logic},
  journal = {Artif. Intell.},
  volume  = {25},
  number  = {1},
  pages   = {75--94},
  year    = {1985},
  url     = {https://doi.org/10.1016/0004-3702(85)90042-6}
}

@article{GelfondL91,
  author  = {Michael Gelfond and Vladimir Lifschitz},
  title   = {Classical Negation in Logic Programs and Disjunctive Databases},
  journal = {New Gener. Comput.},
  volume  = {9},
  number  = {3/4},
  pages   = {365--386},
  year    = {1991},
  url     = {https://doi.org/10.1007/BF03037169}
}

@article{Bennett11,
  title     = {Standpoint semantics: a framework for formalising the variable meaning of vague terms},
  author    = {Bennett, Brandon},
  journal   = {Understanding Vagueness. Logical, Philosophical and Linguistic Perspectives},
  editor    = {Cintula, Petr and Fermüller, Christian G and Godo, Lluís and Hájek, Petr},
  publisher = {College Publications},
  series    = {Studies in Logic},
  pages     = {261--278},
  year      = {2011}
}

@inproceedings{GomezAlvarezR24,
  author    = {Lucía {Gómez Álvarez} and
               Sebastian Rudolph},
  editor    = {Pierre Marquis and
               Magdalena Ortiz and
               Maurice Pagnucco},
  title     = {Reasoning in {SHIQ} with Axiom- and Concept-Level Standpoint Modalities},
  booktitle = {Proceedings of the 21st International Conference on Principles of
               Knowledge Representation and Reasoning, {KR} 2024, Hanoi, Vietnam.
               November 2--8, 2024},
  year      = {2024},
  url       = {https://doi.org/10.24963/kr.2024/36},
  optdoi    = {10.24963/KR.2024/36}
}

@inproceedings{LeisegangMV2025,
  author    = {Leisegang, Nicholas and Meyer, Thomas and Varzinczak, Ivan},
  title     = {Extending Defeasibility for Propositional Standpoint Logics},
  year      = {2025},
  isbn      = {978-3-032-04589-8},
  publisher = {Springer-Verlag},
  address   = {Berlin, Heidelberg},
  url       = {https://doi.org/10.1007/978-3-032-04590-4_4},
  doi       = {10.1007/978-3-032-04590-4_4},
  booktitle = {Logics in Artificial Intelligence: 19th European Conference, JELIA 2025, Kutaisi, Georgia, September 1--4, 2025, Proceedings, Part II},
  pages     = {43--57},
  numpages  = {15},
  location  = {Kutaisi, Georgia}
}

@inproceedings{ReinfrankDB89,
  author    = {Michael Reinfrank and
               Oskar Dressler and
               Gerhard Brewka},
  editor    = {N. S. Sridharan},
  title     = {On the Relation Between Truth Maintenance and Autoepistemic Logic},
  booktitle = {Proceedings of the 11th International Joint Conference on Artificial Intelligence. Detroit, MI, USA, August 1989},
  pages     = {1206--1212},
  publisher = {Morgan Kaufmann},
  year      = {1989},
  url       = {http://ijcai.org/Proceedings/89-2/Papers/057.pdf}
}

@inproceedings{JunkerK90,
  author    = {Ulrich Junker and
               Kurt Konolige},
  editor    = {Howard E. Shrobe and
               Thomas G. Dietterich and
               William R. Swartout},
  title     = {Computing the Extensions of Autoepistemic and Default Logics with a Truth Maintenance System},
  booktitle = {Proceedings of the 8th National Conference on Artificial Intelligence.  Boston, Massachusetts, USA, July 29 -- August 3, 1990, 2 Volumes},
  pages     = {278--283},
  publisher = {{AAAI} Press / The {MIT} Press},
  year      = {1990},
  url       = {http://www.aaai.org/Library/AAAI/1990/aaai90-043.php}
}

@inproceedings{ChenWZZ10,
  author    = {Yin Chen and
               Hai Wan and
               Yan Zhang and
               Yi Zhou},
  editor    = {Tomi Janhunen and Ilkka Niemel{\"{a}}},
  title     = {dl2asp: Implementing Default Logic via Answer Set Programming},
  booktitle = {Logics in Artificial Intelligence -- 12th European Conference, {JELIA} 2010, Helsinki, Finland, September 13--15, 2010. Proceedings},
  series    = {Lecture Notes in Computer Science},
  volume    = {6341},
  pages     = {104--116},
  publisher = {Springer},
  year      = {2010},
  url       = {https://doi.org/10.1007/978-3-642-15675-5\_11}
}

@inproceedings{JiS14,
  author    = {Jianmin Ji and
               Hannes Strass},
  editor    = {Torsten Schaub and
               Gerhard Friedrich and
               Barry O'Sullivan},
  title     = {From Default and Autoepistemic Logics to Disjunctive Answer Set Programs via the Logic of {GK}},
  booktitle = {{ECAI} 2014 -- 21st European Conference on Artificial Intelligence,
               18--22 August 2014, Prague, Czech Republic -- Including Prestigious
               Applications of Intelligent Systems {(PAIS 2014)}},
  series    = {Frontiers in Artificial Intelligence and Applications},
  volume    = {263},
  pages     = {1039--1040},
  publisher = {{IOS} Press},
  year      = {2014},
  url       = {https://doi.org/10.3233/978-1-61499-419-0-1039}
}

@article{teede2010pcos,
  author    = {H. Teede and A. Deeks and L. Moran},
  title     = {Polycystic ovary syndrome: a complex condition with psychological, reproductive and metabolic manifestations that impacts on health across the lifespan},
  journal   = {BMC Medicine},
  volume    = {8},
  pages     = {41},
  year      = {2010},
  month     = {June},
  optdoi    = {10.1186/1741-7015-8-41},
  pmid      = {20591140},
  pmc       = {PMC2909929},
  issn      = {1741-7015},
  publisher = {BioMed Central},
  address   = {England},
  language  = {English},
  note      = {Published online 2010-06-30.}
}

@inproceedings{GS2024,
  author    = {Piotr Gorczyca and Hannes Strass},
  title     = {Adding Standpoint Modalities to Non-Monotonic {S4F:} Preliminary Results},
  editor    = {Nina Gierasimczuk and Jesse Heyninck},
  booktitle = {Proceedings of the 22nd International Workshop on Non-Monotonic Reasoning},
  year      = {2024},
  month     = {November}
}

@article{EiterG95,
  author  = {Thomas Eiter and
             Georg Gottlob},
  title   = {On the Computational Cost of Disjunctive Logic Programming: Propositional Case},
  journal = {Ann. Math. Artif. Intell.},
  volume  = {15},
  number  = {3--4},
  pages   = {289--323},
  year    = {1995},
  url     = {https://doi.org/10.1007/BF01536399}
}

@article{HalpernM92,
  author  = {Joseph Y. Halpern and
             Yoram Moses},
  title   = {A Guide to Completeness and Complexity for Modal Logics of Knowledge and Belief},
  journal = {Artif. Intell.},
  volume  = {54},
  number  = {2},
  pages   = {319--379},
  year    = {1992},
  url     = {https://doi.org/10.1016/0004-3702(92)90049-4}
}

@inproceedings{DemriW24,
  author    = {St{\'{e}}phane Demri and
               Przemysław Andrzej Wałega},
  editor    = {Ulle Endriss and
               Francisco S. Melo and
               Kerstin Bach and
               Alberto Jos{\'{e}} Bugar{\'{\i}}n Diz and
               Jose Maria Alonso{-}Moral and
               Sen{\'{e}}n Barro and
               Fredrik Heintz},
  title     = {Computational Complexity of Standpoint {LTL}},
  booktitle = {{ECAI} 2024 -- 27th European Conference on Artificial Intelligence, 19--24 October 2024, Santiago de Compostela, Spain -- Including 13th Conference on Prestigious Applications of Intelligent Systems {(PAIS 2024)}},
  series    = {Frontiers in Artificial Intelligence and Applications},
  volume    = {392},
  pages     = {1206--1213},
  publisher = {{IOS} Press},
  year      = {2024},
  url       = {https://doi.org/10.3233/FAIA240616}
}

@article{KrausLM90,
  author  = {Sarit Kraus and
             Daniel Lehmann and
             Menachem Magidor},
  title   = {Nonmonotonic Reasoning, Preferential Models and Cumulative Logics},
  journal = {Artif. Intell.},
  volume  = {44},
  number  = {1--2},
  pages   = {167--207},
  year    = {1990},
  url     = {https://doi.org/10.1016/0004-3702(90)90101-5}
}

@article{abs-2502-20193,
  author     = {Rajab Aghamov and
                Christel Baier and
                Toghrul Karimov and
                Rupak Majumdar and
                Jo{\"{e}}l Ouaknine and
                Jakob Piribauer and
                Timm Spork},
  title      = {Model Checking Linear Temporal Logic with Standpoint Modalities},
  journal    = {CoRR},
  volume     = {abs/2502.20193},
  year       = {2025},
  url        = {https://doi.org/10.48550/arXiv.2502.20193},
  eprinttype = {arXiv},
  eprint     = {2502.20193}
}

@inproceedings{Shvarts90,
  author    = {Grigori F. Shvarts},
  editor    = {Rohit Parikh},
  title     = {Autoepistemic Modal Logics},
  booktitle = {Proceedings of the 3rd Conference on Theoretical Aspects of Reasoning about Knowledge, Pacific Grove, CA, USA, March 1990},
  pages     = {97--109},
  publisher = {Morgan Kaufmann},
  year      = {1990}
}

@phdthesis{GomezAlvarez19,
  author = {Lucía {Gómez Álvarez}},
  title  = {Standpoint logic: a logic for handling semantic variability, with applications to forestry information},
  school = {University of Leeds, {UK}},
  year   = {2019},
  url    = {https://ethos.bl.uk/OrderDetails.do?uin=uk.bl.ethos.804558}
}

@article{Rosati06,
  author  = {Riccardo Rosati},
  title   = {Multi-modal nonmonotonic logics of minimal knowledge},
  journal = {Ann. Math. Artif. Intell.},
  volume  = {48},
  number  = {3-4},
  pages   = {169--185},
  year    = {2006},
  url     = {https://doi.org/10.1007/s10472-007-9046-5},
  optdoi  = {10.1007/S10472-007-9046-5}
}

@inproceedings{LeisegangMR24,
  author    = {Nicholas Leisegang and Thomas Meyer and Sebastian Rudolph},
  title     = {Towards Propositional {KLM-Style} Defeasible Standpoint Logics},
  editor    = {Aurona Gerber and Jacques Maritz and Anban W. Pillay},
  booktitle = {Proceedings of the 5th Southern African Conference on {AI} Research (SACAIR'24)},
  series    = {CCIS},
  volume    = {2326},
  publisher = {Springer},
  year      = {2024},
  pages     = {459--475},
  optdoi    = {10.1007/978-3-031-78255-8_27}
}

@article{MarekST93,
  author  = {V. Wiktor Marek and
             Grigori F. Shvarts and
             Miroslaw Truszczynski},
  title   = {Modal Nonmonotonic Logics: Ranges, Characterization, Computation},
  journal = {J. {ACM}},
  volume  = {40},
  number  = {4},
  pages   = {963--990},
  year    = {1993},
  url     = {https://doi.org/10.1145/153724.153773},
  optdoi  = {10.1145/153724.153773}
}

@inproceedings{EmmrichAS23,
  author    = {Florian Emmrich and
               Lucía {Gómez Álvarez} and
               Hannes Strass},
  editor    = {Fumiaki Toyoshima and
               Megan Katsumi and
               Guendalina Righetti and
               Stefano De Giorgis and
               Maria M. Hedblom and
               Oliver Kutz and
               Boyan Brodaric and
               Michael Gr{\"{u}}ninger and
               Torsten Hahmann and
               Damion Dooley Simon and
               Matthew Lange and
               Hande K{\"{u}}{\c{c}}{\"{u}}k{-}McGinty and
               Anoosha Sehar Simon and
               Rhiannon Cameron and
               Bart Gajderowicz and
               Daniela Rosu and
               Janna Hastings and
               Davide Audrito and
               Luigi Di Caro and
               Francesca Grasso and
               Roberto Nai and
               Emilio Sulis and
               Loris Bozzato and
               Cogan Shimizu and
               Antoine Zimmermann and
               Riccardo Baratella and
               Stefano Borgo and
               Sergio de Cesare and
               Tiago Prince Sales},
  title     = {Automated Reasoning Support for Standpoint-OWL 2},
  booktitle = {Proceedings of the Joint Ontology Workshops 2023 Episode {IX:} The Qu{\'{e}}bec Summer of Ontology co-located with the 13th International Conference on Formal Ontology in Information Systems {(FOIS 2023)}, Sherbrooke, Qu{\'{e}}bec, Canada, July 19--20, 2023},
  series    = {{CEUR} Workshop Proceedings},
  volume    = {3637},
  publisher = {CEUR-WS.org},
  year      = {2023},
  url       = {https://ceur-ws.org/Vol-3637/paper49.pdf}
}

@article{BidoitF91,
  author  = {Nicole Bidoit and
             Christine Froidevaux},
  title   = {Negation by Default and Unstratifiable Logic Programs},
  journal = {Theor. Comput. Sci.},
  volume  = {78},
  number  = {1},
  pages   = {86--112},
  year    = {1991},
  url     = {https://doi.org/10.1016/0304-3975(51)90004-7},
  optdoi  = {10.1016/0304-3975(51)90004-7}
}

@article{MarekT91,
  author  = {V. Wiktor Marek and
             Mirosław Truszczyński},
  title   = {Autoepistemic Logic},
  journal = {J. {ACM}},
  volume  = {38},
  number  = {3},
  pages   = {588--619},
  year    = {1991},
  url     = {https://doi.org/10.1145/116825.116836},
  optdoi  = {10.1145/116825.116836}
}

@inbook{DunneW09,
  author    = {Dunne, Paul E.
               and Wooldridge, Michael},
  editor    = {Simari, Guillermo
               and Rahwan, Iyad},
  title     = {Complexity of Abstract Argumentation},
  booktitle = {Argumentation in Artificial Intelligence},
  year      = {2009},
  publisher = {Springer US},
  address   = {Boston, MA},
  pages     = {85--104},
  isbn      = {978-0-387-98197-0},
  optdoi    = {10.1007/978-0-387-98197-0_5},
  url       = {https://doi.org/10.1007/978-0-387-98197-0_5}
}

@inproceedings{GomezAlvarezR25,
  title     = {{Putting Perspective into OWL [sic]: Complexity-Neutral Standpoint Reasoning for Ontology Languages via Monodic S5 over Counting Two-Variable First-Order Logic}},
  author    = {{Gómez Álvarez}, Lucía and Rudolph, Sebastian},
  booktitle = {{Proceedings of the 22nd International Conference on Principles of Knowledge Representation and Reasoning}},
  pages     = {366--375},
  year      = {2025},
  month     = {10},
  url       = {https://doi.org/10.24963/kr.2025/36},
  optdoi    = {10.24963/kr.2025/36}
}

\end{document}